\documentclass[12pt,oneside]{report}
\usepackage{index}
\usepackage{named}
\usepackage{frontmatter}
\usepackage{setspace} 
\usepackage[pdftex]{hyperref}
\usepackage[pdftex]{graphicx}
\usepackage{fancyhdr}

\usepackage{amsmath,amscd} 
\usepackage{amssymb}       
\usepackage[all]{xy}
\usepackage{graphicx}

\hypersetup{
  pdftitle={\thesistitle},
  pdfauthor={\thesisauthor},
  pdfkeywords={\thesisfield}
}

\usepackage[pdftex,letterpaper,left=1in,top=1in,bottom=1in,right=1in,includeheadfoot,headheight=15pt]{geometry}

\newlength{\defaultparindent}
\renewcommand{\doublespace} {\setstretch{1.5}}
\renewcommand{\singlespace} {\setstretch{1}}
\newcommand{\indentmode}  {\setlength{\parindent}{\defaultparindent}}
\newcommand{\noindentmode}{\setlength{\parindent}{0.0in}}

\usepackage{epsfig}
\usepackage{amsmath}
\usepackage{amssymb}
\usepackage{amsthm}
\usepackage{color}
\usepackage[raggedright,normalsize,sf,SF,hang]{subfigure}
\usepackage{caption}

\usepackage[toc,page,title,titletoc]{appendix}
\renewcommand{\restoreapp}{}
\renewcommand{\appendixname}{Appendix}
\usepackage[nottoc]{tocbibind}

\usepackage{bm}

\usepackage{amsmath,amssymb}
\usepackage{graphicx} 
\usepackage{subfigure}
\usepackage{algorithm}
\usepackage{algorithmic}
\usepackage{hyperref}
\usepackage{comment} 

\usepackage{sectsty}
\allsectionsfont{\sffamily}
\chapterfont{\centering \sffamily}

\newcommand{\dd}{{\mathrm{d}}}
\newcommand{\pd}[2]{\frac{\partial #1}{\partial #2}}

\newcommand{\avg}[1]{\left< #1 \right>}
\newcommand{\mb}{\mathbf}

\newcommand{\gij}[2]{T_{#1}\lp \mb x_{#2} | \mb x_{#1} \rp}
\newcommand{\gijr}[2]{\widetilde{T}_{#2}\lp \mb x_{#2} | \mb x_{#1} \rp}
\newcommand{\ygij}[2]{T_{\cup #1}\lp \mb y_{#2} | \mb y_{#1} \rp}
\newcommand{\ygijr}[2]{{\widetilde{T}}_{\cup #2}\lp \mb y_{#2} | \mb y_{#1} \rp}
\newcommand{\pit}[2]{\pi_{#1}\lp \mb x_{#2} \rp}
\newcommand{\Ept}[2]{E_{\pi_{#1}}\lp x_{#2} \rp}

\newcommand{\p}[2]{p_{#1}^{(#2)}}
\newcommand{\argmin}{\operatornamewithlimits{argmin}}
\newcommand{\argmax}{\operatornamewithlimits{argmax}}

\newcommand{\doublebar}{\bigl|\!\bigr|}

\def \R { {\mathbb{R}} }

\def \1 { {\mathbf{1}} }
\def \I { {\mathbf{I}} }

\newcommand{\mc}{\mathcal}
\newcommand{\lp}{\left(}
\newcommand{\rp}{\right)}

\makeindex

\begin{document}


\singlespace
\setcounter{page}{0} 
\pagestyle{empty}
\noindentmode

\begin{center}

\hfill

\vspace{54pt} 

\textbf{\thesisTitle}

\vspace{12pt}

by

\vspace{12pt}

{\thesisauthor}


\vspace{24pt}

A dissertation submitted in partial satisfaction \\
of the requirements for the degree of

\vspace{12pt}

{\thesisdegree}

\vspace{12pt}

in

\vspace{12pt}

{\thesisField}

\vspace{12pt}

in the

\vspace{12pt}

{\thesisDivision}

\vspace{12pt}

of the

\vspace{12pt}

{\thesisSchool}

\vspace{24pt}

Committee in charge:

\vspace{12pt}

{\thesiscommitteeone}   \\
{\thesiscommitteetwo}   \\
{\thesiscommitteethree} \\
{\thesiscommitteefour} 

\vspace{24pt}

{\thesissemester} {\thesisyear} 

\end{center}

%
%
%
%


\clearpage
\setcounter{page}{-2} 
\thispagestyle{empty}
\doublespace
\indentmode

\begin{center}

\hfill

\vspace{2in}

{\thesistitle}

\vspace{12pt}

Copyright {\copyright} {\thesisyear}

\vspace{12pt}

by

\vspace{12pt}

{\thesisauthor}

\end{center}


\clearpage

\makeatletter
\newcommand{\ps@abstract}{
  \renewcommand{\@oddhead}{}
  \renewcommand{\@evenhead}{}
  \renewcommand{\@oddfoot}{\hfill \arabic{page} \hfill}
  \renewcommand{\@evenfoot}{\hfill \arabic{page} \hfill}
}
\makeatother

\pagestyle{abstract}
\pagenumbering{Roman} 
\doublespace
\indentmode

\begin{center}

\textbf{Abstract}

\hfill

{\thesistitle}

by

{\thesisauthor}

{\thesisdegree} in {\thesisfield}

{\thesisschool}

{\thesiscommitteeone}

{\thesiscommitteetwo}

\hfill

\end{center}
\singlespace

High dimensional probabilistic models are used for many modern scientific and engineering data analysis tasks.  Interpreting neural spike trains, compressing video, identifying features in DNA microarrays, and recognizing particles in high energy physics all rely upon the ability to find and model complex structure in a high dimensional space.  Despite their great promise, high dimensional probabilistic models are frequently computationally intractable to work with in practice.  In this thesis I develop solutions to overcome this intractability, primarily in the context of energy based models.

A common cause of intractability is that model distributions cannot be analytically normalized. Probabilities can only be computed up to a constant, making training exceedingly difficult. To solve this problem I propose `minimum probability flow learning', a variational technique for parameter estimation in such models. The utility of this training technique is demonstrated in the case of an Ising model, a Hopfield auto-associative memory, an independent component analysis model of natural images, and a deep belief network.

A second common difficulty in training probabilistic models arises when the parameter space is ill-conditioned. This makes gradient descent optimization slow and impractical, but can be alleviated using the natural gradient. I show here that the natural gradient can be related to signal whitening, and provide specific prescriptions for applying it to learning problems.

It is also difficult to evaluate the performance of models that cannot be analytically normalized, providing a particular challenge to hypothesis testing and model comparison.  To overcome this, I introduce a method termed `Hamiltonian annealed importance sampling,' which more efficiently estimates the normalization constant of non-analytically-normalizable models.  This method is then used to calculate and compare the log likelihoods of several state of the art probabilistic models of natural image patches. 

Finally, many tasks performed with a trained probabilistic model (for instance, image denoising or inpainting and speech recognition) involve generating samples from the model distribution, which is typically a very computationally expensive process.  I introduce a modification to Hamiltonian Monte Carlo sampling that reduces the tendency of sampling trajectories to double back on themselves, and enables statistically independent samples to be generated more rapidly.

Taken together, it is my hope that these contributions will help scientists and engineers to build and manipulate probabilistic models.

\doublespace


\clearpage
\pagestyle{plain}
\pagenumbering{roman}
\doublespace
\indentmode

\begin{center}

\singlespace

\textbf{Acknowledgements}

\doublespace

\end{center}

\hfill

\singlespace

Thank you to my advisor Bruno Olshausen, for countless thoughtful and inspiring conversations, and for giving me the freedom to pursue my interests; my mentor Mike DeWeese, for long nights working and innumerable helpful conversations; Tony Bell for identifying the interesting questions; Fritz Sommer for many interesting conversations, and a supply of reading material; Jack Culpepper, Peter Battaglino, Charles Cadieu, Jimmy Wang, Chris Hillar, Kilian Koepsell, Amir Khosrowshahi, Urs Koester, Pierre Garrigues, and the rest of the Redwood Center for diverse and esoteric interests, many, many fascinating conversations, and useful feedback.

\doublespace
%








\clearpage
\doublespace
\indentmode
\tableofcontents






\clearpage
\pagenumbering{arabic}
\doublespace
\indentmode
\pagestyle{plain}


\pagestyle{fancy}
\renewcommand{\sectionmark}[1]{}
\renewcommand{\chaptermark}[1]{%
 \markboth{\chaptername
 \ \thechapter.\ #1}{}} 
\fancyhead[L]{\sffamily \selectfont \leftmark \normalfont}
\fancyfoot[C]{\sffamily \selectfont \thepage \normalfont}
\fancyhead[R]{}

\singlespace

\chapter{Introduction}
\label{chap intro}

Scientists and engineers increasingly confront large and
complex data sets that defy traditional modeling and analysis
techniques.
For example, fitting well-established probabilistic models from physics to population neural activity recorded in retina \cite{Schneidman_Nature_2006,Shlens_JN_2006,Schneidman_Nature_2006} or cortex \cite{Tang:2008p13129,Marre:2009p13182,Yu:2008p4351} is currently impractical for populations of more than about 100 neurons \cite{Broderick:2007p2761}.  
Similar difficulties occur
in many other fields, including computer science \cite{mackay:book}, genomics \cite{genomics_review_2009}, and physics \cite{physics_param_est_book_2005}.  
Thus, development of new techniques to train, evaluate, and sample from complex probabilistic models is of fundamental importance to many scientific and engineering disciplines.

This thesis identifies and addresses a number of important difficulties that are encountered when working with these models.  I focus on energy-based models which cannot be normalized in closed form, posing unique challenges for learning. 
I begin this chapter with a review of probabilistic models, current state of the art parameter estimation methods, and techniques for estimating intractable normalization constants.  I end this chapter in Section \ref{intro contributions} with a summary of the contributions made in this thesis, all of which improve our ability to evaluate, train, or work with challenging probabilistic models.

\section{Parameter Estimation in Probabilistic Models}
\label{intro formalism}

The common goal of parameter estimation is to find the parameters that cause a probabilistic model to best
agree with a list $\mathcal{D}$ of (assumed iid) observations of the state of a system.  In this section we provide formalism for writing data and model distributions, introduce the canonical Kullback-Leibler (KL) divergence objective for parameter estimation, and present a number of relevant parameter estimation techniques.

\subsection{Distributions}\label{sec:distributions}

\subsubsection{Discrete Distributions}

\begin{figure}
\begin{center}
\begin{tabular}{cc}
	\includegraphics[width=0.2\linewidth]{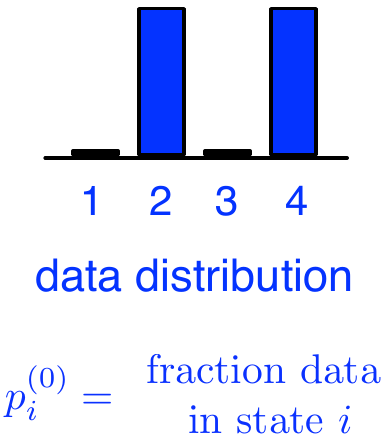}
	&
	\includegraphics[width=0.2\linewidth]{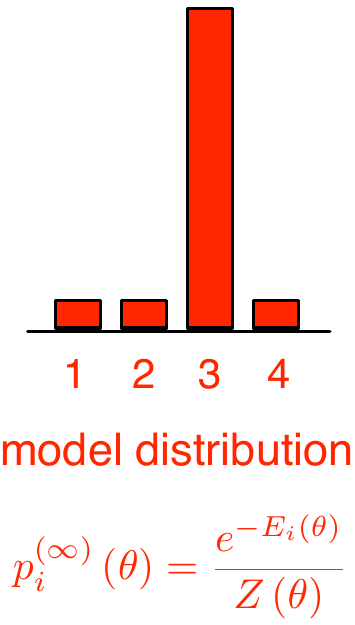}
\\
	(a) & (b)
\end{tabular}
\end{center}
\caption{Both data and model distributions can be viewed as vectors, with a dimensionality equal to the number of possible states, and a value for each entry equal to the probability of the corresponding state.  {\textbf{(a)}} illustrates a data distribution.  In this case there were an equal number of observations of states 2 and 4, and no observations of states 1 and 3, so states 2 and 4 each have probability 0.5.  {\textbf{(b)}} illustrates a model distribution parameterized by $\theta$, with the probability assigned to each state $i$ determined by the energy $E_i\left(\theta\right)$ assigned to that state and by a normalization constant $Z\left(\theta\right)$.
}
\label{fig:distributions histogram}
\end{figure}

The data distribution is represented by a vector $\mb{p}^{(0)}$, with
$p^{(0)}_i$ the fraction of the observations $\mathcal{D}$ in state $i$.  The superscript $(0)$ represents time $t=0$ under system
dynamics (which will be described for MPF in Section~\ref{sec:dynamics}).  For example, in
a two variable binary system, $\mb{p}^{(0)}$ would have
four entries representing the fraction of the data in states $00$, $01$,
$10$ and $11$ (Figure \ref{fig:distributions histogram}).

Our goal is to find the parameters $\theta$ that cause a model distribution
$\mb{p}^{(\infty)}\left( \theta \right)$ to best match the data
distribution $\mb{p}^{(0)}$. The superscript
$(\infty)$ on the model distribution indicates that this is the equilibrium distribution reached after
running the dynamics (again described for MPF in Section~\ref{sec:dynamics}) for infinite time. Without loss of generality, we assume the
model distribution is of the form
\begin{align}
\label{eqn:p infinity}
p^{(\infty)}_i\left( \theta \right) &= \frac
       {\exp \left( -E_i\left( \theta \right) \right) }
       {Z\left(\theta\right)}
       ,
\end{align}
where $\mb{E}\left( \theta \right)$ is referred to as the energy function,
and the normalizing factor $Z\left(\theta\right)$ is the partition
function,
\begin{align}
Z\left(\theta\right) = \sum_i \exp \left( -E_i\left( \theta \right) \right)
\end{align}
(this can be thought of as a Boltzmann distribution of a physical
system with $k_B T$ set to 1).


\subsubsection{Continuous Distributions}

Data and model distributions over a continuous state space $\mb x \in \mathcal R^d$ take the forms,
\begin{align}
\label{eqn:p cont zero}
p^{(0)}\left( \mb x \right) &= 
\frac{1}{\left| \mathcal D \right|}
\sum_{\mb x' \in \mathcal D} \delta\left( \mb x - \mb x' \right)
\\
\label{eqn:p cont infinity}
p^{(\infty)}\left( \mb x; \theta \right) &= \frac
       {\exp \left( -E\left( \mb x; \theta \right) \right) }
       {Z\left(\theta\right)}
       ,
\end{align}
respectively, where $\left| \mathcal D \right|$ is the number of observations, $\delta\left(\cdot\right)$ is the Dirac delta function, and $Z\left(\theta\right)$ is the partition function
\begin{align}
\label{eq Z intro}
Z\left(\theta\right) = \int d\mb x \exp \left( -E\left( \mb x;  \theta \right) \right)
.
\end{align}

\subsection{Kullback-Leibler (KL) Divergence}

The standard goodness of fit measure of a model distribution to a data distribution is the KL divergence between data and model distributions \cite{cover_thomas},
\begin{align}
\label{eq KL}
D_{KL}\left( \mathbf p^{(0)} \doublebar  \mathbf p^{(\infty)}\left( \theta \right) \right) & = \sum_i p_i^{(0)} \log p_i^{(0)}
 - \sum_i p_i^{(0)} \log p_i^{(\infty)} \left( \theta \right) 
.
\end{align}
Because the first term in \ref{eq KL} is constant, and the second term is the negative log likelihood of the model distribution, finding the parameters which minimize the KL divergence $D_{KL}\left( \mathbf p^{(0)} \doublebar  \mathbf p^{(\infty)}\left( \theta \right) \right)$ is equivalent to finding the parameters which minimize the negative log likelihood, and which maximize the likelihood.  Given a list of data points $\mathcal D$,
\begin{align}
\label{eq KL L comparison}
\argmin 
D_{KL}\left( \mathbf p^{(0)} \doublebar  \mathbf p^{(\infty)}\left( \theta \right) \right) & = 
\argmin\left[- \sum_i p_i^{(0)} \log p_i^{(\infty)} \left( \theta \right) \right] \\
&= \argmin\left[ -\log L\left(\theta\right)\right] = \argmax L\left(\theta\right) \\
L\left(\theta\right) & = 
\prod_{i \in \mathcal D} p_i^{(\infty)} \left( \theta \right)
.
\end{align}
The gradient of the log likelihood is
\begin{align}
\label{eq:log like grad}
\pd{\left[\log L\left(\theta\right)\right]}{\theta}
 & =
-\sum_i p_i^{(0)} \pd{E_i\left( \theta \right)}{\theta}
+\sum_i p_i^{(\infty)}\left(\theta\right) \pd{E_i\left( \theta \right)}{\theta}
.
\end{align}

\subsection{Parameter Estimation Techniques}

Exact parameter estimation involves evaluation of $D_{KL}\left( \mathbf p^{(0)} \doublebar  \mathbf p^{(\infty)}\left( \theta \right) \right)$ or its derivatives.  Unfortunately, this involves evaluating $Z\left(\theta\right)$, which includes the sum over all system states in Equation \ref{eq Z intro}, or a similar integral in the case of a continuous state space.  This sum is intractable for most systems of a reasonable size - for instance involving $2^{100}$ terms for a 100 bit binary system.  For this reason, exact parameter estimation is frequently impractical.

Many approaches exist for approximate parameter estimation, including mean field theory and
its expansions, variational Bayes techniques and a variety of sampling or numerical
integration based methods \cite{Tanaka:1998p1984,Kappen:1997p6,Jaakkola:1997p4985,haykin2008nnc}.  The approaches which most closely relate to the new techniques introduced in this thesis include contrastive divergence (CD),
developed by Hinton, Welling and Carreira-Perpi\~{n}\'an \cite{Welling:2002p3,Hinton02,CD_04}, Hyv\"arinen's score matching (SM) \cite{Hyvarinen05}, Besag's pseudolikelihood (PL) \cite{besag}, Lyu's Minimum KL Contraction \cite{mkc}, and the minimum velocity learning
framework proposed by Movellan \cite{Movellan:2008p7643,movellan2008cdg,Movellan93}. 

\subsubsection{Contrastive Divergence}
\label{sec CD intro}
Contrastive divergence \cite{Welling:2002p3,Hinton02} is a variation on steepest gradient descent of the maximum (log) likelihood (ML) objective function.  Rather than integrating over the full model distribution, CD approximates the partition function term in the gradient by averaging over the distribution obtained after taking a few, or only one, Markov chain Monte Carlo (MCMC) steps away from the data distribution (Equation \ref{eq:CD-k intro}).  CD is frequently abbreviated CD-$k$, where $k$ is the number of MCMC steps taken away from the data distribution.  
Qualitatively, one can imagine that the data distribution is contrasted against a distribution that has evolved only a small distance towards the model distribution, whereas it would be contrasted against the true model distribution in traditional MCMC approaches.  Although CD is not guaranteed to converge to the right answer, or even to a fixed point, it has proven to be an effective and fast heuristic for parameter estimation \cite{MacKay:2001p8372,Yuille04}.  The CD-$k$ update rule can be written
\begin{align}
\label{eq:CD-k intro}
\Delta \theta_{CD} & \propto
-\sum_i p_i^{(0)} \pd{E_i\left( \theta \right)}{\theta}
+\sum_i p_i^{(k)} \pd{E_i\left( \theta \right)}{\theta}
,
\end{align}
where $p^{(k)}$ is the distribution resulting after applying $k$ MCMC updates to samples from $p^{(0)}$.  This update rule should be compared against the gradient of the log likelihood in Equation~\ref{eq:log like grad}.

\subsubsection{Score Matching}
\label{sec SM intro}
Score matching \cite{Hyvarinen05} is a method that learns parameters in a probabilistic model with a continuous state space using only derivatives of the energy function evaluated over the data distribution.  This sidesteps the need to explicitly sample or integrate over the model distribution. In score matching one minimizes the expected square distance of the score function with respect to spatial coordinates given by the data distribution from the similar score function given by the model distribution.  The score function is the gradient of the log likelihood.  A number of connections have been made between score matching and other learning techniques \cite{Hyvarinen:2007p5984,sohldickstein,Movellan:2008p7643,siwei2009}.  The score matching objective function can be written
\begin{align}
\label{eq:SM intro}
K_{SM}\left( \theta \right) & =
 \sum_{\mb x\in\mathcal{D}} \left[ \frac{1}{2}\nabla E(\mb x; \theta)\cdot \nabla E(\mb x; \theta) - \nabla^2 E(\mb x; \theta) \right]
.
\end{align}
Parameter estimation is performed by finding $\argmin_\theta K_{SM}\left( \theta \right)$.  Performing gradient descent on $K_{SM}\left( \theta \right)$ involves computing 3rd derivatives of $E(\mb x; \theta)$, which is frequently unwieldy.

\subsubsection{Pseudolikelihood}

Pseudolikelihood \cite{besag} approximates the joint probability distribution of a collection of random variables by a computationally tractable product of conditional distributions, where each factor is the distribution of a single random variable conditioned on the others. This approach often leads to surprisingly good parameter estimates, despite the extreme nature of the approximation.  Recent work suggests that pseudolikelihood is a consistent estimator of model parameters \cite{mkc}, meaning that if the data distribution has the same form as the model distribution then in the limit of infinite data the exact correct distribution will be recovered.  The pseudolikelihood objective function can be written
\begin{align}
\label{eq:PL intro}
K_{PL}\left(\theta\right) & =
\sum_{\mb x\in\mathcal{D}}
\sum_m
\log p\left( x_m | \mb x_{\backslash m}; \theta \right)
,
\end{align}
where $m$ indexes the dimensions of the state space, and the expression $p\left( x_m | \mb x_{\backslash m} \right)$ indicates the probability distribution over the $m$th dimension of the state space conditioned on the remaining dimensions.  For clarity we have written the pseudolikelihood objective function for a continuous state space, but it is defined for both continuous and discrete state spaces.

\subsubsection{Minimum Velocity Learning}

Minimum velocity learning is an approach recently proposed by
Movellan \cite{Movellan:2008p7643} that recasts a number of the ideas behind CD, treating the
minimization of the initial dynamics away from the data distribution as the goal
itself rather than a surrogate for it.  Rather than directly minimize the difference between the data and the model, Movellan's proposal is to introduce system dynamics that have the model as their equilibrium distribution,
and minimize the initial flow of probability away from the data under those
dynamics. If the model looks exactly like the data there
will be no flow of probability, and if model and data are similar the flow of probability will tend to be minimal.  Movellan applies this intuition to the specific case of distributions over continuous state spaces evolving via diffusion dynamics, and recovers the score matching objective function (Equation \ref{eq:SM intro}). The velocity in minimum velocity learning is the difference in average drift velocities between particles diffusing under the model distribution and particles diffusing under the data distribution.

\subsubsection{Minimum KL contraction}

Minimum KL contraction \cite{mkc} involves applying a special class of mapping (a contraction mapping) to both the data and model distributions, and minimizing the amount by which this mapping shrinks the KL divergence between the data and model distributions.  As the KL divergence between the data and model distributions becomes more similar there is less room for the contraction mapping to further shrink it, and the KL contraction objective becomes smaller.  Like minimum probability flow (introduced in Chapter \ref{chap MPF}), minimum KL contraction appears to be a generalization of a number of existing parameter estimation techniques based on ``local'' information about the model distribution.
%

\section{Hamiltonian Monte Carlo Sampling}
\label{sec HMC intro}

Generating samples from probability distributions over high dimensional state spaces is frequently extremely expensive.  Hamiltonian Monte Carlo (HMC) \cite{Horowitz1991,Neal:HMC} is a family of techniques for fast sampling in continuous state spaces, which work by extending the state space to include auxiliary momentum variables, and then simulating Hamiltonian dynamics from physics in order to traverse long iso-probability trajectories which rapidly explore the state space.

In HMC, the state space $\mb x \in \R^M$ is expanded to include auxiliary momentum variables $\mb v \in \R^M$ with a simple independent distribution,
\begin{align}
p\left( \mb v \right) &= \frac{
\exp
\left[ -\frac{1}{2} \mb v^T \mb v
\right]
}{\left( 2\pi \right)^{\frac{M}{2}}} 
.
\end{align}
The joint distribution over $\mb x$ and $\mb v$ is then
\begin{align}
p\left( \mb x, \mb v \right) &= p\left( \mb x \right)p\left( \mb v \right) = \frac{\exp\left[
	-H\left( \mb x, \mb v \right)
\right]}{Z_H}  \\
H\left( \mb x, \mb v \right) &= E\left(\mb x\right) + \frac{1}{2}\mb v^T\mb v
,
\end{align}
where $H\left( \mb x, \mb v \right)$ is the total energy, $Z_H$ is a normalization constant, and $E\left(\mb x\right)$ and $\frac{1}{2}\mb v^T\mb v$ are analogous to the potential and kinetic energies in a physical system.

Sampling alternates between drawing the momentum $\mb v$ from its marginal distribution $p\left( \mb v \right)$, and simulating Hamiltonian dynamics for the joint system described by $H\left( \mb x, \mb v \right)$.  Hamiltonian dynamics are described by the differential equations
\begin{align}
\label{ham dyn 1}
\dot{\mb x} &= \pd{H\left( \mb x, \mb v \right)}{\mb v} = \mb v \\
\label{ham dyn 2}
\dot{\mb v} &= -\pd{H\left( \mb x, \mb v \right)}{\mb x} = -\pd{E\left(\mb x\right)}{\mb x}
.
\end{align}
Because Hamiltonian dynamics conserve the total energy $H\left( \mb x, \mb v \right)$ and thus the joint probability $p\left( \mb x, \mb v \right)$, and preserve volume in the joint space of $\mb x$ and $\mb v$, new samples proposed by integrating Equations \ref{ham dyn 1} and \ref{ham dyn 2} can be accepted with probability one, yet also have traversed a large distance from the previous sample.  HMC thus allows independent samples to be rapidly drawn from $p\left( \mb x, \mb v \right)$.  Because $p\left( \mb x, \mb v \right)$ is factorial, samples from $p\left( \mb x \right)$ can be recovered by discarding the $\mb v$ variables and taking the marginal distribution over $\mb x$.  Additional issues which must be addressed in implementation involve choosing a numerical integration scheme for the dynamics, and correctly accounting for discretization errors.
 
\section{Annealed Importance Sampling}
\label{sec AIS intro}

Annealed Importance Sampling (AIS) \cite{Neal:AIS} is a sequential Monte Carlo method \cite{Moral:2006p12473} which allows the partition function of a non-analytically-normalizable distribution to be estimated in an unbiased fashion. This is accomplished by starting at a distribution with a known normalization, and gradually transforming it into the distribution of interest through a chain of Markov transitions. Its practicality depends heavily on the chosen Markov transitions.  Here we review the derivations of both importance sampling and annealed importance sampling.  An extension of annealed importance sampling to better incorporate Hamiltonian Monte Carlo is presented in Chapter \ref{chap HAIS}.

\subsection{Importance Sampling}

Importance sampling \cite{Kahn:1953p12988} allows an unbiased estimate $\hat Z_p$ of the partition function (or normalization constant) $Z_p$ of a non-analytically-normalizable target distribution $p\lp \mb x \rp$ over $\mb x \in \mathbb R^M$,
\begin{align}
p\lp \mb x \rp &= \frac
{e^{-E_p\lp \mb x \rp}}
{Z_p}   \label{eq qx} \\
Z_p & = \int d\mb x\  
{e^{-E_p\lp \mb x \rp}}
,
\end{align}
to be calculated.  This is accomplished by averaging over samples $\mc S_q$ from a proposal distribution $q\lp \mb x \rp$,
\begin{align}
q\lp \mb x \rp &= \frac
{e^{-E_q\lp \mb x \rp}}
{Z_q} \\
Z_p & = \int d\mb x\  q\lp \mb x \rp
\frac
{e^{-E_p\lp \mb x \rp}}
{q\lp \mb x \rp}
                       \label{eq IS q ratio}
\\
\hat Z_p & = \frac{1}{\left| \mc S_q \right|}\sum_{x \in \mc S_q}
\frac
{e^{-E_p\lp \mb x \rp}}
{q\lp \mb x \rp}
                       \label{eq IS}
,
\end{align}
where $\left| \mc S_q \right|$ is the number of samples.  $q\lp \mb x \rp$ is chosen to be easy both to sample from and to evaluate exactly, and must have support everywhere that $p\lp \mb x \rp$ does.
Unfortunately, unless $q\lp \mb x \rp$ has significant mass everywhere $p\lp \mb x \rp$ does, it takes an impractically large number of samples from $q\lp \mb x \rp$ for $\hat Z_p$ to accurately approximate $Z_p$\footnote{
The expected variance of the estimate $\hat Z_p$ is given by an $\alpha$-divergence between $p\lp \mb x \rp$ and $q\lp \mb x \rp$, times a constant and plus an offset - see \cite{Minka:2005p12627}.
}.

\subsection{Annealed Importance Sampling}

Annealed importance sampling \cite{Neal:AIS} extends the state space $\mb x$ to a series of vectors, $\mb X = \left\{ \mb x_{1}, \mb x_{2} \ldots \mb x_{N} \right\}$, $\mb x_{n} \in \mathbb R^M$.  It then transforms the proposal distribution $q\lp \mb x \rp$ to a distribution $Q\lp \mb X \rp$ over $\mb X$, by setting $q\lp \mb x \rp$ as the distribution over $\mb x_1$ and then multiplying by a series of Markov transition distributions,
\begin{align}
Q\lp \mb X \rp & = q\lp \mb x_{1} \rp \prod_{n=1}^{N-1} \gij{n}{n+1}
,
\end{align}
where $\gij{n}{n+1}$ represents a \emph{forward} transition distribution from $\mb x_{n}$ to $\mb x_{n+1}$.
The target distribution $p\lp \mb x \rp$ is similarly transformed to become a reverse chain $P\lp \mb X \rp$, starting at $\mb x_{N}$, over $\mb X$,
\begin{align}
P\lp \mb X \rp & = \frac
   {e^{-E_p\lp \mb x_{N} \rp}}
   {Z_p}
   \prod_{n=1}^{N-1} \gijr{n+1}{n}
,
\end{align}
where $\gijr{n+1}{n}$ is a \emph{reverse} transition distribution from $\mb x_{n+1}$ to $\mb x_{n}$.  The transition distributions are, by definition, normalized (eg, $\int d\mb x\ _{n+1} \gij{n}{n+1} = 1$).

In a similar fashion to Equations \ref{eq IS q ratio} and \ref{eq IS}, samples $\mc S_Q$ from the forward proposal chain $Q\lp \mb X \rp$ can be used to estimate the partition function $Z_p$,
\begin{align}
Z_p & =     \int d\mb x_{N}\  {e^{-E_p\lp \mb x_{N} \rp}} \\
& =     \int d\mb x_{N}\  {e^{-E_p\lp \mb x_{N} \rp}}
   \int d\mb x_{N-1}\  \gijr{N}{N-1}
   \cdots
   \int d\mb x_{1}\  \gijr{2}{1} \label{eq Z_p int1}
\end{align}
(note that all integrals but the first in Equation \ref{eq Z_p int1} go to 1)
\begin{align}
Z_p & =     \int d\mb X\  Q\lp \mb X \rp \frac
       {e^{-E_p\lp \mb x_{N} \rp}}
       {Q\lp \mb X \rp}
   \gijr{N}{N-1}
     \cdots \
   \gijr{2}{1}        \\
\hat Z_p & = \frac{1}{\left| \mc S_Q \right|}\sum_{X \in \mc S_Q}
   \frac
       {e^{-E_p\lp \mb x_{N} \rp}}
       {q\lp \mb x_{1} \rp}
   \frac
       {\gijr{2}{1}}
       {\gij{1}{2}}
     \cdots
   \frac
       {\gijr{N}{N-1}}
       {\gij{N-1}{N}}
                       \label{eq AIS general}
.
\end{align}

In order to further define the transition distributions, Neal introduces intermediate distributions $\pi_{n}\lp \mb x \rp$ between $q\lp \mb x \rp$ and $p\lp \mb x \rp$,
\begin{align}
\pi_{n}\lp \mb x \rp & = \frac{e^{-E_{\pi_{n}}\lp x \rp
}}{Z_{\pi_{n}}} \\
E_{\pi_{n}}\lp \mb x \rp & = \lp1-\beta_{n}\rp E_q\lp \mb x \rp + \beta_{n} E_p\lp \mb x \rp
,
\end{align}
where the mixing fraction $\beta_{n} = \frac{n}{N}$ for all the results in this thesis.
$\gij{n}{n+1}$ is then chosen to be any Markov chain transition for $\pi_{n}\lp \mb x \rp$, meaning that it leaves $\pi_{n}\lp \mb x \rp$ invariant
\begin{align}
T_{n} \circ \pi_{n} & = \pi_{n}
.
\end{align}
The reverse direction transition distribution $\gijr{n+1}{n}$ is set to the reversal of $\gij{n}{n+1}$,
\begin{align}
\gijr{n+1}{n} & = \gij{n}{n+1} \frac
{\pit{n}{n}}
{\pit{n}{n+1}}
.
\end{align}
Equation \ref{eq AIS general} thus reduces to
\begin{align}
\hat Z_p & = \frac{1}{\left| \mc S_Q \right|}\sum_{X \in \mc S_Q}
   \frac
       {e^{-E_p\lp \mb x_{N} \rp}}
       {q\lp \mb x_{1} \rp}
   \frac
       {\pit{1}{1}}
       {\pit{1}{2}}
     \cdots
   \frac
       {\pit{N-1}{N-1}}
       {\pit{N-1}{N}} \\
& =     \frac{1}{\left| \mc S_Q \right|}\sum_{X \in \mc S_Q}
   \frac
       {e^{-E_p\lp \mb x_{N} \rp}}
       {q\lp \mb x_{1} \rp}
   \frac{e^{
       -\Ept{1}{1}
   }}
   {e^{
       - \Ept{1}{2}
   }}
   \cdots
   \frac{e^{
       -\Ept{N-1}{N-1}
   }}
   {e^{
       - \Ept{N-1}{N}
   }}
                       \label{eq AIS}
.
\end{align}

If the number of intermediate distributions $N$ is large, and the transition distributions $\gij{n}{n+1}$ and $\gijr{n+1}{n}$ mix effectively, then the distributions over intermediate states $\mb x_{n}$ will be nearly identical to $\pit{n}{n}$ in both the forward and backward chains.  $P\lp \mb X \rp$ and $Q\lp \mb X \rp$ will then be extremely similar to one another, and the variance in the estimate $\hat Z_p$ will be extremely low\footnote{
There is a direct mapping between annealed importance sampling and the Jarzynski equality in non-equilibrium thermodynamics --- see \cite{Jarzynski:1997p12846}.  It follows from this mapping, and the reversibility of quasistatic processes, that the variance in $\hat Z_p$ can be made to go to 0 if the transition from $q\lp \mb x_{1}\rp$ to $p\lp\mb x_{N}\rp$ is sufficiently gradual.
}.  If the transitions $\gij{n}{n+1}$ do a poor job mixing, then the marginal distributions over $\mb x_{n}$ under $P\lp \mb X \rp$ and $Q\lp \mb X \rp$ will look different from $\pit{n}{n}$.  The estimate $\hat Z_p$ will still be unbiased, but with a potentially larger variance.  Thus, to make AIS practical, it is important to choose Markov transitions $\gij{n}{n+1}$ for the intermediate distributions $\pi_{n}\lp \mb x \rp$ that mix quickly.

\section{Contributions of this Thesis}
\label{intro contributions}

In this thesis I attempt to solve several of the problems that arise in probabilistic modeling.  
I began in Chapter \ref{chap intro} by reviewing existing techniques for working with intractable probabilistic models. 

One of the most significant problems working with probabilistic models is that the majority of them cannot be analytically normalized. Therefore the probabilities they assign to states cannot be exactly computed, and are expensive even to approximate. 
Training a model with an intractable normalization constant is extremely difficult using traditional methods based on sampling.  I present an alternative technique for parameter estimation in such models, Minimum Probability Flow (MPF), in Chapter \ref{chap MPF}.


In Chapter \ref{chap MPF examples} I present experiments demonstrating the effectiveness of MPF for a number of applications. These quantitative results include comparisons of estimation speed and quality for an Ising model, the application of MPF to storing memories in a Hopfield auto-associative memory, and an evaluation of estimation quality for an Independent Component Analysis (ICA) model and a Deep Belief Network (DBN).



Difficulties in training probabilistic models can stem from ill conditioning of the model's parameter space as well as from an inability to analytically normalize the model.   In Chapter \ref{chap natgrad} I review how an ill conditioned parameter space can undermine learning, and present a novel interpretation of the natural gradient, a common technique for dealing with this ill conditioning.  In addition, I present tricks and specific prescriptions for applying the natural gradient to learning problems.

Even after a probabilistic model has been trained, it remains difficult to objectively judge and compare its performance to that of other models unless it can be normalized. To address this, in Chapter \ref{chap HAIS} Hamiltonian Annealed Importance Sampling (HAIS) is presented. This is a method which can be used for more efficient log likelihood estimation which combines Hamiltonian Monte Carlo (HMC) with Annealed Importance Sampling (AIS). It is then applied to compare the log likelihoods of several state of the art probabilistic models of natural image patches.


Finally, many of the tasks commonly performed with probabilistic models, for instance image denoising or inpainting \cite{roth2005fields} and speech recognition \cite{zweig1998speech}, require samples from the model distribution. Generating those samples has a high computational cost, often making it the bottleneck in a machine learning task. In Chapter \ref{chap HMC} an extension to HMC sampling is introduced which reduces the frequency with which sampling trajectories double back on themselves, and thus enables statistically independent samples to be generated more rapidly.

%


Additional research involving high dimensional probabilistic models, not incorporated into this thesis, includes developing multilinear generative models for natural scenes \cite{Culpepper2011}, training Lie groups to describe the transformations which occur in natural video \cite{Sohl-Dickstein2010,DCC_11}, exploring the statistical structure of MRI and CT scans of breast tissue \cite{Abbey2009}, applying a super-resolution algorithm to images from the Mars Exploration Rover Panoramic Camera \cite{Hayes2011,Grotzinger2005,Bell2004a,Bell2004}, photometric modeling of Martian dust \cite{Kinch2007,Johnson2006}, and modeling of camera systems on the Mars Exploration Rover \cite{Bell2006,Herkenhoff2003}.  


%
%


\chapter{Minimum Probability Flow}
\label{chap MPF}

\begin{figure}
\begin{center}
\parbox[c]{0.8\linewidth}{
\includegraphics[width=1.0\linewidth]{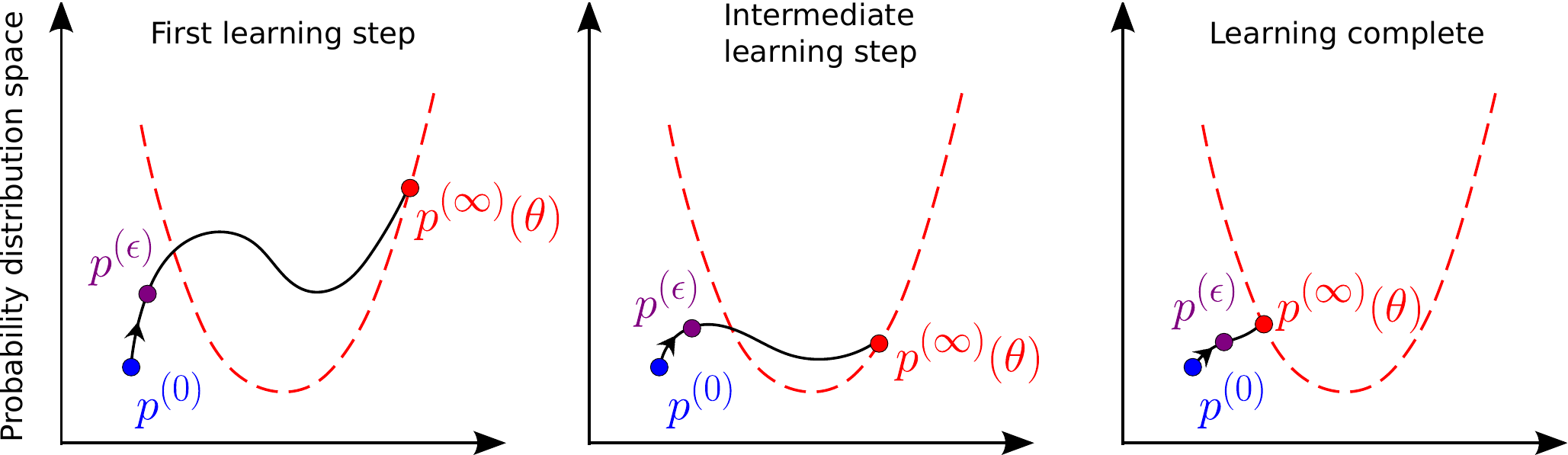}
} \\ \vspace{0.15in} \textsc{Progression of Learning} \vspace{-0.15in}
\end{center}
\caption{
An illustration of parameter estimation using minimum probability flow (MPF). In each panel, the axes represent the space of all probability distributions. The three successive panels illustrate the sequence of parameter updates that occur during learning. The dashed red curves indicate the family of model distributions $\mb{p}^{(\infty)}(\theta)$ parametrized by $\theta$.  The black curves indicate deterministic dynamics that transform the data distribution $\mb{p}^{(0)}$ into the model distribution $\mb{p}^{(\infty)}(\theta)$.  Under maximum likelihood learning, model
parameters $\theta$ are chosen so as to minimize the Kullback--Leibler (KL) divergence
between the data distribution $\mb{p}^{(0)}$ and the model distribution
$\mb{p}^{(\infty)}(\theta)$. Under MPF, however, the KL divergence
between $\mb{p}^{(0)}$ and $\mb{p}^{(\epsilon)}$ is minimized instead, where
$\mb{p}^{(\epsilon)}$ is the distribution obtained by initializing the dynamics at the data distribution $\mb{p}^{(0)}$ and then evolving them for
an infinitesimal time $\epsilon$. Here we represent graphically how parameter updates that pull
$\mb{p}^{(\epsilon)}$ towards $\mb{p}^{(0)}$ also tend to pull
$\mb{p}^{(\infty)}(\theta)$ towards $\mb{p}^{(0)}$.
}
\label{fig:KL}
\end{figure}

As discussed in Chapter \ref{chap intro}, most probabilistic learning techniques require calculating the
normalization factor, or partition function, of the probabilistic model in question, or at least calculating
its gradient. For the overwhelming majority of models there are no known analytic solutions, and this calculation is intractable.  In this chapter we will present a technique for parameter estimation in probabilistic models, even in cases where the normalization factor cannot be calculated.  Material in this chapter is taken from \cite{MPF_ICML,SohlDickstein2011a,SohlDickstein2009p8641}.

Our goal is to find the parameters that cause a probabilistic model to best
agree with a list $\mathcal{D}$ of (assumed iid) observations of the state of a system.  We
will do this by introducing deterministic dynamics that guarantee the transformation of the data
distribution into the model distribution, and then minimizing the KL divergence between the data distribution and the distribution that results from running those dynamics for a short time $\epsilon$ (see Figure~\ref{fig:KL}).  Formalism used below is introduced in Section \ref{intro formalism}.

\section{Dynamics}\label{sec:dynamics}
\begin{figure}
\center{
\parbox[c]{0.65\linewidth}{
\center{
\includegraphics[width=0.95\linewidth]{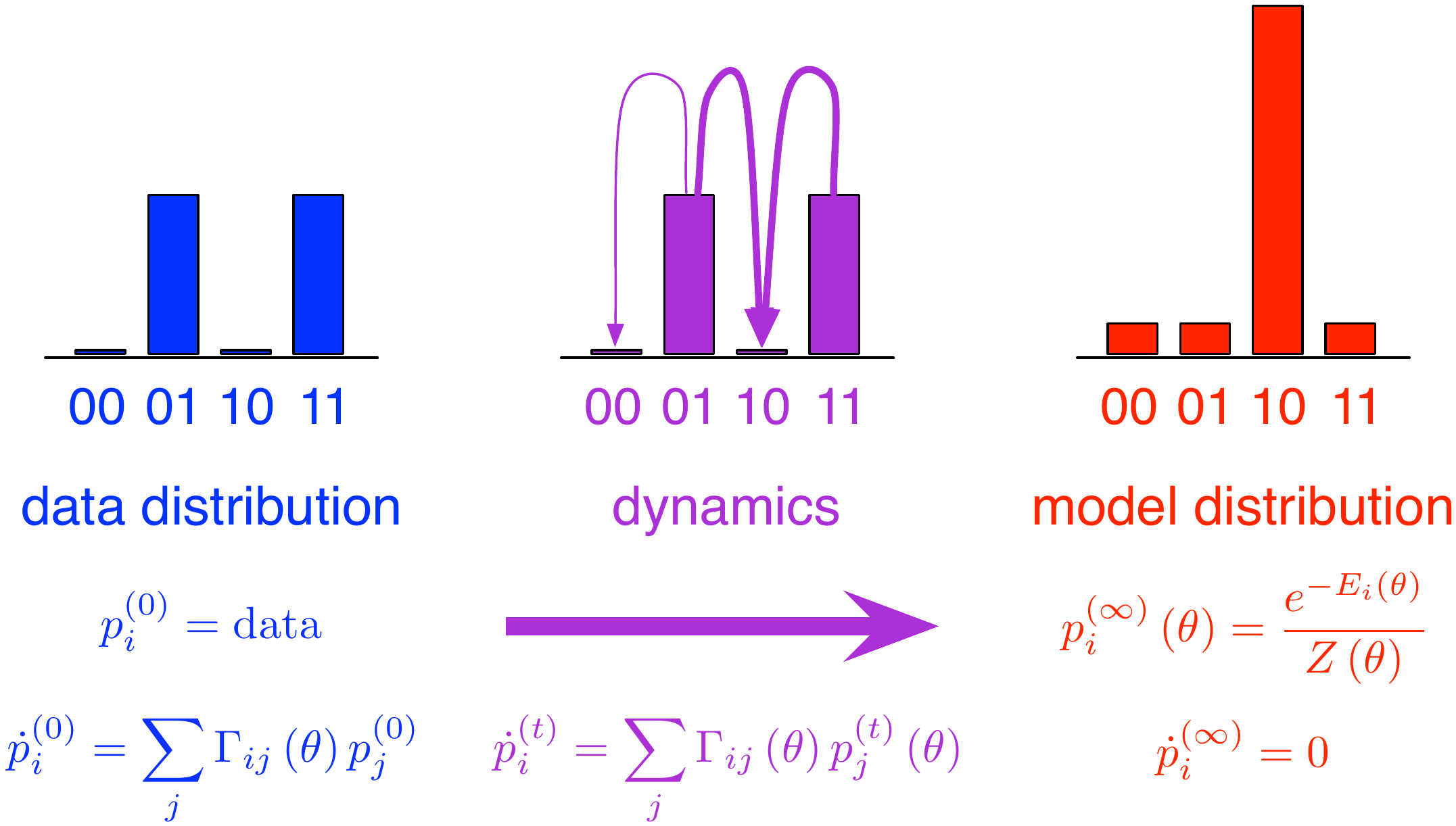}
}
}
}
\caption{
Dynamics of minimum probability flow learning. Model dynamics represented by
the probability flow matrix $\mb{\Gamma}$ ({\it middle}) determine how probability flows from
the empirical histogram of the sample data points ({\it left}) to the equilibrium
distribution of the model ({\it right}) after a sufficiently long time. In
this example there are only four possible states for the system, which consists
of a pair of binary variables, and the particular model parameters favor state
$10$ whereas the data falls on other states.
}
\label{fig:dynamics}
\end{figure}
Most Monte-Carlo algorithms rely on two core concepts from statistical physics,
the first being conservation of probability as enforced by the master equation for the time evolution of a distribution $\mb p^{(t)}$ \cite{Pathria:1972p5861}:
\begin{align}\label{eq:mastereqn}
\dot{p}_i^{(t)} = \sum_{j\neq i} \Gamma_{ij}(\theta)\,p_j^{(t)} - \sum_{j\neq i} \Gamma_{ji}(\theta)\, p_i^{(t)},
\end{align}
where $\dot{p}_i^{(t)}$ is the time derivative of $p_i^{(t)}$.  Transition rates $\Gamma_{ij}(\theta)$, for $i \neq j$, give the rate at which probability flows from a
state $j$ into a state $i$.
The first term of Equation~\eqref{eq:mastereqn} captures the flow of probability out of other states $j$ into the state $i$, and the second
captures flow out of $i$ into other states $j$.
The dependence on $\theta$ results from the requirement that the chosen dynamics cause $\mb{p}^{(t)}$ to flow to the equilibrium distribution $\mb{p}^{(\infty)}(\theta)$. For readability, explicit dependence on $\theta$ will be dropped except where necessary.
If we choose to set the diagonal elements of $\mb{\Gamma}$ to obey $\Gamma_{ii} = -\sum_{j\neq i}\Gamma_{ji}$, then we can write the
dynamics as
\begin{align}\label{eq:dynamics}
\dot{\mb{p}}^{(t)} = \mb{\Gamma}\mb{p}^{(t)}
\end{align}
(see Figure~\ref{fig:dynamics}).  The unique solution for $\mb{p}^{(t)}$ is given by\footnote{ The form chosen for
$\mb{\Gamma}$ in Equation~\eqref{eq:dynamics}, coupled with the satisfaction of detailed balance and ergodicity introduced in section \ref{sec:detailed balance},
guarantees that there
is a unique eigenvector $\mb{p}^{(\infty)}$ of $\mb{\Gamma}$ with
eigenvalue zero, and that all other eigenvalues of $\mb{\Gamma}$ are real and negative.}
\begin{align}
\mb{p}^{(t)} = \exp\left(\mb{\Gamma}t\right)\mb{p}^{(0)},
\end{align}
where $\exp\left(\mb{\Gamma}t\right)$ is a matrix exponential.

\section{Detailed Balance}
\label{sec:detailed balance}
The second core concept is detailed balance,
\begin{align}
\label{eq:detailed_balance}
\Gamma_{ji}\ p^{(\infty)}_i\left(\theta\right) = \Gamma_{ij}\ p^{(\infty)}_j\left(\theta\right)
,
\end{align}
which states that at equilibrium the probability flow from state $i$ into state $j$ equals the probability flow from $j$ into $i$.  When satisfied, detailed balance guarantees that the distribution $\mb p^{(\infty)}\left(\theta\right)$ is a fixed point of the dynamics.  Sampling in most Monte Carlo methods is performed by choosing $\mb \Gamma$ consistent with Equation \ref{eq:detailed_balance} (and the added requirement of ergodicity), then stochastically running the dynamics of Equation \ref{eq:mastereqn}.
Note that there is no need to restrict the dynamics defined by $\mb{\Gamma}$ to those of any real physical process, such as diffusion.

Equation \ref{eq:detailed_balance} can be written in terms of the model's energy function $\mb E\left( \theta \right)$ by substituting in Equation~\ref{eqn:p infinity} for $\mb p^{(\infty)}\left(\theta\right)$:
\begin{align}
{\Gamma_{ji}}
\exp \left( -E_i\left( \theta \right) \right)
=
{\Gamma_{ij}}
\exp \left( -E_j\left( \theta \right) \right)
.
\end{align}
$\mb{\Gamma}$ is underconstrained by the above equation. Introducing the additional
constraint that $\Gamma$ be invariant to the addition of a constant to the energy function (as the
model distribution $\mb p^{(\infty)}\left(\theta\right)$ is), we choose the following form for the non-diagonal entries in $\mb{\Gamma}$
\begin{align}
\label{eqn:gamma symmetric}
\Gamma_{ij} &=
g_{ij} \exp \left[ \frac{1}{2} \left( E_j\left( \theta \right)-E_i\left(  \theta \right) \right)\right] \qquad \left(i \neq j\right) 
,
\end{align}
where the connectivity function
\begin{align}
\label{eqn:g_ij}
g_{ij} = g_{ji} = &
\left\{\begin{array}{ccc}
   0 &  & \mathrm{unconnected\ states} \\
   1 &  & \mathrm{connected\ states}
\end{array}\right.
\qquad \left(i \neq j\right)
\end{align}
determines which states are allowed to directly exchange probability with each other.  The non-zero $\mb \Gamma$ may also be sampled from a proposal distribution rather than set via a deterministic scheme, in which case $g_{ij}$ takes on the role of proposal distribution - see Section \ref{sampling connectivity}. 
$g_{ij}$ can be set such that $\mb \Gamma$ is {\em extremely} sparse (see Section~\ref{sec:tractability}).  Theoretically, to guarantee convergence to the model distribution, the non-zero elements of $\mb{\Gamma}$ must be chosen such that, given sufficient time, probability can flow between any pair of states (ergodicity).
\section{Objective Function}
Maximum likelihood parameter estimation involves maximizing the likelihood of some observations $\mathcal{D}$ under a model, or equivalently minimizing the KL divergence between the data distribution $\mb p^{(0)}$ and model distribution $\mb p^{(\infty)}$,
\begin{align}
\hat{\theta}_{\mathrm{ML}} &= \argmin_\theta D_{KL}\left(
\mb{p^{(0)}} ||\mb{p^{(\infty)}}
\left(\theta\right)\right)
\end{align}
Rather than running the dynamics for infinite time, we propose to minimize the KL divergence after running the dynamics for an infinitesimal time $\epsilon$,
\begin{align}
\hat{\theta}_{\mathrm{MPF}} &= \argmin_\theta K\left( \theta \right) \\
K\left( \theta \right) &=
D_{KL}\left(
\mb{p^{(0)}} ||\mb{p^{(\epsilon)}}
\left(\theta\right)\right)
\label{eq:KL_unexpanded}
.
\end{align}
For small $\epsilon$, $D_{KL}\left(
\mb{p^{(0)}} ||\mb{p^{(\epsilon)}}
\left(\theta\right)\right)$ can be approximated by a first order Taylor expansion,
\begin{align}
K\left( \theta \right) & \approx D_{KL}\left(
\mb{p^{(0)}} ||\mb{p^{(t)}}
\left(\theta\right)\right)\Big |_{t=0}
\nonumber \\ & \qquad
\qquad + \epsilon \frac
{\partial D_{KL}\left(
\mb{p^{(0)}} ||\mb{p^{(t)}}
\left(\theta\right)\right)}
{\partial t}\Big |_{t=0}
.
\end{align}
Further algebra (see Appendix \ref{app:MPF Taylor}) reduces $K\left( \theta \right)$ to a measure of the flow of probability, at time $t = 0$ under the dynamics, out of data states $j \in \mathcal{D}$ into non-data states $i\notin \mathrm{\mathcal{D}}$,
\begin{align}
\label{eq:Kfinal}
K\left( \theta \right) &= \frac{\epsilon}{|\mathcal{D}|}\sum_{i\notin \mathrm{\mathcal{D}}}\sum_{j\in \mathrm{\mathcal{D}}} \Gamma_{ij} \\
& = \frac{\epsilon}{|\mathcal{D}|} \sum_{j\in \mathrm{\mathcal{D}}
} \sum_{
  i\notin \mathrm{\mathcal{D}}
}
g_{ij} \exp \left[ \frac{1}{2} \left( E_j\left( \theta \right)-E_i\left(  \theta \right) \right)\right]
\label{eq:Kfinal_expanded}
\end{align}
with gradient
\begin{align}
\label{eq:Kgrad}
\pd{K\left( \theta \right) }{\theta}
&= \frac{\epsilon}{|\mathcal{D}|} \sum_{j\in \mathrm{\mathcal{D}}
} \sum_{
    i\notin \mathrm{\mathcal{D}}
}
\left[ \pd{E_j\left( \theta \right)}{\theta}-\pd{E_i\left(  \theta \right)}{\theta} \right]
\nonumber \\ & \qquad
g_{ij} \exp \left[ \frac{1}{2} \left( E_j\left( \theta \right)-E_i\left(  \theta \right) \right)\right]
,
\end{align}
where $|\mathcal{D}|$ is the number of observed data points.
Note that Equations~\eqref{eq:Kfinal} and~\eqref{eq:Kgrad} do not depend on the partition function $Z\left( \theta \right)$ or its derivatives.

$K\left( \theta \right)$ is uniquely zero when $\mb{p}^{(0)}$ and $\mb{p}^{(\infty)}\left( \theta \right)$ are equal.  This implies consistency, in that if the data comes from the model class, in the limit of infinite data $K\left( \theta \right)$ will be minimized by exactly the true $\theta$.
In addition, $K\left( \theta \right)$ is convex for all models $\mb p^{(\infty)}\left( \theta \right)$ in the exponential family - that is, models whose energy functions ${\mb E}\left( \theta \right)$ are linear in their parameters $\theta$ \cite{macke} (see Appendix \ref{app:convex}).  The MPF objective additionally provides an upper bound on the log likelihood of the data if the first non-zero eigenvalue of $\mb \Gamma$ is known (see Appendix \ref{app:bound}).

\section{Tractability}
\label{sec:tractability}
The dimensionality of the vector $\mb{p}^{(0)}$ is typically huge, as is that of
$\mb{\Gamma}$ ({\it{e.g.}}, $2^d$ and $2^d \times 2^d$, respectively, for a $d$-bit binary
system).  Na\"ively, this would seem to prohibit evaluation and minimization of
the objective function.  Fortunately, we need only visit those columns of $\Gamma_{ij}$ corresponding to data states, $j \in \mathcal D$.  Additionally, $g_{ij}$ can be populated so as to connect each state $j$ to only a small fixed number of additional states $i$.
The cost in both memory and time to evaluate the objective function is thus $\mathcal{O}(|\mathcal{D}|)$, and does not depend
on the number of system states, only on the (much smaller) number of observed data points.

\section{Choosing the Connectivity Function $\mb g$}

Qualitatively, the most informative states to connect data states to are those that are most probable under the model.  In discrete state spaces, nearest neighbor connectivity schemes for $g_{ji}$ work extremely well (eg Equation \ref{eqn:g_ij_ising} below). This is because, as learning converges, the states that are near data states become the states that are probable under the model.

\section{Continuous State Spaces}\label{sec:cont_systems}
Although we have motivated this technique using systems with a large, but
finite, number of states, it generalizes to
continuous state spaces.  $\Gamma_{ji}$, $g_{ji}$, and $p^{(t)}_i$ become continuous functions $\Gamma\left( \mb x_j, \mb x_i\right)$, $g\left( \mb x_j, \mb x_i\right)$, and $p^{(t)}\left( \mb x_i\right)$.
$\Gamma\left( \mb x_j, \mb x_i\right)$ can be populated stochastically and extremely sparsely, preserving the $\mathcal{O}(|\mathcal{D}|)$ cost.

In continuous state spaces, the estimated parameters are much more sensitive to the choice of $g\left( \mb x_j, \mb x_i\right)$.  Practically, we have implemented MPF in continuous state spaces using the persistent particle extensions in Section \ref{persistent mpf}, and Hamiltonian Monte Carlo (HMC) to sample the connected states.

\section{Connection to Other Learning Techniques}\label{sec:connections}

\subsection{Contrastive Divergence}\label{sec:contrastive_divergence}
The contrastive divergence update rule (introduced in Section \ref{sec CD intro}) can be written in the form
\begin{align}
\label{eq:CDstep}
\Delta \theta_{CD} & \propto
-\sum_{j\in \mathrm{\mathcal{D}}
} \sum_{
   i\notin \mathrm{\mathcal{D}}
}
\left[ \pd{E_j\left( \theta \right)}{\theta}-\pd{E_i\left(  \theta \right)}{\theta} \right]
T_{ij}
,
\end{align}
where $T_{ij}$ is the probability of transitioning from state $j$ to state $i$ in a single Markov chain Monte Carlo step (or $k$ steps for CD-$k$).  Equation \ref{eq:CDstep} has obvious similarities to the MPF learning gradient in Equation \ref{eq:Kgrad}.  Thus, steepest gradient descent under MPF resembles CD updates, but with the MCMC sampling/rejection step $T_{ij}$ replaced by a weighting factor $g_{ij} \exp \left[ \frac{1}{2} \left( E_j\left( \theta \right)-E_i\left(  \theta \right) \right)\right]$.

Note that this difference in form provides MPF with a well-defined objective function. One important consequence of the existence of an objective function is that MPF can readily utilize general purpose, off-the-shelf optimization packages for gradient descent, which would have to be tailored in some way to be applied to CD. This is part of what accounts for the dramatic difference in learning time between CD and MPF in some cases (see Figure \ref{fig:ising_compare}).

\subsection{Score Matching}\label{sec:score_matching}
For a continuous state space, MPF reduces to score matching (introduced in Section \ref{sec SM intro}) if the connectivity function $g\left( \mb x_j, \mb x_i \right)$ is set to connect all states within a small distance $r$ of each other,
\begin{align}
\label{eqn:g_ij SM}
g(\mb x_i,\mb x_j) = g(\mb x_j,\mb x_i) = &
\left\{\begin{array}{ccc}
  0 &  & d(\mb x_i,\mb x_j) > r \\
  1 &  & d(\mb x_i,\mb x_j) \leq r
\end{array}\right.
,
\end{align}
where $d(\mb x_i,\mb x_j)$ is the Euclidean distance between states $\mb x_i$ and $\mb x_j$. In the limit as $r$ goes to 0 (within an overall constant and scaling factor),
\begin{align}
\label{eq:score matching limit}
\lim_{r \rightarrow 0} K\left( \theta \right) & \sim K_{\mathrm{SM}}\left( \theta \right) \nonumber \\
& = \ \sum_{\mb x\in\mathcal{D}} \left[ \frac{1}{2}\nabla E(\mb x)\cdot \nabla E(\mb x) - \nabla^2 E(\mb x) \right ],
\end{align}
where $K_{\mathrm{SM}}\left( \theta \right)$ is the SM objective function.  The full derivation is presented in Appendix \ref{app:score_matching}.
Unlike SM, MPF is applicable to any parametric model, including discrete systems, and it does not require evaluating a third order derivative, which can result in unwieldy expressions.

%

\section{Sampling the Connectivity Function  $g_{ij}$}
\label{sampling connectivity}

The MPF learning scheme is  blind to regions in state space which are not directly connected via $g_{ij}$ to data states.  One way to more flexibly and thoroughly connect states is to treat $g_{ij}$ as the probability of a connection from state $j$ to state $i$, rather than as a binary indicator function.  In this case, $g_{ij}$ has the constraints required of a probability distribution,
\begin{align}
\sum_i g_{ij} & = 1 \\
g_{ij} & \geq 0
,
\label{eq gij}
\end{align}
as well as the added constraint that if $g_{ij} >0$ then $g_{ji} > 0$.  Given these constraints for $g_{ij}$, the following form can be chosen for the transition rates $\Gamma_{ij}$,
\begin{align}
\Gamma_{ij}\left(\theta\right) =
 & 
	\left\{\begin{array}{ccc}
		g_{ij} \left(\frac{g_{ji}}{g_{ij}}\right)^{\frac{1}{2}} \exp\left[\frac{1}{2}\left( E_j\left(\theta\right) - E_i\left(\theta\right) \right) \right] &  & i \neq j \\
		-\sum_{k\neq j} \Gamma_{kj}\left(\theta\right) &  & i=j
	\end{array}\right.
.
\label{eq Gammaij}
\end{align}
It can be seen by substitution that the form for $\Gamma_{ij}$ in Equation \ref{eq Gammaij} still satisfies detailed balance.  Additional motivations for this form are that $\Gamma_{ij}$ have a linear factor $g_{ij}$ so that a sum over $i$ can be approximated using samples from $g_{ij}$, and that the contribution not included in the linear factor be a function solely of the ratio $\frac{g_{ji}}{g_{ij}}$, so that any ($j$ independent) normalization term in $g_{ij}$ cancels out.

Using $\Gamma_{ij}$ from Equation \ref{eq Gammaij}, the MPF objective function becomes
\begin{align}
K_{MPF}\left( \theta; \mb g \right) = 	\frac{1}{\left|\mathcal D\right|}
		\sum_{j\in \mathcal{D}}\sum_{i\notin \mathcal{D}}
	g_{ij}
	\left(\frac{g_{ji}}{g_{ij}}\right)^{\frac{1}{2}}
	\exp\left[
		\frac{1}{2}\left(
			E_j\left(\theta\right) - E_i\left(\theta\right)
		\right)
	\right]
.
\label{eq kmpf nest}
\end{align}
This is identical to the original MPF objective function, except for the addition of a scaling term $\left(\frac{g_{ji}}{g_{ij}}\right)^{\frac{1}{2}}$ which compensates for the differences between the forward and backward connection probabilities $g_{ij}$ and $g_{ji}$.

Because $g_{ij}$ is a probability distribution, the inner sum in Equation \ref{eq kmpf nest} is an expectation over $g_{ij}$, and can be approximated by averaging over sample states $i$ drawn from the distribution $g_{ij}$.

\section{Persistent MPF}
\label{persistent mpf}

Recent work has shown that persistent particle techniques \cite{Tieleman2008} outperform other sample driven learning techniques.  In direct analogy to Persistent Contrastive Divergence (PCD), and using the sampled connectivity function $g_{ij}$ introduced in Section \ref{sampling connectivity}, MPF can be extended to perform learning with persistent particles.

Nearest neighbor schemes for setting the connectivity function $\mb g$ do not work nearly as well in continuous state spaces as in discrete state spaces, while Persistent MPF (PMPF) works quite well in continuous state spaces, so PMPF is particularly applicable to the continuous state space case. 

\subsection{Factoring $K_{MPF}$}

In order to modify MPF to work with persistent samples, we first take advantage of a restricted form for $g_{ij}$ to rewrite the MPF objective function.  If the proposed connectivity function $g_{ij}$ depends only on the destination state, $i$, and not the initial state, $j$, then the nested sums in Equation \ref{eq kmpf nest} can be factored apart.  For the case that $g_{ij}$ does not depend on $j$, we write it simply as $g_i$.  The MPF objective function $K_{MPF}$ becomes
\begin{align}
K_{MPF}\left( \theta; \mb g \right) & = 	\frac{1}{\left|\mathcal D\right|}
		\sum_{j\in \mathcal{D}}\sum_{i\notin \mathcal{D}}
	g_i
	\left(\frac{g_j}{g_i}\right)^{\frac{1}{2}}
	\exp\left[
		\frac{1}{2}\left(
			E_j\left(\theta\right) - E_i\left(\theta\right)
		\right)
	\right] \\
& = \left(
	\frac{1}{\left|\mathcal D\right|}
	\sum_{j\in \mathcal{D}}
	\exp\left[
		\frac{1}{2}\left(
			E_j\left(\theta\right) + \log g_j
		\right)
	\right]\right) \cdot \nonumber  \\ & \qquad  \qquad
      \left(
	\sum_{i\notin \mathcal{D}}
	g_i
	\exp\left[
		-\frac{1}{2}\left(
			 E_i\left(\theta\right) + \log g_i
		\right)
	\right]	
	\right)
	.
\label{eq kmpf}
\end{align}
The second sum is an expectation under $g_i$, and can be approximated by averaging over samples from $g_i$.

\subsection{Iterative Improvement of $g_{i}$} \label{sec it}

The most informative states to connect to for learning are those which are most probable under the model distribution.  
Therefore, it is useful for learning to make $g_{i}$ as similar to $p^{(\infty)}_{i}\left( \theta \right)$ as possible.  An effective learning procedure alternates between updating $g_{i}$ to resemble the current estimate of the model distribution $p^{(\infty)}_i\left(\hat\theta\right)$, and updating the estimated model parameters $\hat\theta$ using samples from a fixed connectivity function $g_{i}$.  Defining a sequence of estimated parameter vectors $\hat \theta^n$ and proposed connectivity distributions $g_{i}^n$, where $n$ indicates the learning iteration, this learning procedure becomes
\begin{enumerate}
\item Set $\hat\theta^0 = $ initial parameter guess
\item For $n \in \mathcal{Z}_+$ iterate
\begin{enumerate}
\item Set $g^{n}_{i} = p^{(\infty)}_i\left(\hat\theta^{n-1}\right) = \frac{\exp\left[-E_i\left(\hat\theta^{n-1}\right)\right]}{Z\left(\hat\theta^{n-1}\right)}$ 
					\label{g set}
\item Find $\hat\theta^{n}$ such that $K^n_{MPF}\left( \hat\theta^{n} \right) < K^n_{MPF}\left( \hat\theta^{n-1} \right)$
\end{enumerate}
\end{enumerate} 
The MPF objective function at learning step $n$, $K^n_{MPF}\left( \theta \right)$, is written using the proposal distribution $g_i^n$ set in step \ref{g set},
\begin{align}
K^n_{MPF}\left( \theta \right)
& = \left(
	\frac{1}{\left|\mathcal D\right|}
	\sum_{j\in \mathcal{D}}
	\exp\left[
		\frac{1}{2}\left(
			E_j\left(\theta\right) - E_j\left(\hat\theta^{n-1}\right)
		\right)
	\right]\right)\cdot \nonumber  \\ & \qquad \qquad
      \left(
	\sum_{i\notin \mathcal{D}}
	g^n_i
	\exp\left[
		-\frac{1}{2}\left(
			 E_i\left(\theta\right) - E_i\left(\hat\theta^{n-1}\right)
		\right)
	\right]	
	\right)
\label{eq kmpf eng}
\end{align}
(the normalization terms in $\log g_i$ cancel out between the two sums).  The expectation in the second sum is still evaluated using samples from $\mb g^n$.  Typically, the number of samples drawn from $\mb g^n$ will be the same as the number of observations, $\left| \mathcal D \right|$.

\subsection{Persistent Samples}

The procedure in Section \ref{sec it} will usually leave the proposal distribution at learning step $n$, $\mb g^{n}$, very similar to the proposal distribution from step $n-1$, $\mb g^{n-1}$.  Significant time can thus be saved when generating samples from $\mb g^{n}$ by initializing with samples from $\mb g^{n-1}$, and taking only a small number of sampling steps.


\subsection{Full Procedure for Persistent MPF}

Using PMPF in an $M$-dimensional continuous states space $\mathcal R^M$, the parameter estimation procedure is as given in the steps below.  $\mathcal S^n$ is the list of samples at learning step $n$.  $\left| \mathcal S^n \right|$ is the number of samples - typically it will be the same as the number of observations $\left| \mathcal D \right|$.
\begin{enumerate}
\item Set $\hat\theta^0 = $ initial parameter guess
\item Initialize samples $\mathcal S^0$ (eg from a Gaussian) 
\item For $n \in \mathcal{Z}_+$ iterate
\begin{enumerate}
\item Draw samples $\mathcal S^n$ from the distribution $p^{(\infty)}\left( \mb x; \hat\theta^{n-1} \right)$ via an MCMC sampler intialized at $\mathcal S^{n-1}$ (eg using Hamiltonian Monte Carlo)
\item Find $\hat\theta^{n}$ such that $K^n_{MPF}\left( \hat\theta^{n} \right) < K^n_{MPF}\left( \hat\theta^{n-1} \right)$ (eg via 10 steps of LBFGS gradient descent)
\end{enumerate}
\end{enumerate}
$K^n_{MPF}\left( \theta \right)$ is the MPF objective function at learning step $n$, and is written
\begin{align}
K^n_{MPF}\left( \theta \right)
= & \left(
	\frac{1}{\left|\mathcal D\right|}
	\sum_{\mb x \in \mathcal{D}}
	\exp\left[
		\frac{1}{2}\left(
			E\left(\mb x;  \theta\right) - E\left(\mb x; \hat\theta^{n-1}\right)
		\right)
	\right]\right) \cdot \nonumber  \\ & \qquad
      \left(
	\frac{1}{\left|\mathcal S^n\right|}
	\sum_{\mb x \in \mathcal{S}^n}
	\exp\left[
		-\frac{1}{2}\left(
			 E\left(\mb x; \theta\right) - E\left(\mb x; \hat\theta^{n-1}\right)
		\right)
	\right]	
	\right)
,
\label{eq PMPF cont}
\end{align}
with derivative
\begin{align}
\pd
	{K^n_{MPF}\left( \theta \right)}
	{\theta}
= & \frac{1}{2} \left(
	\frac{1}{\left|\mathcal D\right|}
	\sum_{\mb x \in \mathcal{D}}
	\exp\left[
		\frac{1}{2}\left(
			E\left(\mb x;  \theta\right) - E\left(\mb x; \hat\theta^{n-1}\right)
		\right)
	\right]   \pd{E\left(\mb x;  \theta\right)}{\theta}   \right) \cdot \nonumber  \\ & \qquad
      \left(
	\frac{1}{\left|\mathcal S^n\right|}
	\sum_{\mb x \in \mathcal{S}^n}
	\exp\left[
		-\frac{1}{2}\left(
			 E\left(\mb x; \theta\right) - E\left(\mb x; \hat\theta^{n-1}\right)
		\right)
	\right]	
	\right) \nonumber \\ &
-
\frac{1}{2} \left(
	\frac{1}{\left|\mathcal D\right|}
	\sum_{\mb x \in \mathcal{D}}
	\exp\left[
		\frac{1}{2}\left(
			E\left(\mb x;  \theta\right) - E\left(\mb x; \hat\theta^{n-1}\right)
		\right)
	\right]\right) \cdot \nonumber  \\ & \qquad
      \left(
	\frac{1}{\left|\mathcal S^n\right|}
	\sum_{\mb x \in \mathcal{S}^n}
	\exp\left[
		-\frac{1}{2}\left(
			 E\left(\mb x; \theta\right) - E\left(\mb x; \hat\theta^{n-1}\right)
		\right)
	\right]
	\pd{E\left(\mb x;  \theta\right)}{\theta}	
	\right)
.
\end{align}

\section{Summary}


We have presented a novel, general purpose framework, called minimum probability flow learning (MPF), for parameter estimation in probabilistic models that outperforms current techniques in both learning time and accuracy. MPF works for any parametric model without hidden state variables, including those over both continuous and discrete state space systems, and it avoids explicit calculation of the partition function by employing deterministic dynamics in place of the slow sampling required by many existing approaches.
Because MPF provides a simple and well-defined objective function, it can be minimized quickly using existing higher order gradient descent techniques.
Furthermore, the objective function is convex for models in the exponential family, ensuring that the global minimum can be found with gradient descent in these cases.
MPF was inspired by the minimum velocity approach developed by Movellan, and it reduces to that technique as well as to score matching and some forms of contrastive divergence for special cases of the dynamics.






\chapter{Minimum Probability Flow Experimental Results}
\label{chap MPF examples}

In this chapter, we demonstrate experimentally the effectiveness of the Minimum Probability Flow (MPF) learning technique presented in Chapter \ref{chap MPF}.  Matlab code implementing MPF for several of the cases presented in this chapter is available at \cite{MPFcode}.  Unless stated otherwise, minimization was performed using the L-BFGS implementation in minFunc \cite{schmidt}.  Material in this chapter is taken from \cite{Hillar2012,MPF_ICML,SohlDickstein2011a,SohlDickstein2009p8641}.

\section{Ising Model}

The Ising model \cite{ising25} has a long and storied history in physics \cite{RevModPhys.39.883} and machine learning \cite{Ackley85} and it has recently been found to be a surprisingly useful model for networks of neurons in the retina \cite{Schneidman_Nature_2006,Shlens_JN_2006}.  The ability to fit Ising models to the activity of large groups of simultaneously recorded neurons is of current interest given the increasing number of these types of data sets from the retina, cortex and other brain structures.

\subsection{Two Dimensional Ising Spin Glass}

We estimated parameters for an Ising model (sometimes referred to as a fully visible Boltzmann machine or an Ising spin glass) of the form
\begin{align}
p^{(\infty)}(\mb{x};\mb{J}) = \frac{1}{Z(\mb{J})}\exp\left[ -\mb{x}^{\mathrm{T}}\mb{J}\mb{x} \right],
\end{align}
where the coupling matrix $\mb{J}$ only had non-zero elements corresponding to
nearest-neighbor units in a two-dimensional square lattice, and bias terms along the diagonal. The training data $\mathcal{D}$ consisted of $20,000$ $d$-element iid binary samples $\mb x \in \{0,1\}^d$ generated via
Swendsen-Wang sampling~\cite{swendsen1987nonuniversal} from a spin glass with known coupling parameters. We used a square $10 \times 10$ lattice, $d=10^2$.
The non-diagonal nearest-neighbor elements of $\mb{J}$ were set using draws from a normal distribution with variance $\sigma^2 = 10$.
The diagonal (bias) elements of $\mb{J}$ were set in such a way that each column of $\mb{J}$ summed to 0, so that the expected unit activations were $0.5$.
The transition matrix $\mb{\Gamma}$ had $2^d \times 2^d$ elements, but for learning we populated it sparsely,
setting
\begin{align}
\label{eqn:g_ij_ising}
g_{ij} = g_{ji} &=
\left\{\begin{array}{ccc}
   1 &  & \mathrm{states\ }i,j\mathrm{\ differ\ by\ single\ bit\ flip} \\
   0 &  & \mathrm{otherwise}
\end{array}\right.
.
\end{align}
The full derivation of the MPF objective for the case of an Ising model can be found in Appendix \ref{app MPF Ising}.

Figure~\ref{fig:ising_compare} shows the mean square error in the estimated $\mb{J}$ and the mean square error in the corresponding pairwise correlations as a function of learning time for MPF and four competing approaches: mean field theory with TAP corrections \cite{Tanaka:1998p1984}, CD with both one and ten sampling steps per iteration, and pseudolikelihood. Parameter estimation in Minimum Probability Flow and Pseudolikelihood was performed by applying an off the shelf L-BFGS (quasi-Newton gradient descent) implementation \cite{schmidt} to their objective functions evaluated over the full training dataset $\mathcal D$.   CD was trained via stochastic gradient descent, using minibatches of size 100.
The learning rate was annealed in a linear fashion from 3.0 to 0.1 to accelerate convergence.  
Mean field theory requires the computation of the inverse of the magnetic susceptibility matrix, which, for strong correlations, was often singular. A regularized pseudoinverse was used in the following manner:
\begin{align}
A = (\chi^T\chi + \lambda I)^+\chi^T,
\end{align}
where $I$ is the identity matrix, $M^+$ denotes the Moore-Penrose pseudoinverse of a matrix $M$, $\chi$ is the magnetic susceptibility $\chi_{ij} = \avg{x_ix_j}-\avg{x_i}\avg{x_j}$, and $\lambda$ is a regularizing parameter.  This technique is known as stochastic robust approximation \cite{boyd2004convex}.

Using MPF, learning took approximately 60 seconds, compared to roughly 800 seconds for pseudolikelihood and upwards of $20,000$ seconds for 1-step and 10-step CD. Note that given sufficient training samples, MPF would converge exactly to the right answer, as learning in the Ising model is convex (see Appendix \ref{app:convex}), and has its global minimum at the true solution. Table~\ref{table:error_table} shows the relative performance at convergence in terms of mean square error in recovered weights, mean square error in the resulting model's correlation function, and convergence time. MPF was dramatically faster to converge than any of the other models tested, with the exception of MFT+TAP, which failed to find reasonable parameters. MPF fit the model to the data substantially better than any of the other models.
\begin{figure}
\begin{center}
\begin{tabular}{cc}
\begin{tabular}{c}\includegraphics[width=0.4\linewidth]{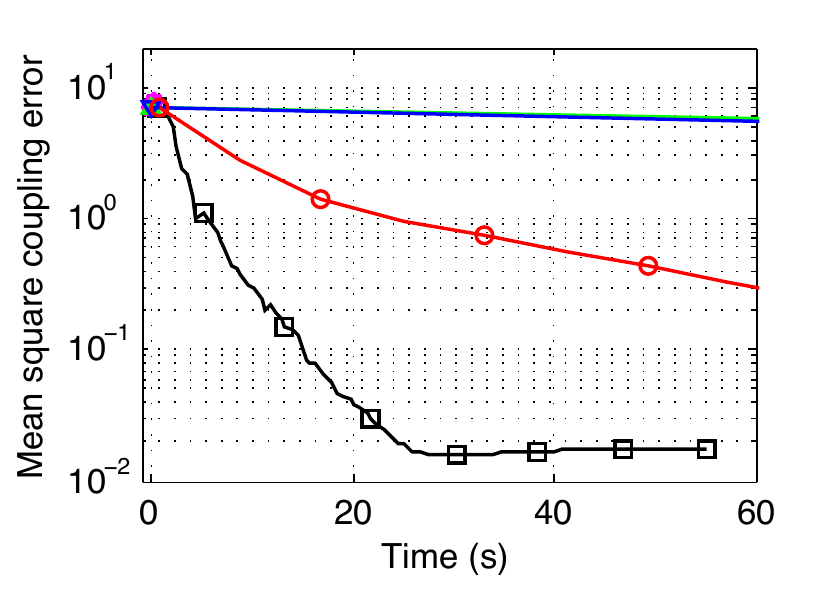}\\(a)\end{tabular}
&
\begin{tabular}{c}\includegraphics[width=0.4\linewidth]{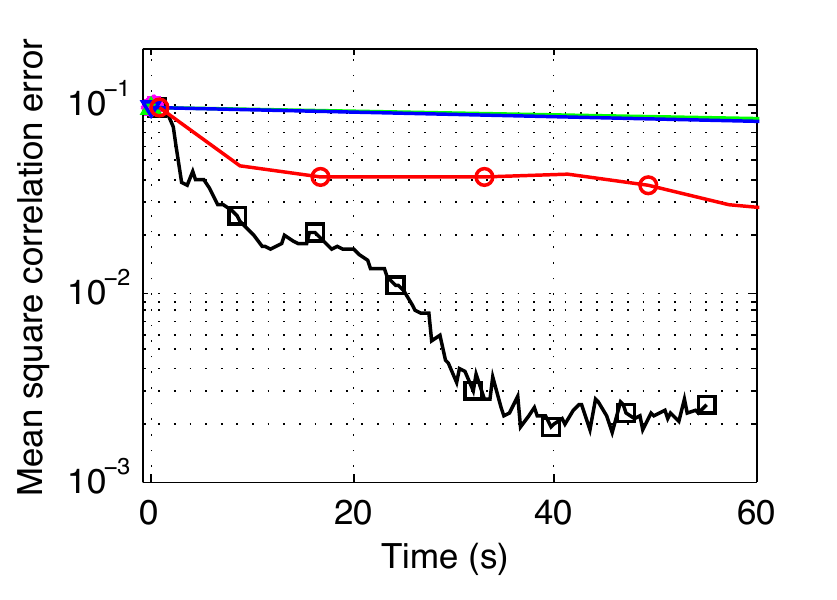}\\(d)\end{tabular}
\\
\begin{tabular}{c}\includegraphics[width=0.4\linewidth]{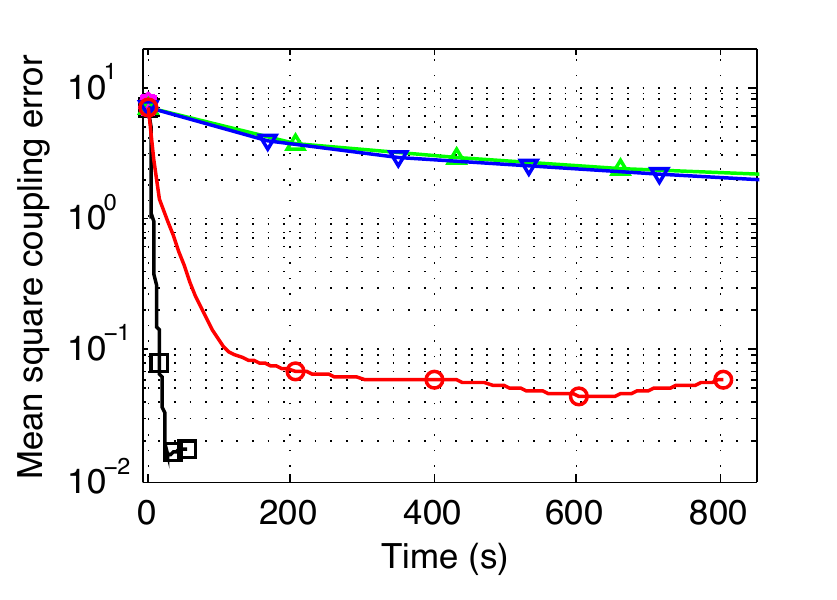}\\(b)\end{tabular}
&
\begin{tabular}{c}\includegraphics[width=0.4\linewidth]{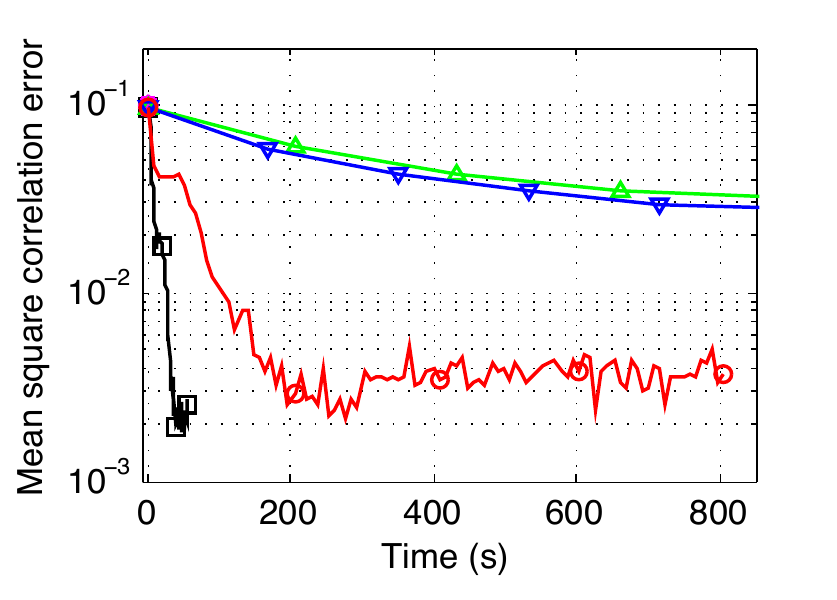}\\(e)\end{tabular}
\\
\begin{tabular}{c}\includegraphics[width=0.4\linewidth]{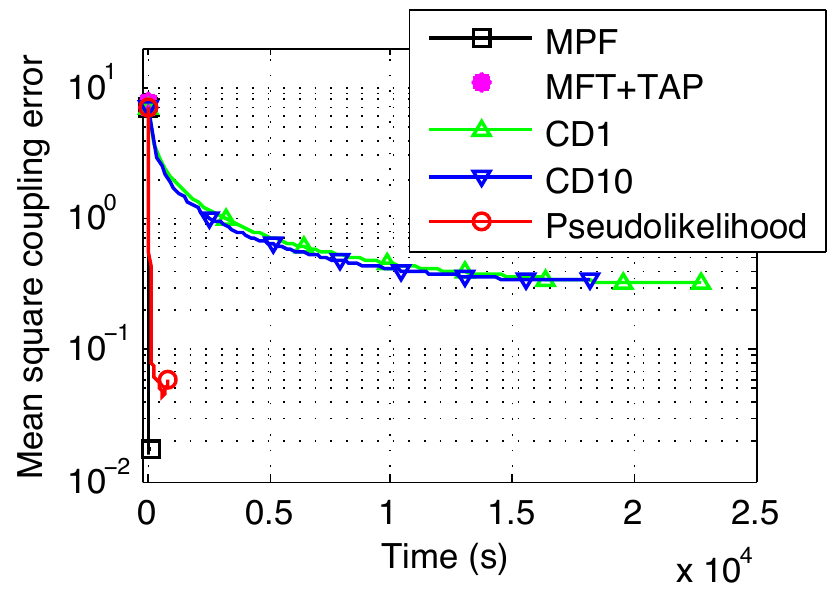}\\(c)\end{tabular}
&
\begin{tabular}{c}\includegraphics[width=0.4\linewidth]{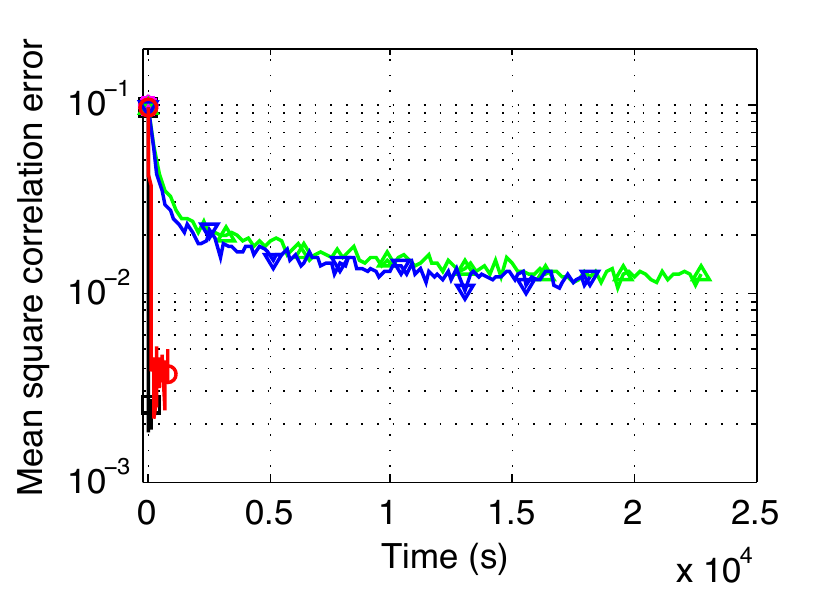}\\(f)\end{tabular}
\end{tabular}
\end{center}
\caption{A demonstration of Minimum Probability Flow (MPF) outperforming existing techniques for parameter recovery in an Ising spin glass.
{\textbf{(a)}} Time evolution of the mean square error in the coupling strengths for 5 methods for the first 60 seconds of learning.  Note that mean field theory with second order corrections (MFT+TAP) actually increases the error above random parameter assignments in this case. {\textbf{(b)}} Mean square error in the coupling strengths for the first 800 seconds of learning. {\textbf{(c)}} Mean square error in coupling strengths for the entire learning period. {\textbf{(d)}}--{\textbf{(f)}} Mean square error in pairwise correlations for the first 60 seconds of learning, the first 800 seconds of learning, and the entire learning period, respectively. In every comparison above MPF finds a better fit, and for all cases but MFT+TAP does so in a shorter time (see Table~\ref{table:error_table}).}
\label{fig:ising_compare}
\end{figure}
\begin{table}[t]
\caption{Mean square error in recovered coupling strengths ($\epsilon_{J}$), mean square error in pairwise correlations ($\epsilon_{\mathrm{corr}}$) and learning time for MPF versus mean field theory with TAP correction (MFT+TAP), 1-step and 10-step contrastive divergence (CD-1 and CD-10), and pseudolikelihood (PL).}
\label{table:error_table}
\vskip 0.15in
\begin{center}
\begin{small}
\begin{sc}
\begin{tabular}{lrrr}
\hline
Technique & $\epsilon_J$ & $\epsilon_{\mathrm{corr}}$& Time (s) \\
\hline
MPF    &\ \ 0.0172&\ \ 0.0025& $\sim$60 \\
MFT+TAP &\ \ 7.7704&\ \ 0.0983& \ \ 0.1\\
CD-1    &\ \ 0.3196&\ \ 0.0127& $\sim$20000 \\
CD-10    &\ \ 0.3341&\ \ 0.0123& $\sim$20000\\
PL    &\ \ 0.0582&\ \ 0.0036& $\sim$800\\
\end{tabular}
\end{sc}
\end{small}
\end{center}
\vskip -0.1in
\end{table}

\begin{figure}
\center{
\parbox[c]{\linewidth}{
\center{
\includegraphics[width= 0.5 \linewidth]{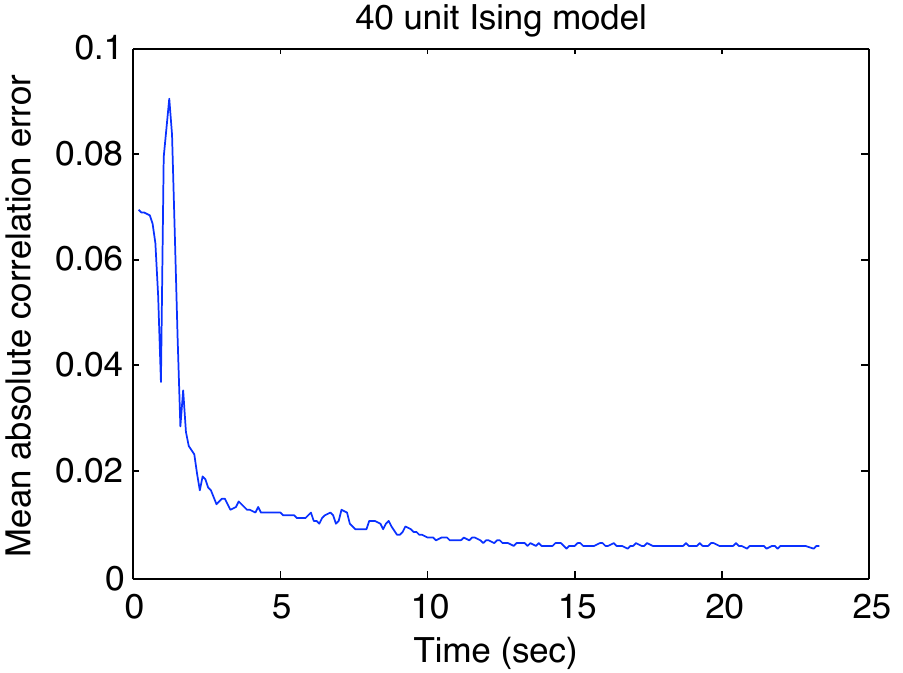}
}
}
}
\caption{
A demonstration of rapid fitting of a fully connected Ising model by minimum probability flow learning.  The mean absolute error in the learned model's correlation matrix is shown as a functions of learning time for a 40 unit fully connected Ising model.  Convergence is reached in about $15$ seconds for $20,000$ samples.
}
\label{fig:ising time}
\end{figure}

\subsection{Fully Connected Ising Model Comparison}

%

In order to allow an additional comparison to earlier work, we recovered the coupling parameters for the 40 unit, fully connected Ising model used in the 2008 paper ``Faster solutions of the inverse pairwise Ising problem" \cite{Broderick:2007p2761}.  
Figure~\ref{fig:ising time} shows the average error in predicted
correlations as a function of learning time for 20,000 samples.  The final absolute correlation error is 0.0058.  The $J_{ij}$ used were graciously provided by Broderick and coauthors, and were identical to those used for synthetic data generation in their paper \cite{Broderick:2007p2761}.  Training was performed on 20,000 samples so as to match the number of samples used in section III.A. of Broderick et al.  On an 8 core 2.33 GHz Intel Xeon,
the learning converges in about $15$ seconds. Broderick et al. perform a similar learning task on a 100-CPU grid computing
cluster, with a convergence time of approximately $200$ seconds.

\section{Deep Belief Network}
\begin{figure}
\center{
\parbox[c]{\linewidth}{
\center{
\begin{tabular}{cc}
\parbox[c]{0.35\linewidth}{
{\em\textbf{(a)}} \includegraphics[width=0.7\linewidth]{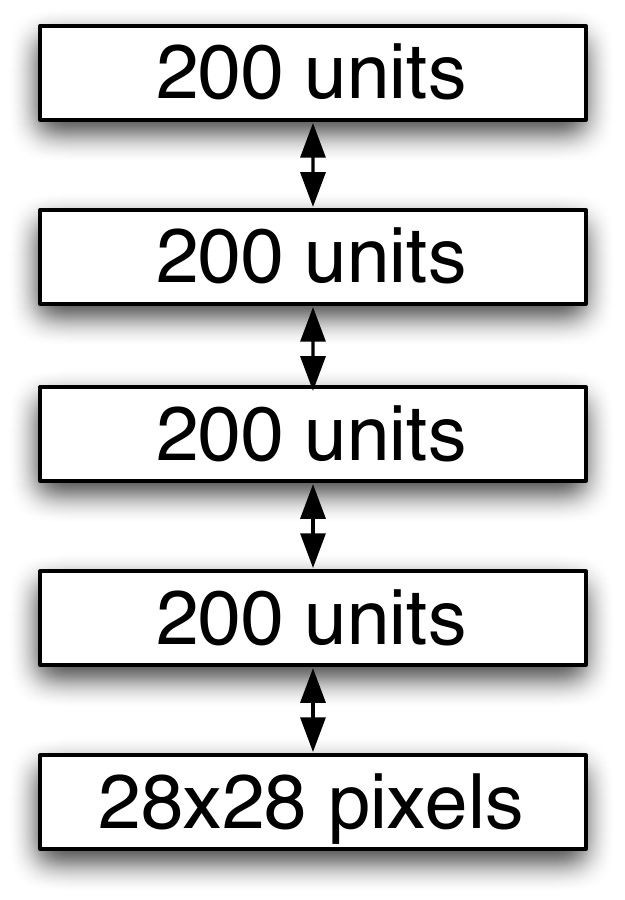}
}
&
\parbox[c]{0.45\linewidth}{
\begin{tabular}{c}
{\em\textbf{(b)}} \includegraphics[width=0.9\linewidth]{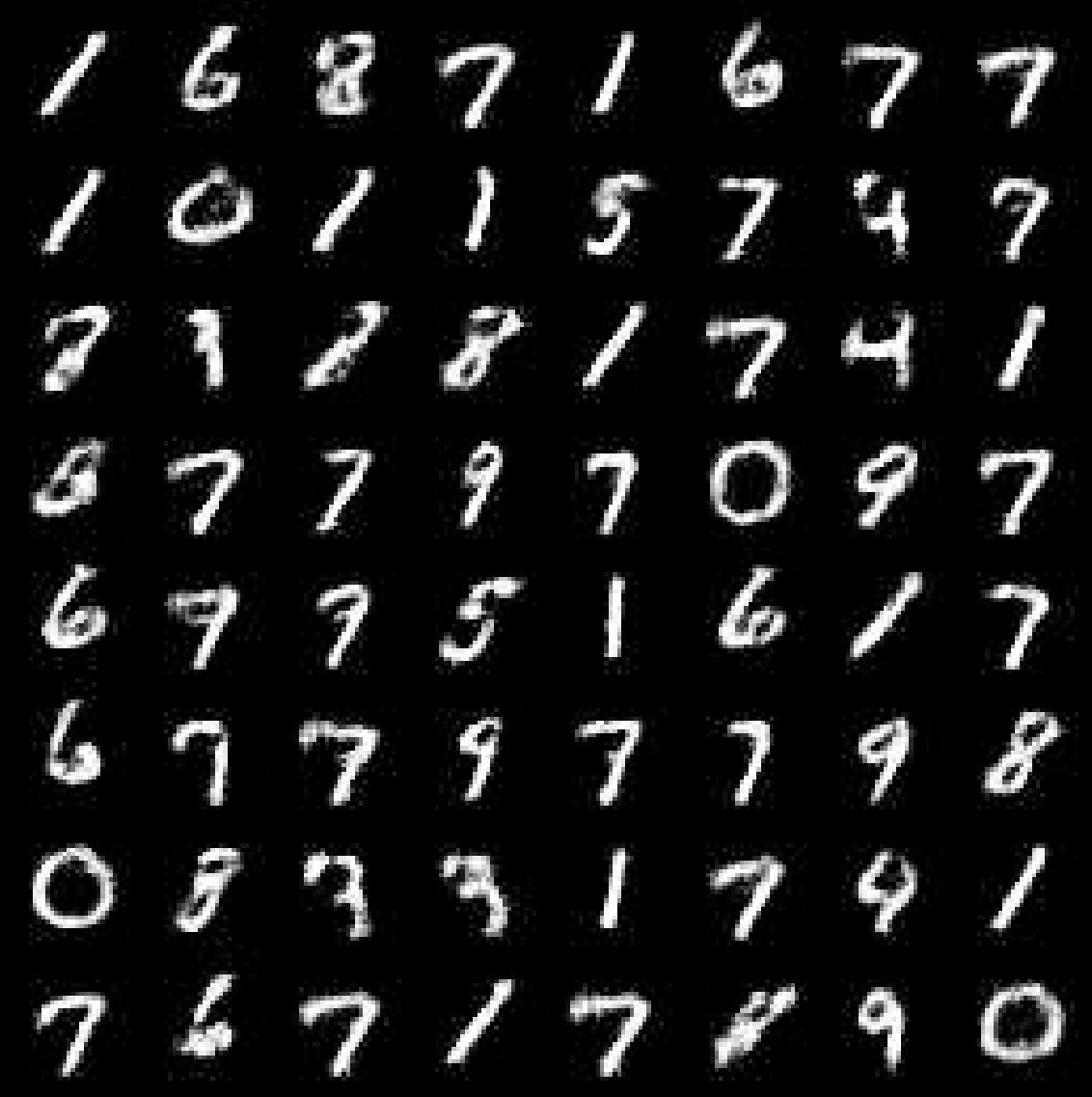} \\
{\em\textbf{(c)}} \includegraphics[width=0.9\linewidth]{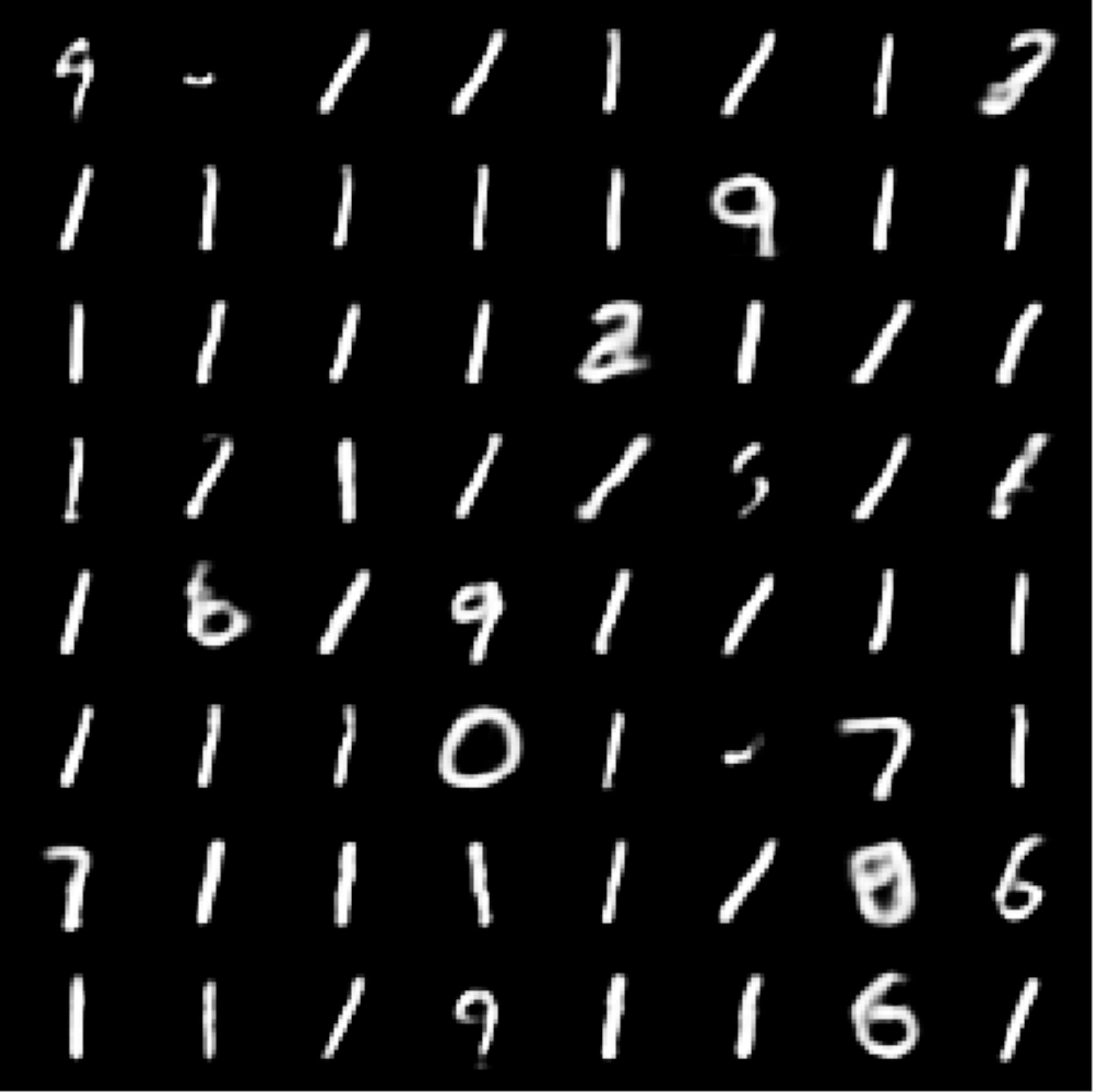}
\end{tabular}
}
\end{tabular}
}
}
}
\caption{
A deep belief network trained using minimum probability flow learning (MPF). \emph{(a)} A four layer deep belief network was trained on the MNIST postal hand written digits dataset by MPF and single step contrastive divergence (CD). \emph{(b)} Samples from the deep belief network after training via MPF.  A reasonable probabilistic model for handwritten digits has been learned.  \emph{(c)}  Samples after training via CD.  The uneven distribution of digit occurrences suggests that CD-1 has learned a less representative model than MPF.
}
\label{fig:dbn}
\end{figure}
As a demonstration of learning on a more complex discrete valued model, we trained a 4 layer deep belief network (DBN) \cite{DBN} on MNIST handwritten digits.  A DBN consists of stacked restricted Boltzmann machines (RBMs), such that the hidden layer of one RBM forms the visible layer of the next.  Each RBM has the form
\begin{align}
&p^{(\infty)}(\mb{x}_{\mathrm{vis}}, \mb{x}_{\mathrm{hid}};\mb{W}) =
\frac{
   \exp\left[ \mb x_{\mathrm{hid}}^T \mb W \mb x_{\mathrm{vis}} \right]
}{
   Z(\mb{W})
}
,
\\
&p^{(\infty)}(\mb{x}_{\mathrm{vis}};\mb{W}) = \frac{
   \exp\left[
\sum_{k} \log
\left(
1 +
\exp\left[
  \mb W_k \mb x_{\mathrm{vis}}
\right]
\right)
\right]
}{
   Z(\mb{W})
}
.
\end{align}
Sampling-free application of MPF requires analytically marginalizing over the hidden units.  RBMs were trained in sequence, starting at the bottom layer, on 10,000 samples from the MNIST postal hand written digits data set. As in the Ising case, the transition matrix $\mb{\Gamma}$ was populated so as to connect every state to all states that differed by only a single bit flip (Equation \ref{eqn:g_ij_ising}).  The full derivation of the MPF objective for the case of an RBM can be found in Appendix \ref{app MPF RBM}.  Training was performed by both MPF and single step CD (note that CD turns into full ML learning as the number of steps is increased, and that many step CD would have produced a superior, more computationally expensive, answer).

Samples were generated by Gibbs sampling from the top layer RBM, then propagating each sample back down to the pixel layer by way of the conditional distribution $p^{(\infty)}(\mb{x}_{\mathrm{vis}} | \mb{x}_{\mathrm{hid}};\mb{W}^k)$ for each of the intermediary RBMs, where $k$ indexes the layer in the stack.   $1,000$ sampling steps were taken between each sample.  As shown in Figure~\ref{fig:dbn}, MPF learned a good model of handwritten digits.

\section{Independent Component Analysis}
\begin{figure}
\center{
\parbox[b]{0.45\linewidth}{
\center{
\begin{tabular}{c}
\includegraphics[width= 0.9\linewidth]{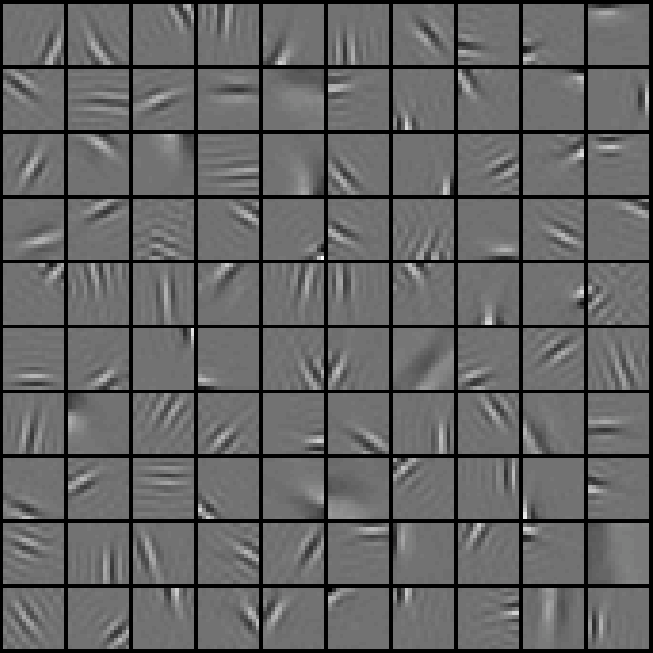} \\
{\em\textbf(a)}
\end{tabular}
}
}
\parbox[b]{0.45\linewidth}{
\center{
\begin{tabular}{c}
\includegraphics[width= 0.9\linewidth]{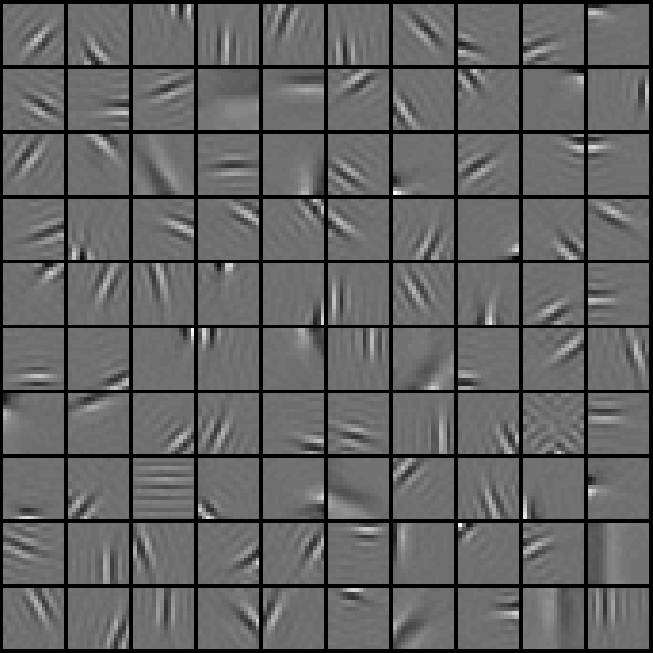} \\
{\em\textbf(b)}
\end{tabular}
}
}
}
\caption{
A continuous state space model fit using minimum probability flow learning (MPF).  Learned $10 \times 10$ pixel independent component analysis receptive fields $\mb J$ trained on natural image patches via \emph{(a)} MPF and \emph{(b)} maximum likelihood learning (ML).  The average log likelihood of the model found by MPF ($-120.61\ \mathrm{nats}$) was nearly identical to that found by ML ($-120.33\ \mathrm{nats}$), consistent with the visual similarity of the receptive fields.
}
\label{fig:ICA}
\end{figure}
As a demonstration of parameter estimation in continuous state space probabilistic models, we trained the receptive fields $\mb J \in R^{K\times K}$ of a $K$ dimensional independent component analysis (ICA) \cite{ICA} model with a Laplace prior,
\begin{align}
p^{(\infty)}\left(\mb x; \mb J \right)
&=
\frac{
e^{
-\sum_k \left|
\mb J_{k} \mb x
\right|
}
}{
2^K
\left| \mb J^{-1} \right|
}
,
\end{align}
on $100,000$ $10 \times 10$ whitened natural image patches from the van Hateren database \cite{hateren_schaaf_1998}.  Since the log likelihood and its gradient can be calculated analytically for ICA, we solved for $\mb J$ via both maximum likelihood learning and MPF, and compared the resulting log likelihoods.  Both training techniques were initialized with identical Gaussian noise, and trained on the same data, which accounts for the similarity of individual receptive fields found by the two algorithms.  The average log likelihood of the model after parameter estimation via MPF was $-120.61\ \mathrm{nats}$, while the average log likelihood after estimation via maximum likelihood was $-120.33\ \mathrm{nats}$.  The receptive fields resulting from training under both techniques are shown in Figure \ref{fig:ICA}.
MPF parameter estimation was performed using the Persistent MPF (PMPF) algorithm described in Section \ref{persistent mpf}, using Hamiltonian Monte Carlo (HMC) to sample from the connectivity function $g\left( \mb x_j, \mb x_i\right)$.

\section{Memory Storage in a Hopfield Network}

In 1982, motivated by the Ising spin glass model from statistical physics \cite{ising25,Little1974}, Hopfield introduced an auto-associative neural-network for the storage and retrieval of binary patterns \cite{hopfield1982}.  Even today, this model and its various extensions \cite{cohen1983absolute,hinton1986learning} provide a plausible mechanism for memory formation in the brain.  However, existing techniques for training Hopfield networks suffer either from limited pattern capacity or excessive training time, and they exhibit poor performance when trained on unlabeled, corrupted memories.

In this section we show that MPF provides a tractable and neurally-plausible algorithm for the optimal storage of patterns in a Hopfield network, and we provide a proof that the capacity of such a network is at least one pattern per neuron.  When compared with standard techniques for Hopfield pattern storage, MPF is shown to be superior in efficiency and generalization.  Another finding is that MPF can store many patterns in a Hopfield network from highly corrupted (unlabeled) samples of them.  This discovery is also corroborated visually by the storage of $64 \times 64$ binary images of human fingerprints from highly corrupted versions, as explained in Fig.~\ref{fig7a}.

\begin{figure}[!t]
\centering
\includegraphics[width=3.4in]{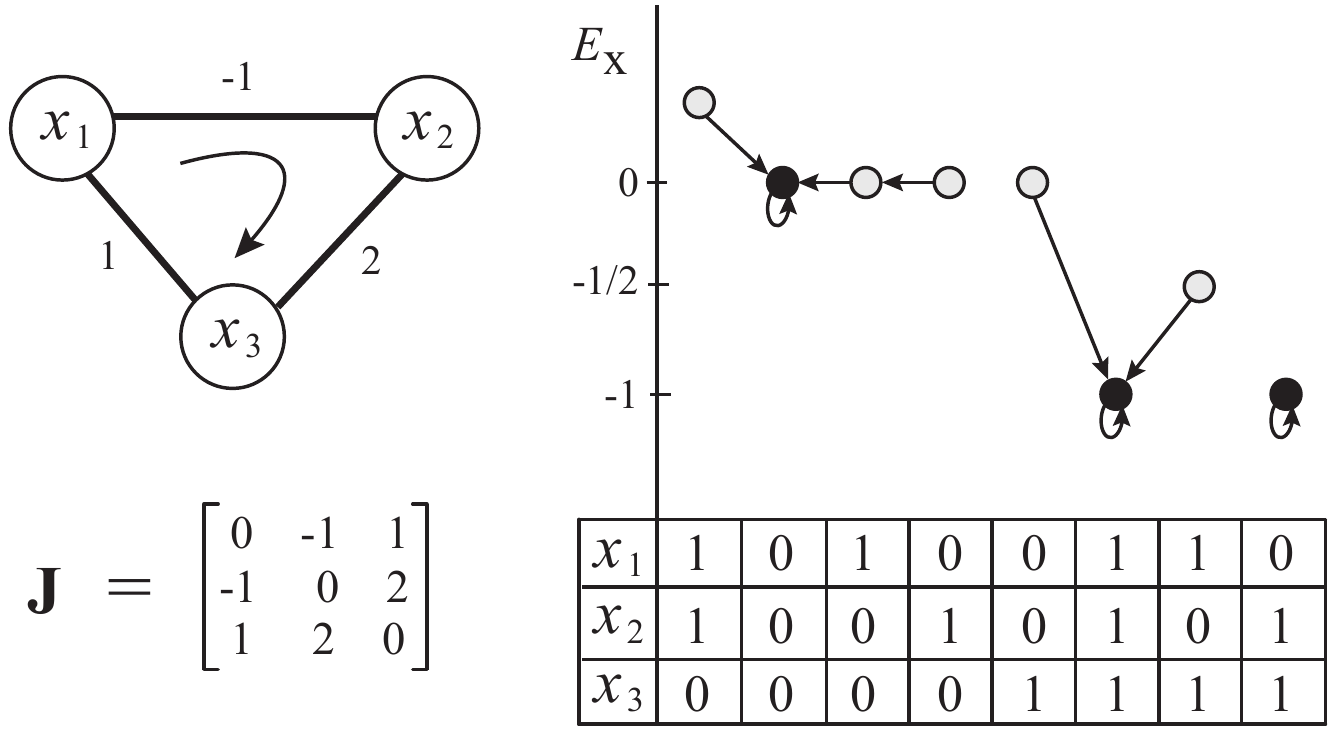}
\caption{\textbf{Example Hopfield Network.}  The figure above displays a $3$-node Hopfield network with weight matrix $\mathbf{J}$ and zero threshold vector.  Each binary state vector  $\mathbf{x} = (x_1,x_2,x_3)^{\top}$ has energy $E_{\mathbf{x}}$ as labeled on the $y$-axis of the diagram on the right.  Arrows between states represent one iteration of the network dynamics; i.e., $x_1$, $x_2$, and $x_3$ are updated by (\ref{Hopdynamics}) in the order indicated by the clockwise arrow in the graph on the left.  The resulting fixed states of the network are indicated by filled circles.}
\vspace{-.25cm}
\label{fighopnet}
\end{figure}

\begin{figure}[!t]
\centering
\includegraphics[width=3.4in]{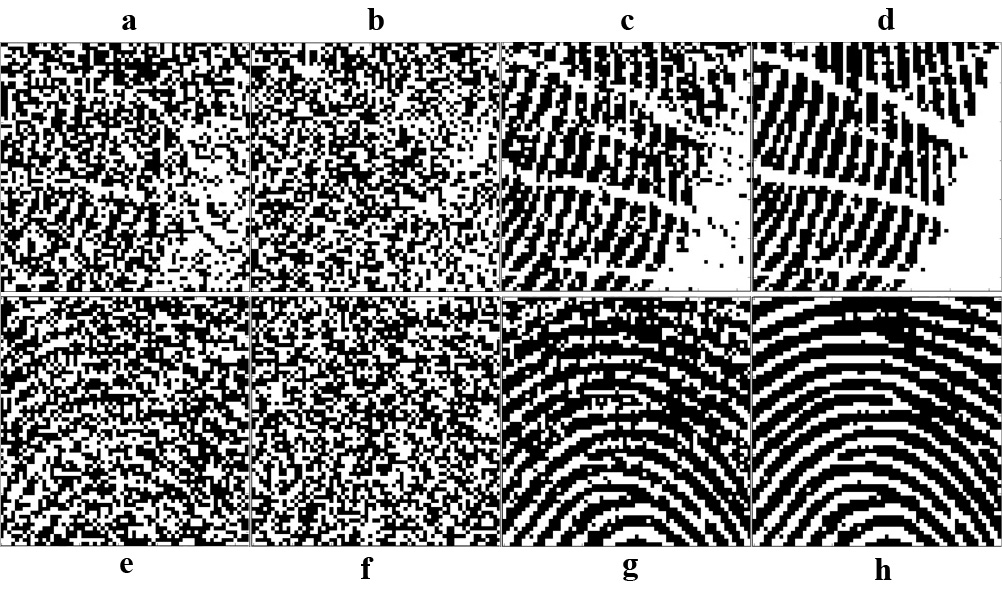}
\caption{\textbf{Learning memories from corrupted samples.} We stored $80$ fingerprints ($64 \times 64$ binary images) in a Hopfield network with $n = 64^2 = 4096$ nodes by minimizing the MPF objective (\ref{MPFobjective}) over a large set of randomly generated (and unlabeled) ``noisy" versions (each training pattern had a random subset of 1228 of its bits flipped; e.g., a,e).  After training, all $80$ original fingerprints were stored as fixed-points of the network.  \textbf{a.} Sample fingerprint with 30\% corruption used for training. \textbf{b.} Sample fingerprint with 40\% corruption. \textbf{c.} State of the network after one update of the dynamics initialized at b.  \textbf{d.} Converged network dynamics equal to original fingerprint. \textbf{e-h.} As in {a-d}, but for a different fingerprint.}
\label{fig7a}
\end{figure}

\subsection{Background} A \textit{Hopfield network} $\mathcal{H} = (\mb J,\theta)$ on $n$ nodes  $\{1,\ldots,n\}$ consists of a symmetric \textit{weight matrix} $\mb J = \mb J^{\top} \in \mathbb R^{n \times n}$  with zero diagonal and a \textit{threshold vector} $\theta = (\theta_1,\ldots,\theta_n)^{\top} \in \mathbb R^n$.  The possible \textit{states} of the network are all length $n$ binary strings $\{0,1\}^n$, which we represent as binary column vectors $\mb x = (x_1,\ldots,x_n)^{\top}$, each $x_i  \in \{0,1\}$ indicating the state $x_i$ of node $i$.  Given any state $\mb x = (x_1,\ldots,x_n)^{\top}$, an (asynchronous) \textit{dynamical update} of $\mb x$ consists of replacing $x_i$ in $\mb x$ (in consecutive order starting with $i = 1$; see Fig \ref{fighopnet}) with the value
\begin{equation}\label{Hopdynamics}
x_i = H(\mb J_i \mb x - \theta_i).
\end{equation}
Here, $\mb J_i$ is the $i$th row of $\mb J$ and $H$ is the \textit{Heaviside function} given by $H(r) = 1$ if $r > 0$ and $H(r) = 0$ if $r \leq 0$.

The \textit{energy} $E_{\mb x}$ of a binary pattern $\mb x$ in a Hopfield network  is  defined to be 
\begin{equation}\label{hopfield_energy}
E_{\mb x}(\mb J, \theta) := -\frac{1}{2}\mb x^{\top} \mb J \mb x + \theta^{\top}\mb  x = - \sum_{i < j} x_i x_j J_{ij} + \sum_{i=1}^n \theta_i x_i,
\end{equation}
identical to the energy function for an Ising spin glass.  In fact, the dynamics of a Hopfield network can be seen as 0-temperature Gibbs sampling of this energy function.  A fundamental  property of Hopfield networks is that asynchronous dynamical updates do not increase the energy (\ref{hopfield_energy}).  
Thus, after a finite number of updates, each initial state $\mb x$ converges to a \textit{fixed-point} $\mb x^{*} = (x_1^*,\ldots,x_n^*)^{\top}$ of the dynamics; that is, $x^{*}_i  = H(\mb J_i \mb x^{*} - \theta_i)$ for each $i$.   See Fig.~\ref{fighopnet} for a sample Hopfield network on $n=3$ nodes.

Given a binary pattern $\mb x$, the \textit{neighborhood} $\mathcal{N}(\mb x)$ of $\mathbf{x}$ consists of those binary vectors which are Hamming distance $1$ away from $\mb x$ (i.e., those with exactly one bit different from $\mb x$).   
We say that $\mb x$ is a \textit{strict local minimum} if every $\mb x' \in \mathcal{N}(\mb x)$ has a strictly larger energy:
\begin{equation}\label{energy_diff}
0 > E_{\mb x}  - E_{\mb x'} =  (\mb J_i \mb x - \theta_i)\delta_i,
\end{equation}
where $\delta_i = 1-2 x_i$ and $x_i$ is the bit that differs between $\mb x$ and $\mb x'$.  It is straightforward to verify that if $\mb x$ is a strict local minimum, then it is a fixed-point of the dynamics.

A basic problem is to construct Hopfield networks with a given set $\mathcal D$ of binary patterns as fixed-points or strict local minima of the energy function (\ref{hopfield_energy}). Such networks are useful for memory denoising and retrieval since corrupted versions of patterns in $\mathcal D$ will converge through the dynamics to the originals. Traditional approaches to this problem consist of iterating over $\mathcal D$ a \textit{learning rule}  \cite{hertz1991} that updates a network's weights and thresholds given a training pattern $\mb x \in \mathcal D$.  We call a rule \textit{local} when the learning updates to the three parameters $J_{ij}$, $\theta_{i}$, and $\theta_{j}$ can be computed with access solely to $x_i, x_j$, the feedforward inputs $\mb J_i \mb x$, $\mb J_j \mb x$, and the thresholds $\theta_{i}$, $\theta_{j}$; otherwise, we call the rule \textit{nonlocal}. Note that a stricter definition of local is sometimes used, in which a learning rule is called local only if updating $J_{ij}$ depends on the states $x_i$ and $x_j$, but not on the feedforward inputs to units $i$ and $j$.  Each unit $i$ necessarily has its feedforward input $\mb J_i \mb x$ locally available, since the feedforward input is compared against the threshold $\theta_i$ when the output $x_i$ is chosen.  We therefore label learning rules which utilize feedforward input as local rules. The locality of a rule is an important feature in a network training algorithm because of its necessity in theoretical models of computation in neuroscience.
 
In \cite{hopfield1982}, Hopfield defined an \textit{outer-product learning rule} (OPR) for finding such networks.  OPR is a local rule since only the binary states of nodes $x_i$ and $x_j$ are required to update a coupling term $J_{ij}$ during training (and only the state of $x_i$ is required to update $\theta_i$).  Using OPR, at most $n/(4\log n)$ patterns can be stored without errors in an $n$-node Hopfield network \cite{weisbuch1985,mceliece1987}.  In particular, the ratio of patterns storable to the number of nodes using this rule is at most $1/(4\log n)$ memories per neuron, which approaches zero as $n$ increases.  If a small percentage of incorrect bits is tolerated, then approximately $0.15n$ patterns can be stored \cite{hopfield1982,amit1987}.

The \textit{perceptron learning rule} (PER) \cite{rosenblatt1957perceptron,minsky1988} provides an alternative method to store patterns in a Hopfield network \cite{jinwen1993}.  PER is also a local rule since updating $J_{ij}$ requires only $\mb J_i\mb x$ and $\mb J_j\mb x$ (and updating $\theta_i$ requires $\mb J_i\mb x$).  Unlike OPR, it achieves optimal storage capacity, in that if it is possible for a collection of patterns $\mathcal D$ to be fixed-points of a Hopfield network, then PER will converge to parameters $\mb J, \theta$ for which all of $\mathcal D$ are fixed-points.  However, training frequently takes many parameter update steps (see Fig.~\ref{fig9}), and the resulting Hopfield networks do not generalize well (see Fig.~\ref{fig2}) nor store patterns from corrupted samples (see Fig.~\ref{fig5a}).

Despite the connection to the Ising model energy function, and the common usage of Ising spin glasses (otherwise referred to as Boltzmann machines \cite{hinton1986learning}) to build probabilistic models of binary data, we are aware of no previous work on associative memories that takes advantage of a probabilistic interpretation during training.  Probabilistic interpretations have been used for pattern recovery \cite{Sommer1998}.

\subsection{Theoretical Results}  We give an efficient algorithm for storing at least $n$ binary patterns as strict local minima (and thus fixed-points) in an $n$-node Hopfield network, and we prove that this algorithm achieves the optimal storage capacity achievable in such a network.  We also present a novel local learning rule for the training of neural networks.

Consider a collection of $m$ binary $n$-bit patterns $\mathcal{D}$ to be stored as strict local minima in a Hopfield network.  Not all collections of $m$ such patterns $\mathcal{D}$ can so be stored; for instance, from (\ref{energy_diff}) we see that no two binary patterns one bit apart can be stored simultaneously.  Nevertheless, we say that the collection $\mathcal{D}$ \textit{can be stored as local minima} of a Hopfield network if there is some $\mathcal{H} = (\mb J, \theta)$ such that each $\mb x \in \mathcal{D}$ is a strict local minimum of the energy function $E_{\mb x}(\mb J,\theta)$ in (\ref{hopfield_energy}).

The \textit{minimum probability flow} (MPF) objective function given the collection $\mathcal{D}$ is
\begin{equation}\label{MPFobjective}
K_{\mathcal{D}}(\mb J,\theta) := \sum_{\mb x \in \mathcal{D}} \ \sum_{\mb x' \in \mathcal{N}(\mb x)} \exp \left(\frac{E_{\mb x}-E_{\mb x'}}{2}\right).
\end{equation}
The function in (\ref{MPFobjective}) is infinitely differentiable and strictly convex in the parameters.
Notice that when $K_{\mathcal{D}}(\mb J,\theta)$ is small, the energy differences $E_{\mb x}-E_{\mb x'}$ between $\mb x \in \mathcal{D}$ and patterns $\mb x'$ in neighborhoods $\mathcal{N}(\mb x)$ will satisfy (\ref{energy_diff}), making $\mb x$ a fixed-point of the dynamics.

As the following result explains, minimizing (\ref{MPFobjective}) given a storable set of patterns will determine a Hopfield network storing those patterns.

\textbf{Theorem 1.} \textit{If a set of binary vectors $\mathcal{D}$ can be stored as local minima of a Hopfield network, then minimizing the convex MPF objective (\ref{MPFobjective}) will find such a network.}

\textit{Proof:} We first claim that $\mathcal{D}$ can be stored as local minima of a Hopfield network $\mathcal{H}$ if and only if the MPF objective (\ref{MPFobjective}) satisfies $K_{\mathcal{D}}(\mb J,\theta) < 1$ for some $\mb J$ and $\theta$. Suppose first that $\mathcal{D}$ can be made strict local minima with parameters $\mb J$ and $\theta$.  Then for each $\mb x \in \mathcal{D}$ and $\mb x' \in \mathcal{N}(\mb x)$, inequality (\ref{energy_diff}) holds.  In particular, a uniform scaling in the parameters will make the energy differences in (\ref{MPFobjective}) arbitrarily large and negative, and thus $K$ can be made less than $1$.  Conversely, suppose that $K_{\mathcal{D}}(\mb J,\theta) < 1$ for some choice of $\mb J$ and $\theta$.  Then each term in the sum of positive numbers (\ref{MPFobjective}) is less than $1$.  This implies that the energy difference between each $\mb x \in \mathcal{D}$ and $\mb x' \in \mathcal N(\mb x)$  satisfies (\ref{energy_diff}).  Thus, $\mathcal{D}$ are all strict local minima.  

We now explain how the claim proves the theorem.  Suppose that $\mathcal{D}$ can be stored as local minima of a Hopfield network; then, $K_{\mathcal{D}}(\mb J,\theta) < 1$ for some $\mb J, \mb \theta$.  Any method producing parameter values $\mb J $ and $\theta$ having objective (\ref{MPFobjective}) arbitrarily close to the infimum of $K_{\mathcal{D}}(\mb J,\theta)$ will produce a network with MPF objective strictly less than $1$, and therefore store $\mathcal{D}$ by above.
\qed

Our next main result is that at least $n$ patterns in an $n$-node Hopfield network can be stored by minimizing (\ref{MPFobjective}).  To make this statement mathematically precise, we introduce some notation.  Let $r(m,n) < 1$ be the probability that a collection of $m$ binary patterns chosen uniformly at random from all ${2^n \choose m}$ $m$-element subsets of $\{0,1\}^n$ can be made local minima of a Hopfield network.  The \textit{pattern capacity} (per neuron) of the Hopfield network is defined to be the supremum of all real numbers $a > 0$ such that 
\begin{equation}\label{capacity_def}
\lim_{n \to \infty} r(an,n) = 1.
\end{equation}

\textbf{Theorem 2.} \textit{The pattern capacity of an $n$-node Hopfield network is at least $1$ pattern per neuron.}

In other words, for any fixed $a < 1$, the fraction of all subsets of $m = an$ patterns that can be made strict local minima (and thus fixed-points) of a Hopfield network with $n$ nodes converges to $1$ as $n$ tends to infinity.  Moreover, by Theorem 1, such networks can be found by minimizing (\ref{MPFobjective}). Although the Cover bound \cite{cover1965} forces $a \leq 2$, it is an open problem to determine the exact critical value of $a$ (i.e., the exact pattern capacity of the Hopfield network).  Note that a perceptron with an asymmetric weight matrix can achieve the Cover bound and store $2N$ arbitrary mappings, but its stored mappings will not be local minima of an associated energy function, and the learned network will not be equivalent to a Hopfield network \cite{Gardner1987}.   Experimental evidence suggests that the limit in (\ref{capacity_def}) is $1$ for all $a < 1.5$, but converges to $0$ for $a > 1.7$ (see Fig.~\ref{fig1}).

We close this section by defining a new learning rule for a neural network. In words, the \textit{minimum probability flow learning rule} (MPF) takes an input training pattern $\mb x$ and moves the parameters $(\mb J,\theta)$ a small amount in the  direction of steepest descent of the MPF objective function $K_{\mathcal{D}}(\mb J,\theta)$ with $\mathcal{D} = \{\mb x\}$.  Mathematically, these updates for $J_{ij}$ and $\theta_i$ take the form (where again, \mbox{$\mb \delta = \mb 1 - 2\mb x$}):
\begin{eqnarray}\label{online_MPF_update}
\Delta J_{ij} &\propto & -\delta_i x_j e^{\frac{1}{2} \left( \mb J_i \mb x - \theta_i \right)\delta_i } 
	 - \delta_j x_i e^{\frac{1}{2} \left( \mb J_j \mb x - \theta_j \right)\delta_j} \\
\Delta \theta_i &\propto &\delta_i e^{ \frac{1}{2} \left( \mb J_i \mb x - \theta_i \right)\delta_i}. \label{online_MPF_thresh_update}
\end{eqnarray}
It is clear from (\ref{online_MPF_update}),(\ref{online_MPF_thresh_update}) that MPF is a local learning rule.

\subsection{Experimental Results}  We performed several experiments comparing standard techniques for fitting Hopfield networks with  minimizing the MPF objective function (\ref{MPFobjective}).  All computations were performed on standard desktop computers, and we used used the limited-memory Broyden-Fletcher-Goldfarb-Shanno (L-BFGS) algorithm \cite{nocedal1980} to minimize (\ref{MPFobjective}).

\begin{figure}[!t]
\centering
\includegraphics[width=3.4in]{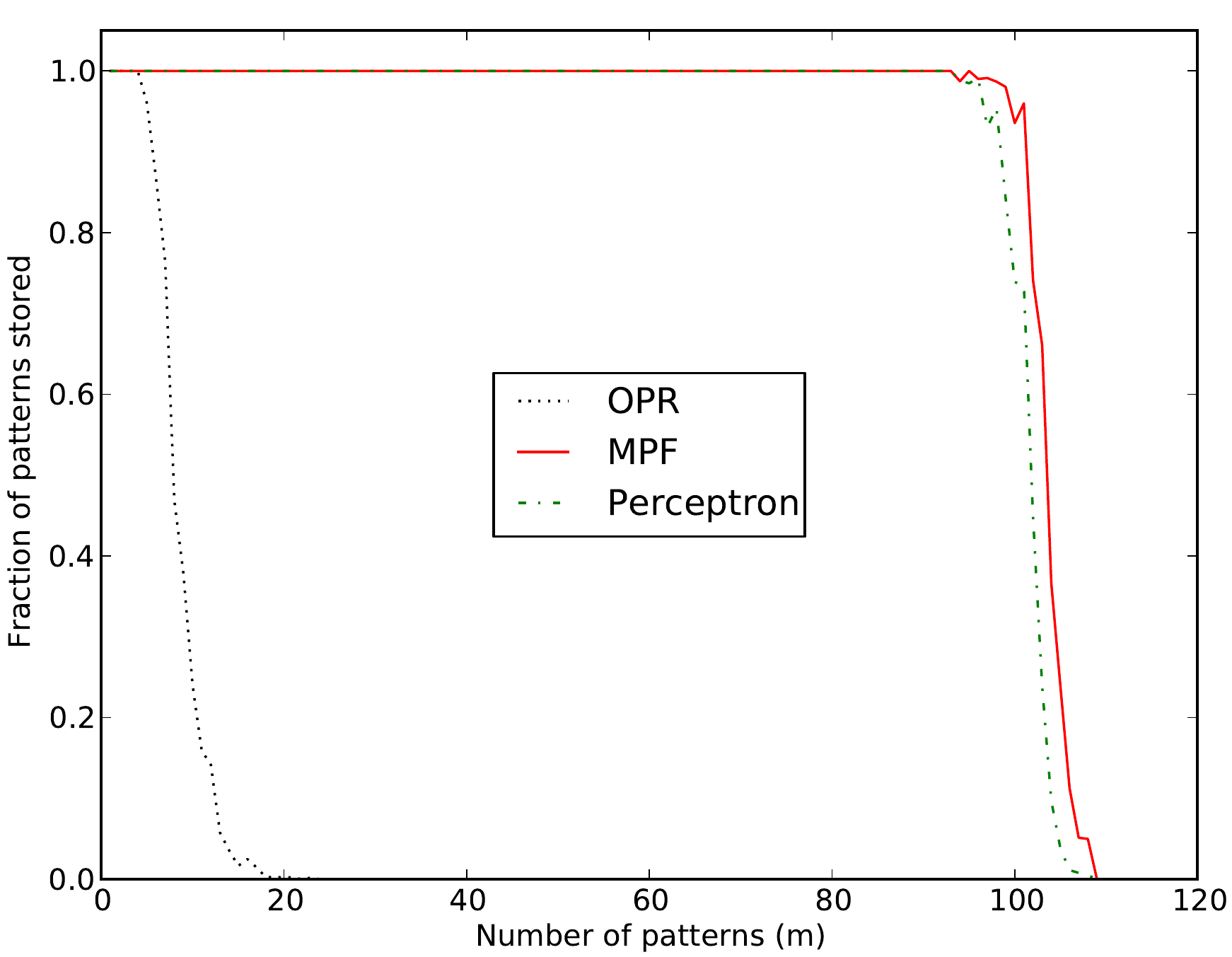}
\caption{Shows fraction of patterns made fixed-points of a Hopfield network using OPR (outer-product rule), MPF (minimum probability flow), and PER (perceptron) as a function of the number of randomly generated training patterns $m$.  Here, $n = 64$ binary nodes and we have averaged over $t = 20$ trials.  The slight difference in performance between MPF and PER is due to the extraordinary number of iterations required for PER to achieve perfect storage of patterns near the critical pattern capacity of the Hopfield network.  See also Fig.~\ref{fig9}.}
\vspace{-.25cm}
\label{fig1}
\end{figure}

In our first experiment, we compared MPF to the two methods OPR and PER for finding $64$-node Hopfield networks storing a given set of patterns $\mathcal{D}$.  For each of $20$ trials, we used the three techniques to store a randomly generated set of $m$ binary patterns, where $m$ ranged from $1$ to $120$.  The results are displayed in Fig.~\ref{fig1} and support the conclusions of Theorem 1 and Theorem 2.  

\begin{figure}[!t]
\centering
\includegraphics[width=3.4in]{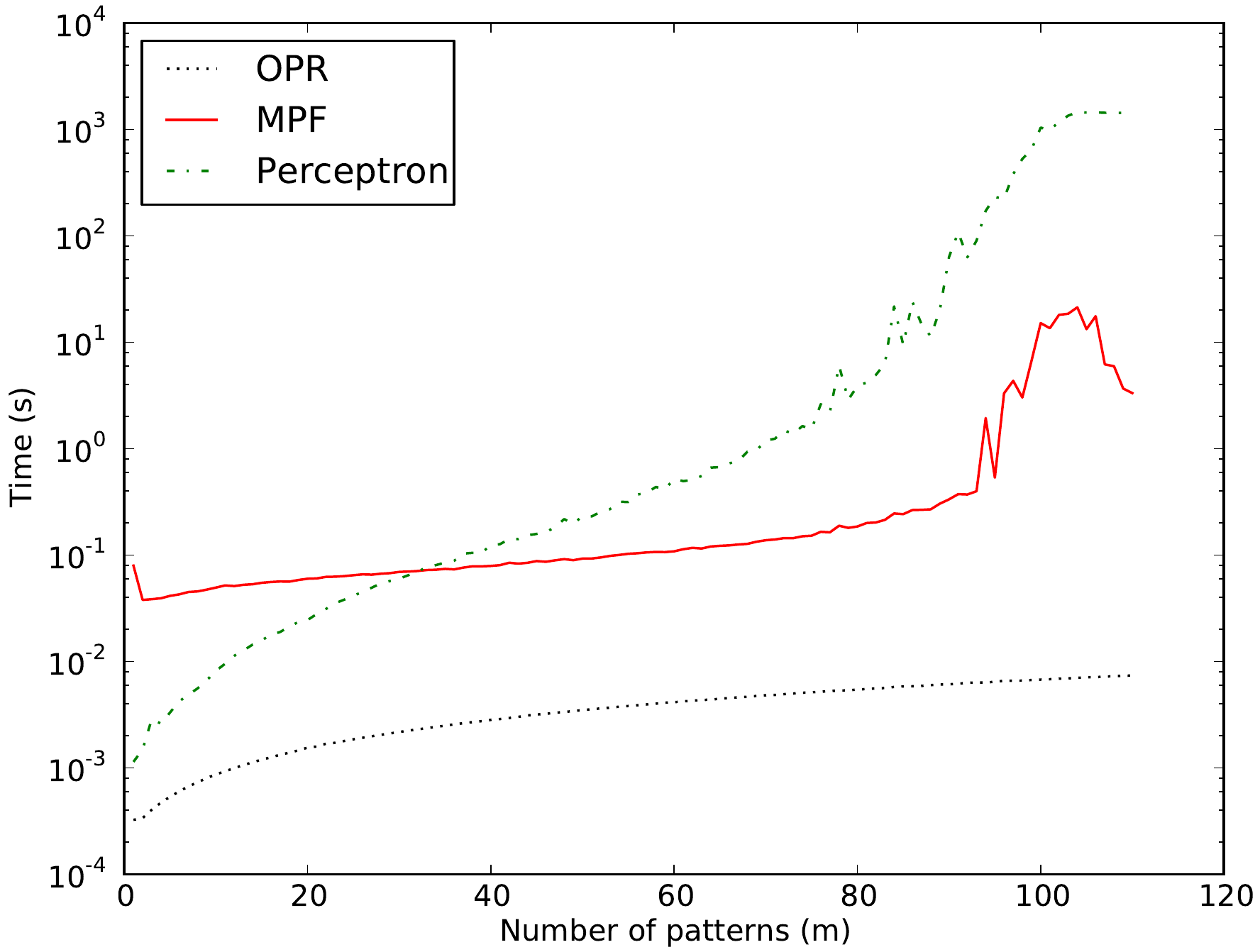}
\caption{Shows time (on a log scale) to train a Hopfield network with $n = 64$ neurons to store $m$ patterns using OPR, PER, and MPF (averaged over $t=20$ trials).}
\label{fig9}
\end{figure}

To study the efficiency of our method, we compared training time of a $64$-node network as in Fig.~\ref{fig1} with the three techniques OPR, MPF, and PER.  The resulting computation times are displayed in Fig.~\ref{fig9} on a logarithmic scale.  Notice that computation time for MPF and PER significantly increases near the pattern capacity threshold of the Hopfield network.

\begin{figure}[!t]
\centering 
\includegraphics[width=5.5in]{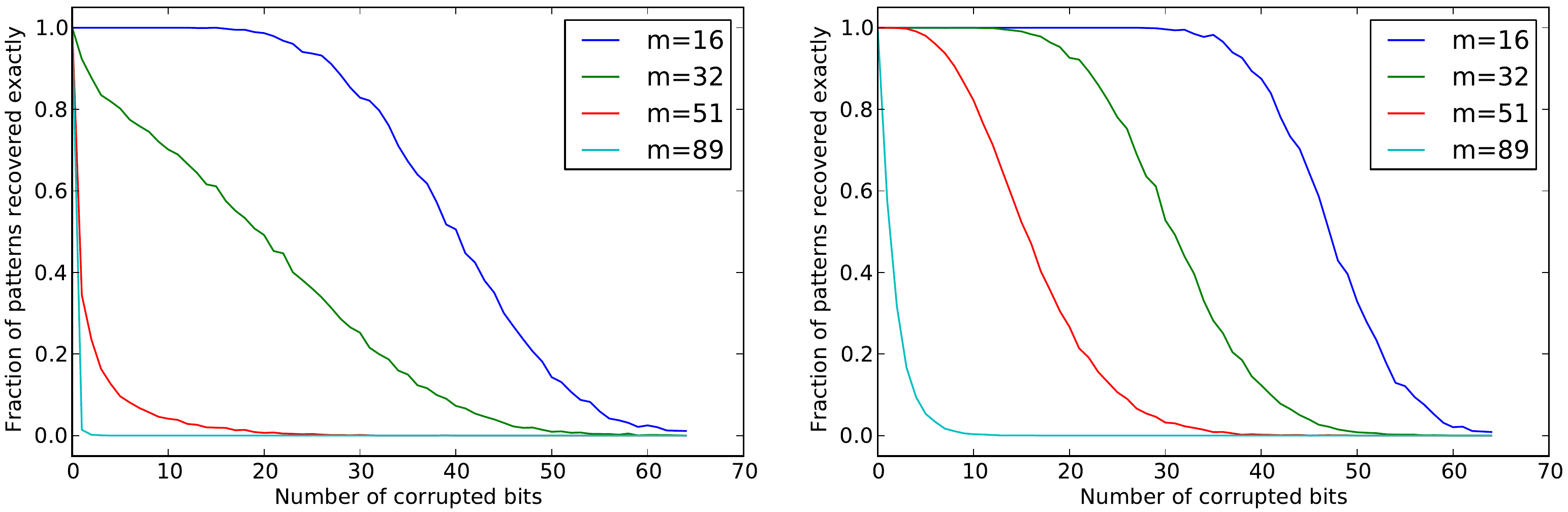}
\caption{Shows fraction of exact pattern recovery for a perfectly trained $n = 128$ Hopfield network using rules PER (figure on the left) and MPF (figure on the right) as a function of bit corruption at start of recovery dynamics for various numbers $m$ of patterns to store.  We remark that this figure and the next do not include OPR as its performance was far worse than either MPF or PER.}
\vspace{-.25cm}
\label{fig2}
\end{figure}

For our third experiment, we compared the denoising performance of MPF and PER.  For each of four values for $m$ in a $128$-node Hopfield network, we determined weights and thresholds for storing all of a set of $m$ randomly generated binary patterns using both MPF and PER.  We then flipped $0$ to $64$ of the bits in the stored patterns and let the dynamics (\ref{Hopdynamics}) converge (with weights and thresholds given by MPF and PER), recording if the converged pattern was identical to the original pattern or not.  Our results are shown in Fig~\ref{fig2}, and they demonstrate the superior corrupted memory retrieval performance of MPF.

\begin{figure}[!t]
\centering
\includegraphics[width=3.4in]{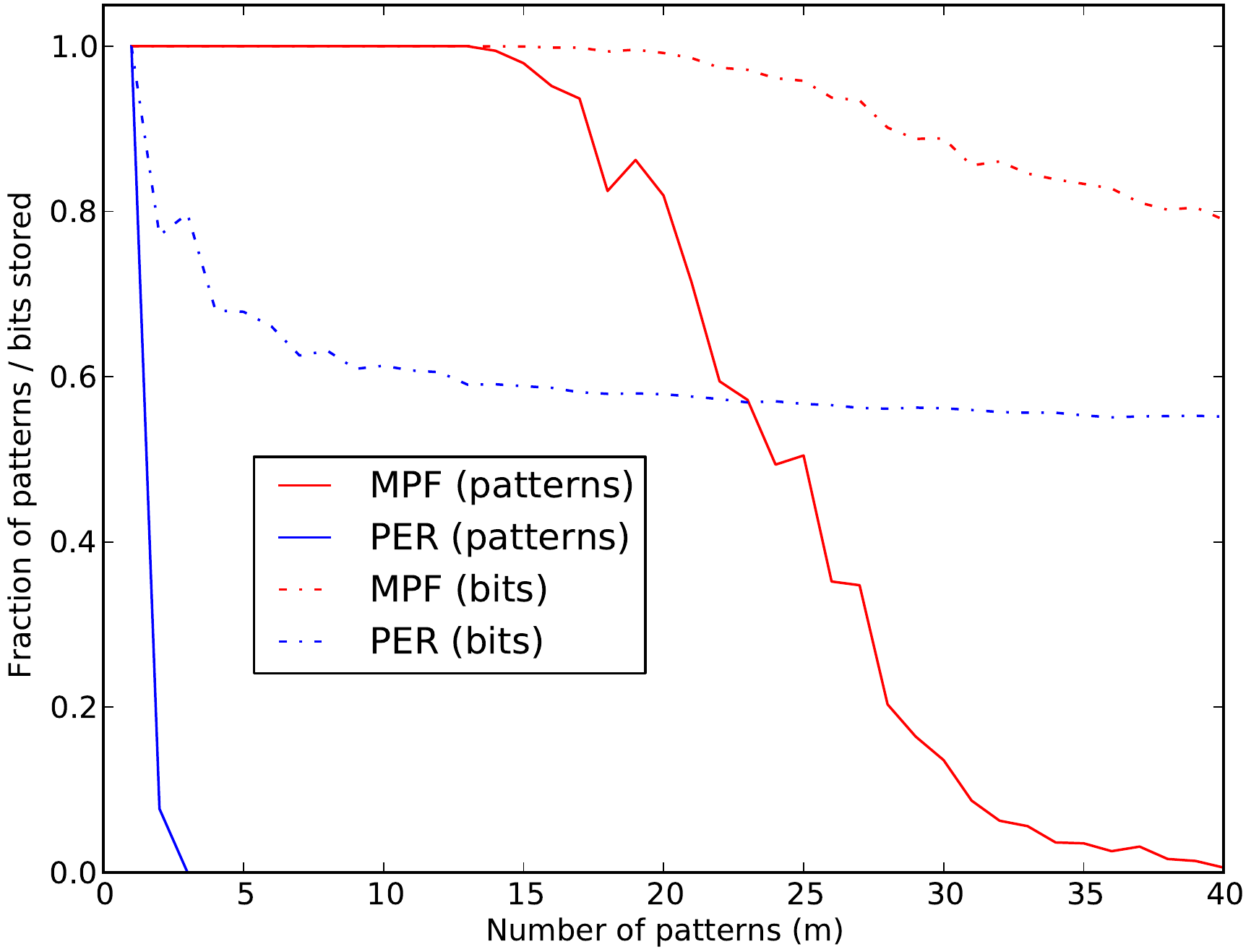}
\caption{Shows fraction of patterns (shown in red for MPF and blue for PER) and fraction of bits (shown in dotted red for MPF and dotted blue for PER) recalled of trained networks (with $n = 64$ nodes each) as a function of the number of patterns $m$ to be stored.  Training patterns were presented repeatedly with 20 bit corruption (i.e., 31\% of the bits flipped). (averaged over t = 13 trials.)}
\vspace{-.25cm}
\label{fig5a}
\end{figure}

A surprising final finding in our investigation was that MPF can store patterns from highly corrupted or noisy versions on its own and without supervision.  This result is explained in Fig~\ref{fig5a}.  To illustrate the experiment visually, we stored $m = 80$ binary fingerprints in a $4096$-node Hopfield network using a large set of training samples which were corrupted by flipping at random $30\%$ of the original bits; see Fig.~\ref{fig7a} for more details.

\subsection{Discussion} We have presented a novel technique for the storage of patterns in a Hopfield associative memory.   The first step of the method is to fit an Ising model using minimum probability flow learning to a discrete distribution supported equally on a set of binary target patterns.  Next, we use the learned Ising model parameters to define a Hopfield network.  We show that when the set of target patterns is storable, these steps result in a Hopfield network that stores all of the patterns as fixed-points.  We have also demonstrated that the resulting (convex) algorithm outperforms current techniques for training Hopfield networks.

We have shown improved recovery of memories from noisy patterns  and improved training speed as compared to training by PER.  We have demonstrated optimal storage capacity in the noiseless case, outperforming OPR.  We have also demonstrated the unsupervised storage of memories from heavily corrupted training data.  Furthermore, the learning rule that results from our method is local; that is, updating the weights between two units requires only their states and feedforward input.  

It is the probabilistic interpretation of the Hopfield network used in the MPF rule that leads to the superior robustness to noise and graceful degradation in the case of more patterns than can be stored as fixed points.  The probabilistic learning objective tries not only to make the observed data states probable, but also to make unobserved states improbable.  This second aspect reduces the probability mass assigned to spurious minima and their attractive wells, improving pattern recovery from noisy initialization.  Additionally, when more patterns are presented than can be stored, the probabilistic objective attempts to carve out a broad minima in the energy landscape around clusters of datapoints.  If many noisy examples of template patterns are presented, the lowest energy states will tend to lie in the center of the minima corresponding to each cluster of data, and will thus tend to correspond to the template states.

As MPF allows the fitting of large Hopfield networks quickly, new investigations into the structure of Hopfield networks are posssible \cite{HTK}.  
It is our hope that the robustness and speed of this learning technique will enable practical use of Hopfield associative memories in both computational neuroscience, computer science, and scientific modeling.

\chapter{
The Natural Gradient by Analogy to Signal Whitening, and Recipes and Tricks for its Use
}
\label{chap natgrad}

Difficulties in training probabilistic models can stem from ill conditioning of the model's parameter space as well as from an inability to analytically normalize the model.  In this chapter we review how an ill conditioned parameter space can undermine learning, and we present a novel interpretation of a common technique for dealing with this ill conditioning, the natural gradient.  In addition, we present tricks and specific prescriptions for applying the natural gradient to learning problems.  Material in this chapter is taken from \cite{Sohl-Dickstein2012b}.

The natural gradient, as introduced by \cite{Amari1987}, allows for more efficient gradient descent by removing dependencies and biases inherent in a function's parameterization.  Several papers present the topic thoroughly and precisely \cite{Amari1987,Amari1998,Amari2000,Theis2005,Amari2010}.  It remains a very difficult idea to get your head around however.  The intent of this chapter is to provide simple intuition for the natural gradient and its uses.  The natural gradient is explained by analogy to the more widely understood concept of signal whitening.  To our knowledge, this is the first time a connection has been made between signal whitening and the natural gradient.

\section{Natural gradient}

\subsection{A simple example \label{simple example}}

We begin with a simple probabilistic model which has clearly been very poorly parametrized.  For this we use a two dimensional gaussian distribution, with means written in terms of the parameters $\theta \in \mathcal R^2$,
\begin{eqnarray}
q\left( \mb x; \theta \right) = \frac{1}{2 \pi}\exp\left[ -\frac{1}{2}\left( x_1 - \left[3 \theta_1 + \frac{1}{3} \theta_2 \right]\right)^2 - \frac{1}{2}\left( x_2 - \left[\frac{1}{3} \theta_1 \right]\right)^2 \right]
.
\end{eqnarray}
As an objective function $J\left( \theta \right)$ we use the negative log likelihood of $q\left( \mb x; \theta \right)$ under an observed data distribution $p\left( \mb x \right)$
\begin{eqnarray}
\label{log like obj}
J\left( \theta \right) = 
-\left< \log q\left( \mb x; \theta \right) \right>_{p\left( \mb x \right)}
.
\end{eqnarray}
Using steepest gradient descent to minimize the negative log likelihood involves taking steps like
\begin{eqnarray}
\Delta \theta &  \propto & -\nabla_\theta J\left( \theta \right) \\
\label{descent equation}
\left[
\begin{matrix}
\Delta \theta_1 \\
\Delta \theta_2
\end{matrix}
\right]
 & \propto & 
\left[
\begin{matrix}
	\left<
		3 \left( x_1 - \left[3 \theta_1 + \frac{1}{3} \theta_2 \right]\right)
		+
		\frac{1}{3} \left( x_2 - \left[\frac{1}{3} \theta_1 \right]\right)
	\right>_{p\left( \mb x \right)} \\
	\left<
		\frac{1}{3} \left( x_1 - \left[3 \theta_1 + \frac{1}{3} \theta_2 \right]\right)
	\right>_{p\left( \mb x \right)}
\end{matrix}
\right]
.
\end{eqnarray}

\begin{figure}
\begin{center}
\begin{tabular}{cc}
(a)\includegraphics[width=0.4\linewidth]{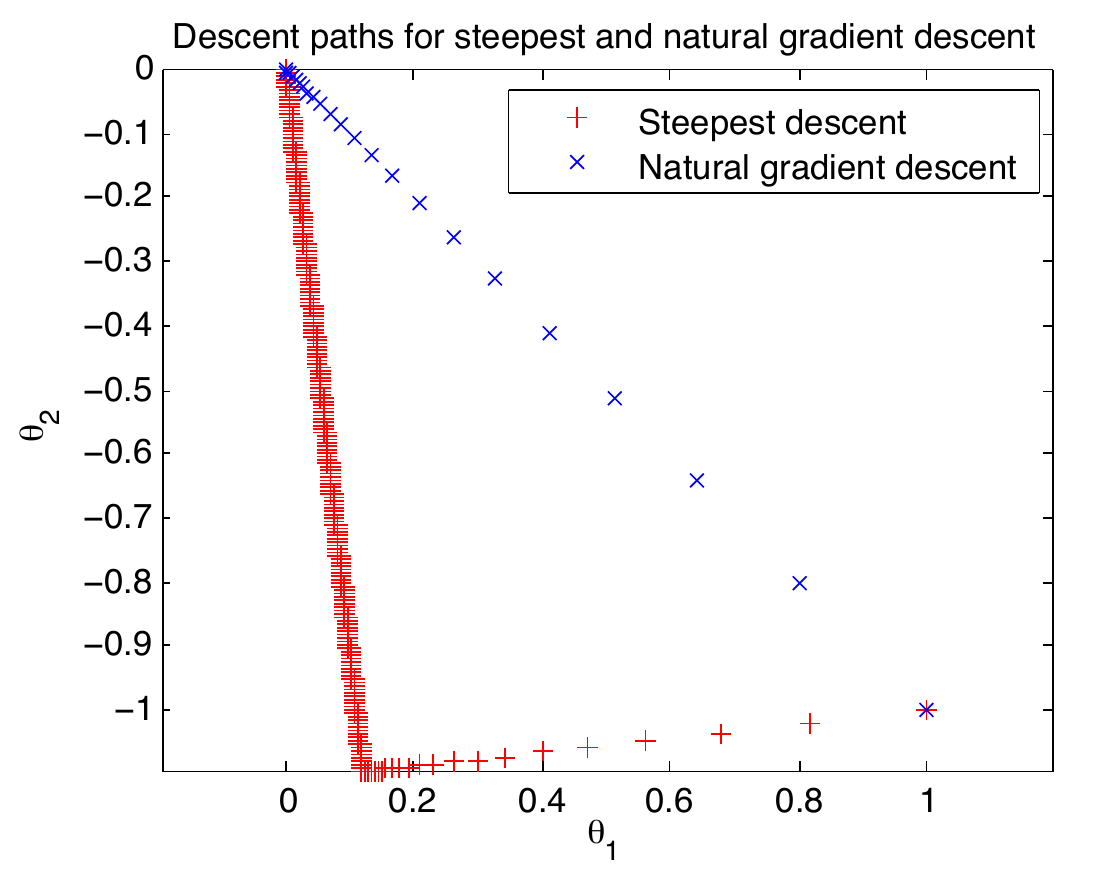}
&
(b)\includegraphics[width=0.4\linewidth]{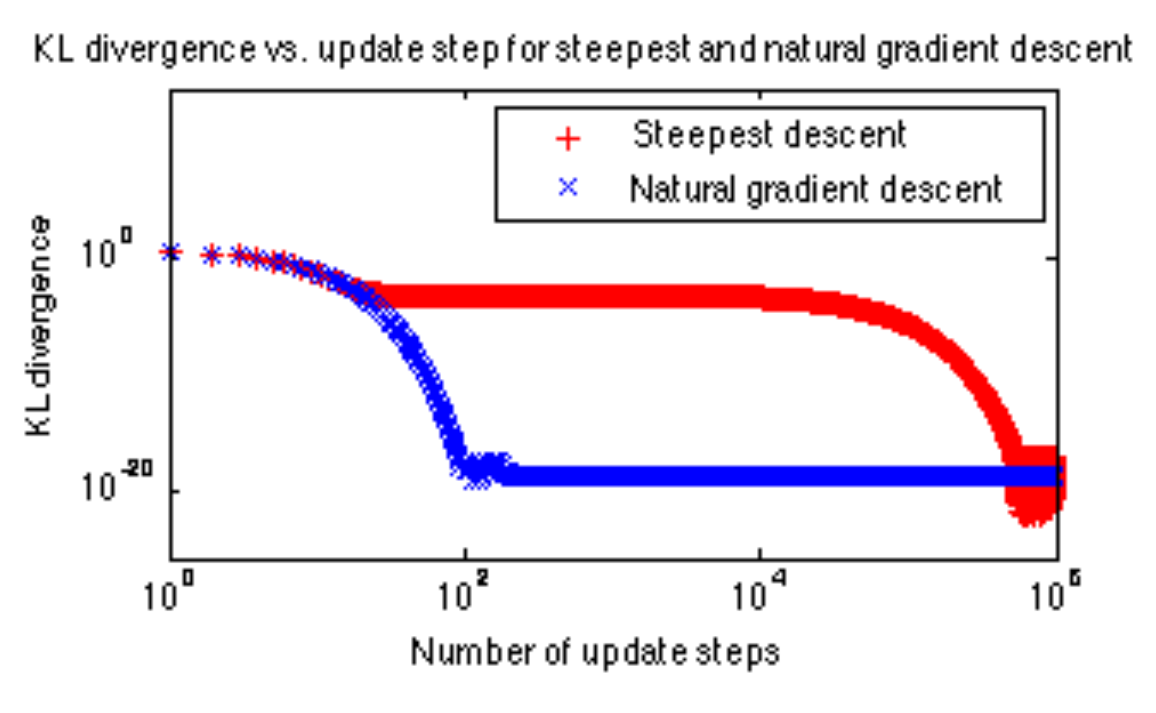}
\\
\\
(c)\includegraphics[width=0.4\linewidth]{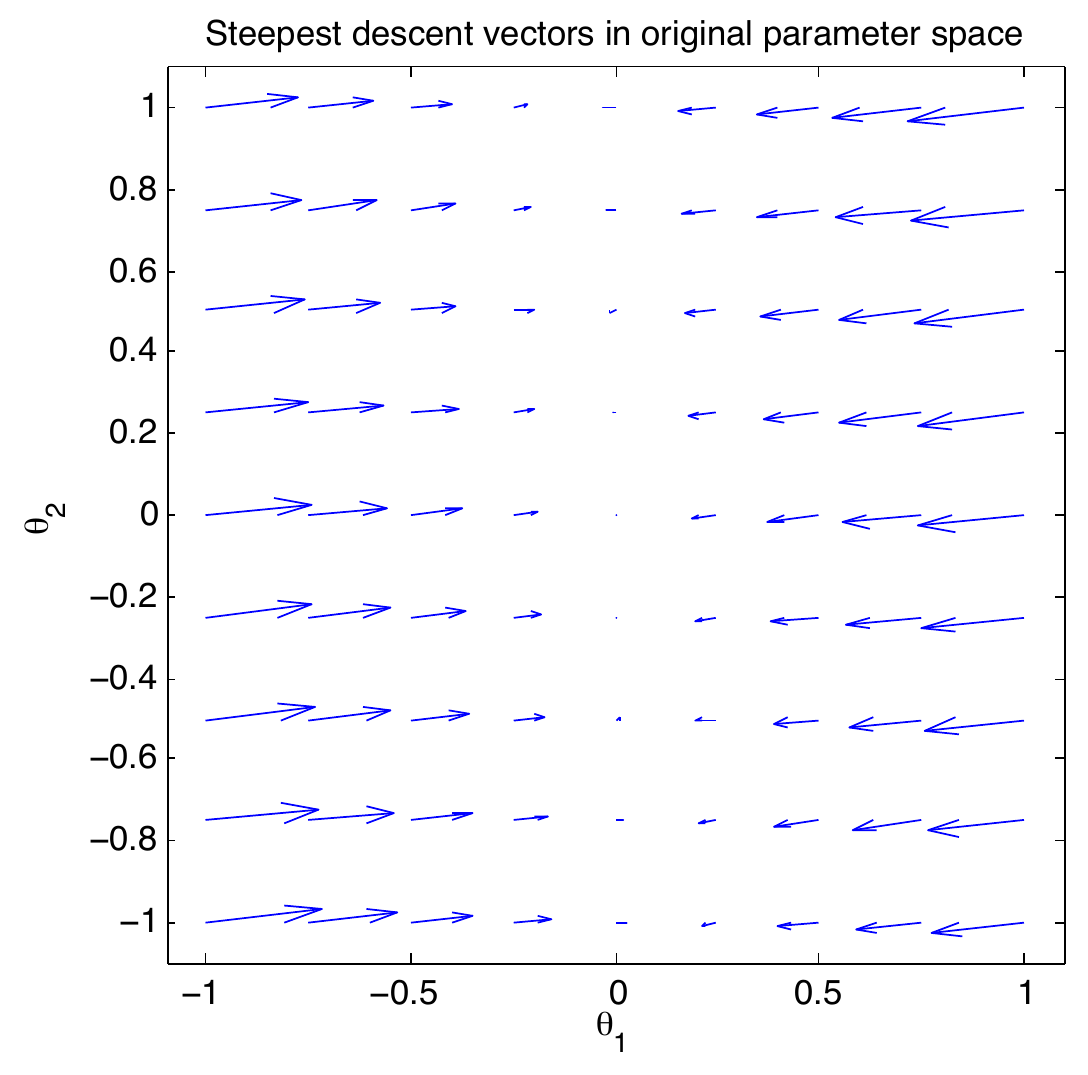}
&
(d)\includegraphics[width=0.4\linewidth]{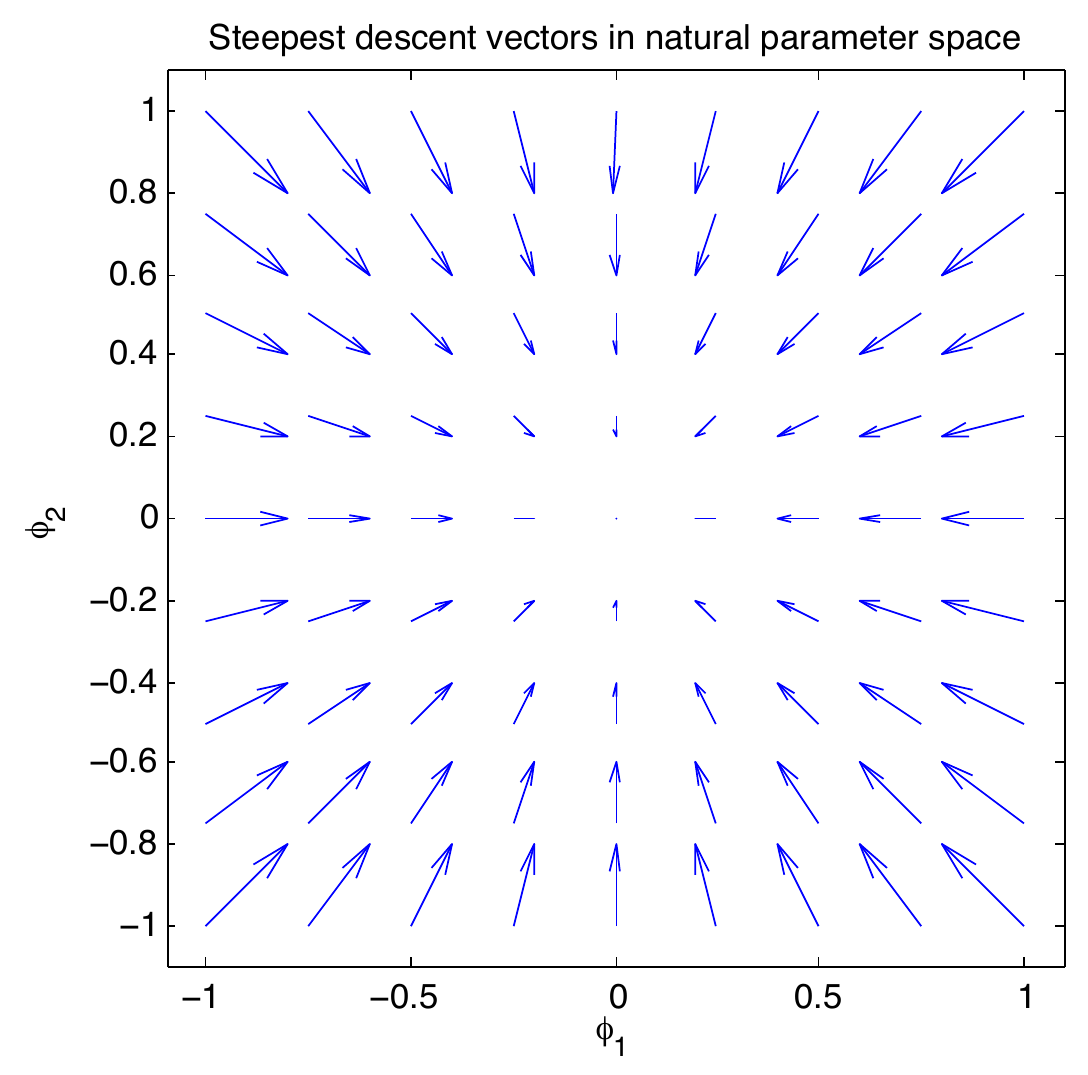}
\end{tabular}
\end{center}
\caption{
\emph{(a)} The parameter descent paths taken by steepest gradient descent (red) and natural gradient descent (blue) for the example given in Section \ref{simple example}.  The parameters are initialized at $\theta_{init} = \left[1, -1\right]^T$, and are fit to data generated with $\theta_{true} = \left[0, 0\right]^T$.  The Fisher information matrix (Equation \ref{fisher recipe}) is used to calculate the natural gradient.  Notice that steepest descent takes a more circuitous and far slower path.  
\emph{(b)} The KL divergence between the data distribution and the fit model as a function of number of gradient descent steps.  Descent using the natural gradient converges more quickly.  
\emph{(c)} The arrows give the gradient of the log likelihood objective (Equation \ref{log like obj}), for a grid of parameter settings.  This is the descent direction provided by Equation \ref{descent equation}.
\emph{(d)} The gradient of the same log likelihood objective (Equation \ref{log like obj}), but in terms of the whitened, natural, parameter space $\phi$ as described in Section \ref{whitened space}.  Note that steepest descent in the whitened space converges directly to the true parameter values $\phi_{true} = \mb G^\frac{1}{2} \theta_{true} = \left[0, 0\right]^T$.
}
\label{gauss_pics}
\end{figure}

As can be seen in Figure \ref{gauss_pics}{\em{a}} the steepest gradient update steps can move the parameters in a direction nearly perpendicular to the desired direction.  $q\left( \mb x; \theta \right)$ is much more sensitive to changes in $\theta_1$ than $\theta_2$, so the step size in $\theta_1$ should be much smaller, but is instead much larger.  In addition, $\theta_1$ and $\theta_2$ are not independent of each other.  They move the distribution in nearly the same direction, making movement in the perpendicular direction particularly difficult.  Getting the parameters here to fully converge via steepest descent is a slow proposition, as shown in Figure \ref{gauss_pics}{\em{b}}.

The pathological learning gradient above is illustrative of a more general problem.  A model's learning gradient is effected by the parameterization of the model as well as the objective function being minimized.  The effects of the parameterization can dominate learning.  The natural gradient is a technique to remove the effects of model parameterization from learning updates.

\subsection{A metric on the parameter space}\label{sec metric}

As a first step towards compensating for differences in relative scaling, and cross-parameter dependencies, the shape of the parameter space $\theta$ is first described by assigning it a measure of distance, or a metric.  This metric is expressed via a symmetric matrix $\mb G\left( \theta \right)$, which defines the length $\left| d \theta \right|$ of an infinitesimal step $d \theta$ in the parameters,
\begin{eqnarray}
\label{metric def}
\left| d \theta \right|^2 = \sum_i \sum_j G_{ij}\left( \theta \right) d \theta_i d \theta_j = d \theta^T \mb G\left( \theta \right) d \theta
.
\end{eqnarray}
$\mb G\left( \theta \right)$ is chosen so that the length $\left| d \theta \right|$ provides a reasonable measure for the expected magnitude of the difference of $J\left( \theta + d \theta \right)$ from $J\left( \theta \right)$.  That is, $\mb G\left( \theta \right)$ is chosen such that $\left| d \theta \right|$ is representative of the expected magnitude of the change in the objective function resulting from a step $d \theta$.  There is no uniquely correct choice for $\mb G\left( \theta \right)$. 

If the objective function $J\left( \theta \right)$ is the log likelihood of a probability distribution $q\left( \mb x; \theta \right)$, then a measure of the information distance between $q\left( \mb x; \theta + d \theta \right)$ and $q\left( \mb x; \theta \right)$ usually works well, and the Fisher information matrix (Equation \ref{fisher recipe}) is frequently used as a metric.  Plugging in the example from Section \ref{simple example}, the resulting Fisher information matrix is $\mb G = \left[\begin{array}{cc}3^2 + \frac{1}{3^2} & 1 \\1 & \frac{1}{3^2}\end{array}\right]$.


\begin{figure}
\parbox[c]{\textwidth}{
\center{
(a)\includegraphics[width= 0.4\linewidth]{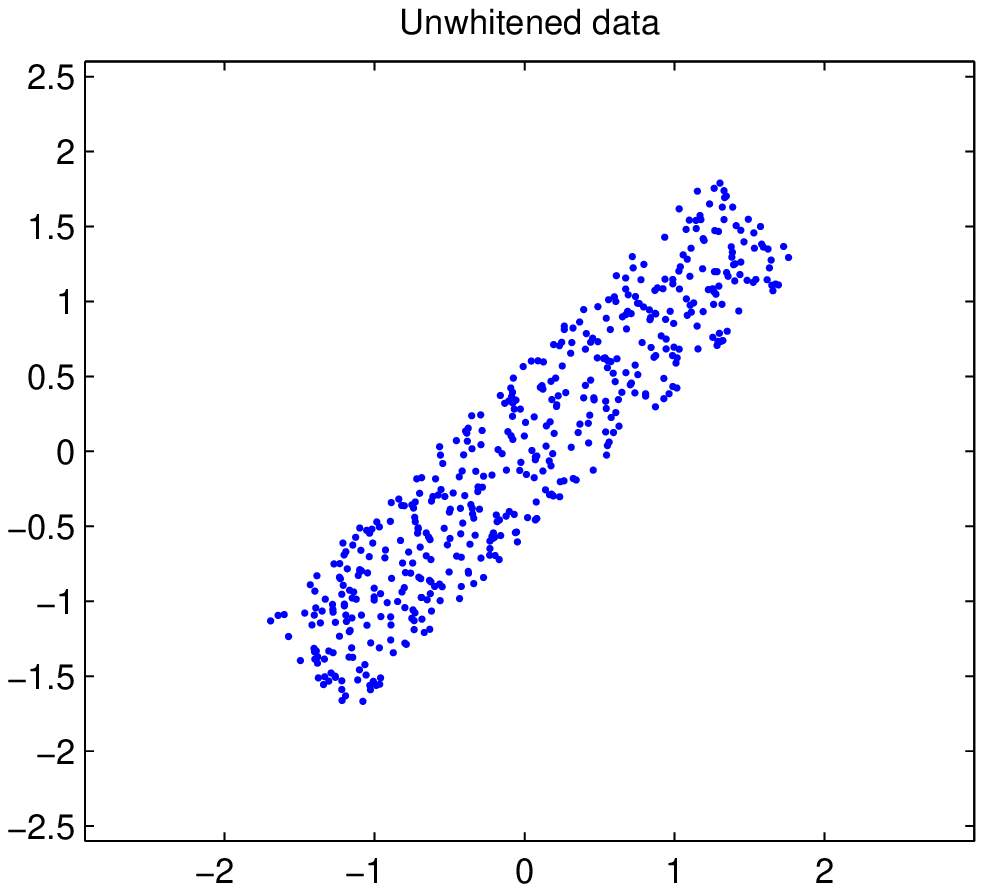} 
(b)\includegraphics[width= 0.4\linewidth]{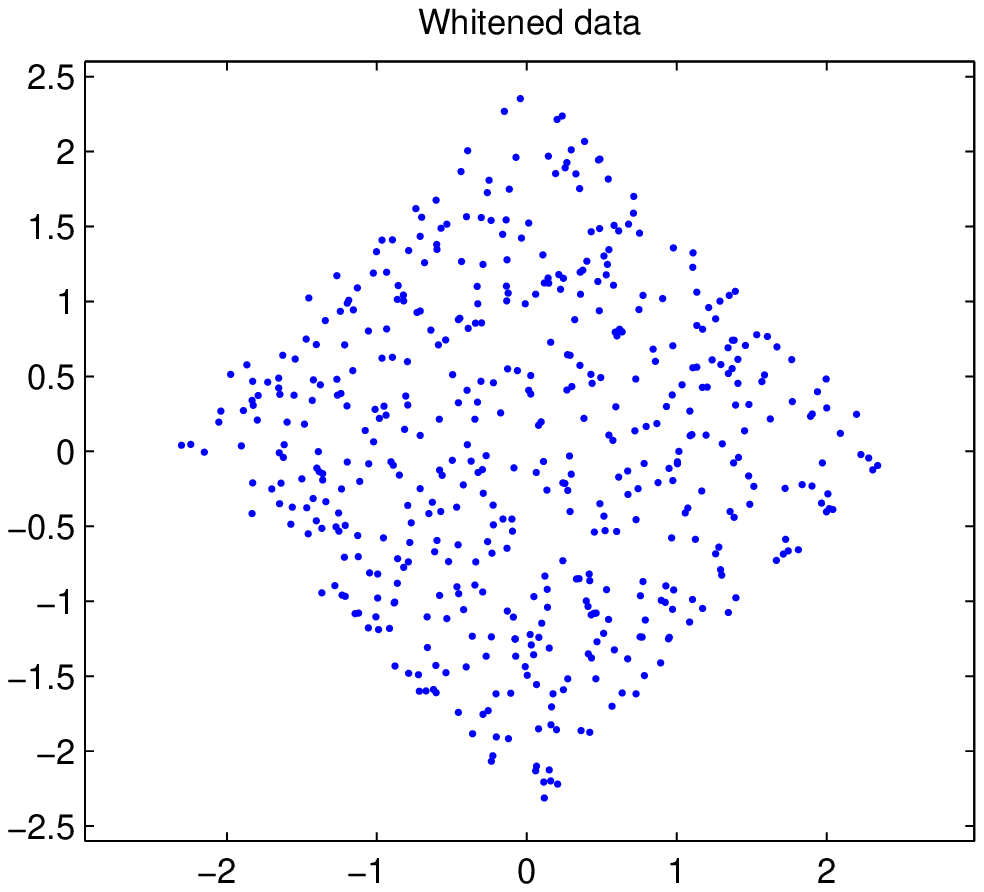}
}
}
\caption{
Example of signal whitening.
\emph{(a)} Samples $\mb x$ from an unwhitened distribution in 2 variables.  
\emph{(b)} The same samples after whitening, in new variables $\mb y = \mb W\mb x = \mb\Sigma^{-\frac{1}{2}}\mb x$.
}
\label{whitening example}
\end{figure}

\subsection{Connection to covariance}

$\mb G\left( \theta \right)$ is an analogue of the inverse covariance matrix $\mb\Sigma^{-1}$.  Just as a signal can be whitened given $\mb\Sigma^{-1}$ --- removing all first order dependencies and scaling the variance in each dimension to unit length --- the parameterization of $J\left( \theta \right)$ can also be ``whitened," removing the dependencies and differences in scaling between dimensions captured by $\mb G\left( \theta \right)$.  See Figure \ref{whitening example} for an example of signal whitening.  

As a quick review, the covariance matrix $\mb\Sigma$ of a signal $\mb x$ is defined as
\begin{eqnarray}
\mb\Sigma = \left<\mb x \mb x^T\right>
.
\end{eqnarray}
The inverse covariance matrix is frequently used as a metric on the signal $\mb x$.  This is called the Mahalanobis distance \cite{mahalanobis1936}.  It has the same form as the definition of $\left| d \theta \right|^2$ in Equation \ref{metric def},
\begin{eqnarray}
\left| d\mb x \right|^2_{\mathrm{Mahalanobis}} = {d\mb x}^T \mb\Sigma^{-1} {d\mb x}
.
\end{eqnarray}

In order to whiten a signal $\mb x$, a whitening matrix $\mb W$ is found such that the covariance matrix for a new signal $\mb y = \mb W\mb x$ is the identity matrix $\mb I$.  The signal $\mb y$ is then a whitened version of $\mb x$,
\begin{eqnarray}
\mb I =  \left<\mb y \mb y^T\right> = \mb W \left<\mb x \mb x^T\right> \mb W^T = \mb W\mb\Sigma \mb W^T
.
\end{eqnarray}
Remembering that $\mb\Sigma^{-1}$ is symmetric, one solution\footnote{
Choosing $\mb W=\mb\Sigma^{-\frac{1}{2}}$ leads to symmetric, or zero-phase, whitening.  In some fields it is referred to as a decorrelation stretch.  It is equivalent to rotating a signal to the PCA basis, rescaling each axis to have unit norm, and then performing the inverse rotation, returning the signal to its original orientation.  All unitary transformations of $\mb\Sigma^{-\frac{1}{2}}$ also whiten the signal.
} to this system of linear equations is
\begin{eqnarray}
\mb W=\mb\Sigma^{-\frac{1}{2}} \\
\mb y = \mb\Sigma^{-\frac{1}{2}} \mb x
.
\end{eqnarray}
If the covariance matrix for $\mb y$ is the identity, then the metric for the Mahalanobis distance in the new variables $\mb y$ is also the identity ($\left| d\mb y \right|^2_{\mathrm{Mahalanobis}} = d \mb y^T d \mb y$).

Whitening is a common preprocessing step in signal processing.  It prevents incidental differences in scaling between dimensions from effecting later processing stages.

\subsection{``Whitening" the parameter space}
\label{whitened space}
If $\mb G$ is not a function of $\mb \theta$, then a similar procedure can be followed to produce a ``whitened" parameterization $\mb \phi$.  We wish to find new parameters $\mb \phi = \mb W \mb \theta$ such that the metric $\mb G$ on $\mb \phi$ is the identity $\mb I$, as the Mahalanobis metric $\mb\Sigma^{-1}$ is the identity for a whitened signal.  This will mean that a small step $d\mb  \phi$ in any direction will tend to have the same magnitude effect on the objective $J\left(\mb  \phi \right)$.
\begin{align}
\phi &= \mb W \theta \\
\left| d \phi \right|^2 & = \left| d \theta \right|^2 \\
d \phi^T \mb I d \phi & = d \theta^T \mb G d \theta \\
d \phi^T d \phi & = d \theta^T \mb G d \theta \\
d \phi & = \mb W d \theta \\
d \theta^T \mb W^T \mb W d \theta & = d \theta^T \mb G d \theta
\end{align}
Noting that $\mb G$ is symmetric, we find that one solution to this system of linear equations is
\begin{eqnarray}
\mb W = \mb G^{\frac{1}{2}} \\
\phi = \mb G^{\frac{1}{2}} \theta
\label{eq phi theta}
.
\end{eqnarray}
Steepest gradient descent steps in terms of $\phi$ descend the objective function in a more direct fashion than steepest gradient descent steps in terms of $\theta$, as is illustrated in Figure \ref{gauss_pics}{\em{c}} and \ref{gauss_pics}{\em{d}}.  In $\phi$, the steepest gradient is the natural gradient.

$\mb G$ is almost always a function of $\theta$, and for most problems there is no parameterization $\phi$ which will be ``white" everywhere.  So long as $\mb G\left( \theta \right)$ changes slowly though, it can be treated as constant for a single learning step.  This suggests the following as an algorithm for learning in a natural parameter space:
\begin{enumerate}
\item Express $J\left( \cdot \right)$ in terms of natural parameters $\phi = \mb G^{\frac{1}{2}}\left( \theta_t \right) \theta$.
\item Calculate an update step $\Delta \phi \propto \nabla_\phi J\left( \phi_t \right)$, where $\phi_t = \mb G^{\frac{1}{2}}\left( \theta_t \right) \theta_t$.
\item Calculate the $\theta_{t+1} = \mb G^{-\frac{1}{2}}\left( \theta_t \right) \left( \phi_t + \Delta \phi \right)$ associated with the update to $\phi$.
\item Repeat.\footnote{Practically, $\mb G\left( \theta \right)$ can usually be treated as constant for many learning steps.  This allows the natural gradient to be combined in a plug and play fashion with other gradient descent algorithms, like L-BFGS, by performing gradient descent on $J\left(\phi\right)$ rather than $J\left(\theta\right)$.}
\end{enumerate}
The resulting update steps more directly and rapidly descend the objective function than steepest descent steps.

\subsection{The natural gradient in $\theta$}

The parameter updates in Section \ref{whitened space} can be performed entirely in the original parameter space $\theta$.  
The natural gradient $\tilde{\nabla}_\theta J\left( \theta \right)$ is the direction in $\theta$ which is equivalent to steepest gradient descent in $\phi$ of $J\left( \phi \right)$. 
In order to find $\tilde{\nabla}_\theta J\left( \theta \right)$, we first write $\Delta \phi$ in terms of $\theta$, then we write the natural gradient update step in $\theta$, $\tilde{\Delta} \theta$, in terms of $\Delta \phi$, 
\begin{eqnarray}
\Delta \phi & \propto &
	 \nabla_\phi J\left( \phi \right) \\
& = &
\left(
	 \frac
		{\partial \theta}
		{\partial \phi^T}\right)^T
	\nabla_\theta J\left( \theta \right) \\
& = &
	\mb G^{-\frac{1}{2}}
	\nabla_\theta J\left( \theta \right) 
\end{eqnarray}
(where $\theta = \mb G^{-\frac{1}{2}}\phi$ from Equation \ref{eq phi theta}, and $\mathbf{\frac{\partial \theta}{\partial \phi^T}}$ is the Jacobian matrix),
\begin{eqnarray}
\tilde{\Delta} \theta & \propto &
	\frac
		{\partial \theta}
		{\partial \phi^T}
	\Delta \phi \\
& = &
	\mb G^{-\frac{1}{2}}
	\Delta \phi \\
& \propto &
	\mb G^{-1}
	\nabla_\theta J\left( \theta \right) 
.
\label{nat_grad_step}
\end{eqnarray}

Since the natural gradient update step is proportional to the natural gradient, $\tilde{\Delta} \theta \propto \tilde{\nabla}_\theta J\left( \theta \right)$, the natural gradient can be written as
\begin{eqnarray}
\tilde{\nabla}_\theta J\left( \theta \right)
= 
\mb G^{-1}\left( \theta \right)
	\nabla_\theta J\left( \theta \right)
.
\label{nat_grad_eq}
\end{eqnarray}

Figure \ref{gauss_pics}{\em{a}} illustrates this gradient applied to the example objective function from Section \ref{simple example}.  If gradient descent is performed by infinitesimal steps in the direction indicated by $\tilde{\nabla}_\theta J\left( \theta \right)$, then the parameterization of the problem will have no effect on the path taken during learning (though choice of $\mb G\left( \theta \right)$ will have an effect).

Surprise opportunity!  The first person to read this far and email me will receive a gift drawn at random from a complex probability distribution, but most likely a bottle of fine wine.  Later respondents may receive a miniature version of a randomly drawn gift, for instance an airline-sized wine bottle.

\section{Recipes and tricks}

In this section we present a reference with key formulas for using the natural gradient, as well as approaches useful for applying the natural gradient in specific cases. 

\subsection{Natural gradient}

The natural gradient is
\begin{eqnarray}
\tilde{\nabla}_\theta J\left( \theta \right)
=
\mb G^{-1}\left( \theta \right)
	\nabla_\theta J\left( \theta \right)
\end{eqnarray}
where $J\left( \theta \right)$ is an objective function to be minimized with parameters $\theta$, and $\mb G\left( \theta \right)$ is a metric on the parameter space.  Learning should be performed with an update rule
\begin{eqnarray}
\theta_{t+1} =  \theta_t + \tilde{\Delta} \theta_t \\
\tilde{\Delta} \theta \propto
-\tilde{\nabla}_\theta J\left( \theta \right)
\end{eqnarray}
with steps taken in the direction given by the natural gradient.

\subsection{Metric $\mb G\left( \theta \right)$}

If the objective function $J\left( \theta \right)$ is the negative log likelihood of a probabilistic model $q\left( \mb x; \theta \right)$ under an observed data distribution $p\left( \mb x \right)$
\begin{eqnarray}
J\left( \theta \right) = 
-\left< \log q\left( \mb x; \theta \right) \right>_{p\left( \mb x \right)}
\label{acceptable form}
\end{eqnarray}
then the Fisher information matrix
\begin{eqnarray}
G_{ij}\left( \theta \right) = \left< 
	\frac
		{\partial \log q\left( \mb x; \theta \right)}{\partial \theta_i}
	\frac
		{\partial \log q\left( \mb x; \theta \right)}{\partial \theta_j}
	 \right>_{q\left( \mb x; \theta \right)}
\label{fisher recipe}
\end{eqnarray}
is a good metric to use.

If the objective function is \textit{not} of of the form given in Equation \ref{acceptable form}, and cannot be transformed into that form, then greater creativity is required.  See Section \ref{non probabilistic} for some basic hints.

Remember, as will be discussed in Section \ref{sec fail}, even if the metric you choose is approximate, it is still likely to accelerate convergence!

\subsection{Fisher information over data distribution}

The Fisher information matrix (Equation \ref{fisher recipe}) requires averaging over the model distribution $q\left( \mb x; \theta \right)$.  For some models this is very difficult to do.  If that is the case, instead taking the average over the empirical data distribution $p\left( \mb x \right)$
\begin{eqnarray}
G_{ij}\left( \theta \right) = \left< 
	\frac
		{\partial \log q\left( \mb x; \theta \right)}{\partial \theta_i}
	\frac
		{\partial \log q\left( \mb x; \theta \right)}{\partial \theta_j}
	 \right>_{p\left( \mb x \right)}
\label{fisher data}
\end{eqnarray}
is frequently an effective alternative.

\subsection{Energy approximation}

Parameter estimation in a probabilistic model of the form
\begin{equation}
\label{basicmodel}
q(\mathbf{x}) = \frac{e^{-E\left(\mathbf{x};\theta\right)}}{Z\left(\theta\right)}
\end{equation}
is in general very difficult, since it requires working with the frequently intractable partition function integral $Z(\theta) = \int{e^{-E(\mathbf{x};\theta)}d\mathbf{x}}$.  There are a number of techniques which can provide approximate learning gradients (eg minimum probability flow \cite{MPF_ICML,SohlDickstein2011a}
, contrastive divergence \cite{Welling:2002p3,Hinton02}, score matching \cite{Hyvarinen05}, mean field theory, and variational bayes \cite{Tanaka:1998p1984,Kappen:1997p6,Jaakkola:1997p4985,haykin2008nnc}).  Turning those gradients into natural gradients is difficult though, as the Fisher information depends on the gradient of $\log Z\left( \theta \right)$.  Practically, simply ignoring the $\log Z\left( \theta \right)$ terms entirely and using a metric
\begin{eqnarray}
G_{ij}\left( \theta \right) = \left< 
	\frac
		{\partial E\left( \mb x; \theta \right)}{\partial \theta_i}
	\frac
		{\partial E\left( \mb x; \theta \right)}{\partial \theta_j}
	 \right>_{p\left( \mb x \right)}
\end{eqnarray}
averaged over the data distribution works surprisingly well, and frequently greatly accelerates learning.

\subsection{Diagonal approximation}

$\mb G\left( \theta \right)$ is a square matrix of size $N\times N$, where $N$ is the number of parameters in the vector $\theta$.  For problems with large $N$, $\mb G^{-1}\left( \theta \right)$ can be impractically expensive to compute and apply.  For almost all problems however, the natural gradient still improves convergence even when off-diagonal elements of $\mb G\left( \theta \right)$ are neglected,
\begin{eqnarray}
G_{ij}\left( \theta \right) = \delta_{ij} \left< \left(
	\frac
		{\partial \log q\left( \mb x; \theta \right)}{\partial \theta_i}
	\right)^2  \right>_{q\left( \mb x; \theta \right)}
,
\label{fisher diagonal}
\end{eqnarray}
making inversion and application cost $O\left( N \right)$ to perform.

If the parameters can be divided up into several distinct classes (for instance the covariance matrix and means of a gaussian distribution), block diagonal forms may also be worth considering.

\subsection{Regularization}

Even if evaluating the full $\mb G$ is easy for your problem, you may still find that $\mb G^{-1}$ is ill conditioned\footnote{This is a general problem when taking matrix inverses.  A matrix $\mb A$ with random elements, or with noisy elements, will tend to have a few very very small eigenvalues.  The eigenvalues of $\mb A^{-1}$ are the inverses of the eigenvalues of $\mb A$.  $\mb A^{-1}$ will thus tend to have a few very very large eigenvalues, which will tend to make the elements of $\mb A^{-1}$ very very large.  Even worse, the eigenvalues and eigenvectors which most dominate $\mb A^{-1}$ are those which were smallest, noisiest and least trustworthy in $\mb A$.}.  Dealing with this --- solving a set of linear equations subject to some regularization, rather than using an unstable matrix inverse --- is an entire field of study in computer science.  Here we give one simple plug and play technique, called stochastic robust approximation (Section 6.4.1 in \cite{boyd2004convex}), for regularizing the matrix inverse.  If $\mb G^{-1}$ is replaced with
\begin{equation}
\mb G^{-1}_{reg} =  \left( \mb G^T \mb G + \epsilon \I \right)^{-1}\mb G^T
\end{equation}
where $\epsilon$ is some small constant (say $0.01$), the matrix inverse will be much better behaved.

Alternatively, techniques such as ridge regression can be used to solve the linear equation
\begin{eqnarray}
\mb G\left( \theta \right) \tilde{\nabla}_\theta J\left( \theta \right) 
= 
\nabla_\theta J\left( \theta \right)
\end{eqnarray}
for $\tilde{\nabla}_\theta J\left( \theta \right)$.

\subsection{Combining the natural gradient with other techniques using the natural parameter space $\phi$ \label{nat space combine}}

It can be useful to combine the natural gradient with other gradient descent techniques.  Blindly replacing all gradients with natural gradients frequently causes problems (line search implementations, for instance, depend on the gradients they are passed being the true gradients of the function they are descending).  For a fixed value of $\mb G$ though there is a natural parameter space
\begin{eqnarray}
\phi = \mb G^{\frac{1}{2}}\left( \theta_{fixed} \right) \theta
\end{eqnarray}
in which the steepest gradient is the same as the natural gradient.

In order to easily combine the natural gradient with other gradient descent techniques, fix $\theta_{fixed}$ to the initial value of $\theta$ and perform gradient descent over $\phi$ using any preferred algorithm.  After a significant number of update steps convert back to $\theta$, update $\theta_{fixed}$ to the new value of $\theta$, and continue gradient descent in the new $\phi$ space.

\subsection{Natural gradient of non-probabilistic models \label{non probabilistic}}

The techniques presented here are not unique to probabilistic models.  The natural gradient can be used in any context where a suitable metric can be written for the parameters.  There are several approaches to writing an appropriate metric.

\begin{enumerate}

\item If the objective function is of a form
\begin{eqnarray}
J\left( \theta \right) = \left<l\left( \mb x; \theta \right)\right>_{p(x)}
\end{eqnarray}
where $\left< \cdot \right>_{p(x)}$ indicates averaging over some data distribution $p(x)$, then it is sensible to choose a metric based on
\begin{eqnarray}
G_{ij}\left( \theta \right) & = & \left< 
	\frac
		{\partial l\left( \mb x; \theta \right)}{\partial \theta_i}
	\frac
		{\partial l\left( \mb x; \theta \right)}{\partial \theta_j}
	 \right>_{p\left( \mb x \right)}
\end{eqnarray}

\item Similarly, the penalty function $l\left( \mb x; \theta \right)$ can be treated as if it is the log likelihood of a probabilistic model, and the corresponding Fisher information matrix used.

For example, the task of minimizing an L2 penalty function $\left|\left| \mb y - \mb f\left(\mb x; \theta \right) \right|\right|^2$ over observed pairs of data $p\left( \mb x, \mb y \right)$ can be made probabilistic.  Imagine that the L2 penalty instead represents a conditional gaussian $q\left( \mb y | \mb x ;\theta \right) \propto \exp \left( - \left|\left| \mb y - \mb f\left(\mb x; \theta \right) \right|\right|^2 \right)$ over $\mb y$, and use the observed marginal $p\left( \mb x \right)$ over $\mb x$ to build a joint distribution $q\left( \mb x, \mb y ; \theta \right) = q\left( \mb y | \mb x ;\theta \right) p\left( \mb x \right)$.\footnote{Amari \cite{Amari1998} suggests using some uninformative model distribution $q\left( \mb x \right)$ over the inputs, such as a gaussian distribution, rather than taking $p\left( \mb x \right)$ from the data.  Either approach will likely work well.
}  This generates the metric:
\begin{eqnarray}
G_{ij}\left( \theta \right) & = & \left< 
	\frac
		{\partial \log \left[ q\left( \mb y|\mb x; \theta \right) p\left( \mb x \right) \right]}{\partial \theta_i}
	\frac
		{\partial \log \left[ q\left( \mb y|\mb x; \theta \right) p\left( \mb x \right) \right]}{\partial \theta_j}
	 \right>_{q\left( \mb y|\mb x; \theta \right) p\left( \mb x \right)} \\
 & = & \left< 
	\frac
		{\partial \log q\left( \mb y|\mb x; \theta \right) }{\partial \theta_i}
	\frac
		{\partial \log q\left( \mb y|\mb x; \theta \right) }{\partial \theta_j}
	 \right>_{q\left( \mb y|\mb x; \theta \right) p\left( \mb x \right)}
\end{eqnarray}

\item Find a set of parameter transformations $T\left( \theta \right)$ which you believe the distance measure $\left| d \theta \right|$ should be invariant to, and then find a metric $\mb G\left( \theta \right)$ such that this invariance holds.  That is find $\mb G\left( \theta \right)$ such that the following relationship holds for any invariant transformation $T\left( \theta \right)$,
\begin{eqnarray}
\left| \left(\theta + d\theta\right) - \theta \right|^2 & = \left| T\left(\theta + d\theta\right) - T\left( \theta \right) \right|^2
.
\end{eqnarray}

A special case of this approach involves functions parametrized by a matrix, as presented in the next section.
\end{enumerate}

\subsection{$\mb W^T\mb W$}

As derived in \cite{Amari1998}, if a function depends on a (square, non-singular) matrix $\mb W$, it frequently aids learning a great deal to take

\begin{equation}
\tilde{\Delta} \mb W_{nat} \propto \frac
	{\partial J\left( \mb W \right) }
	{\partial \mb W} 
	\mb W^T \mb W
.
\end{equation}
The algebra leading to this rule is complex, but as discussed in the previous section it falls out of a demand that the distance measure $\left| d \mb W \right|$ be invariant to a set of transformations applied to $\mb W$.  In this case, those transformations are right multiplication by any (non-singular) matrix $\mb Y$.
\begin{eqnarray}
d \theta^T \mb G\left( \theta \right) d \theta = \left( d \theta Y \right)^T \mb G\left( \theta Y \right) \left( d \theta Y \right)
\end{eqnarray}

\subsection{What if my approximation of ${\Delta \theta}_{nat}$ is wrong?} \label{sec fail}

For any positive definite $\mb H$, movement in a direction
\begin{eqnarray}
\tilde{\Delta} \theta = \mb H \Delta \theta
\end{eqnarray}
will descend the objective function.  If the wrong $\mb H$ is used, gradient descent is performed in a suboptimal way \ldots which is the problem when steepest gradient descent is used as well.  Making an educated guess as to $\mb H$ rarely makes things worse, and frequently helps a great deal.  

\chapter{Hamiltonian Annealed Importance Sampling for Partition Function Estimation}
\label{chap HAIS}

In this chapter we introduce an extension to Annealed Importance Sampling (AIS) that uses Hamiltonian dynamics to rapidly estimate normalization constants. We demonstrate this method by computing log likelihoods in directed and undirected probabilistic image models. We compare the performance of linear generative models with both Gaussian and Laplace priors, product of experts models with Laplace and Student's t experts, the mc-RBM, and a bilinear generative model.  Matlab code implementing the estimation technique presented in this chapter is available at \cite{HAIScode}.  Material in this chapter is taken from \cite{HAIS}.  AIS is introduced in Section \ref{sec AIS intro}.

\section{Introduction}

We would like to use probabilistic models to assign probabilities to data. Unfortunately, this innocuous statement belies an important, difficult problem: many interesting distributions used widely across sciences cannot be analytically normalized.
Historically, the training of probabilistic models has been motivated in terms of maximizing the log probability of the data under the model or minimizing the KL divergence between the data and the model.  
However, for most models it is impossible to directly compute the log likelihood, due to the intractability of the  normalization constant, or partition function.
For this reason, performance is typically measured using a variety of diagnostic heuristics, not directly indicative of log likelihood.  For example, image models are often compared in terms of their synthesis, denoising, inpainting, and classification performance.
This inability to directly measure the log likelihood has made it difficult to consistently evaluate and compare models.


Recently, a growing number of researchers have given their attention to measures of likelihood in image models.
\cite{Salakhutdinov:2008p12541} use annealed importance sampling, and \cite{Murray:2009p12430} use a hybrid of annealed importance sampling and a Chib-style estimator to estimate the log likelihood of a variety of MNIST digits and natural image patches modeled using restricted Boltzmann machines and deep belief networks.
\cite{Bethge:2006p12046} measures the reduction in multi-information, or statistical redundancy, as images undergo various complete linear transformations.  \cite{Chandler:2007p12190} and \cite{Stephens:2008p12575} produce estimates of the entropy inherent in natural scenes, but do not address model evaluation. \cite{Karklin:2007p12497} uses kernel density estimates -- essentially, vector quantization -- to compare different image models, though that technique suffers from severe scaling problems except in specific contexts.
\cite{Zoran:2009p12559} compare the true log likelihoods of a number of image models, but restricts their analysis to the rare cases where the partition function can be solved analytically.

In this work, we merge two existing ideas -- annealed importance sampling (see Section \ref{sec AIS intro}) and Hamiltonian dynamics (see Section \ref{sec HMC intro} and Chapter \ref{chap HMC}) -- into a single algorithm.   The key insight that makes our algorithm more efficient than previous methods is our adaptation of AIS to work with Hamiltonian dynamics. As in HMC, we extend the state space to include auxiliary momentum variables; however, we do this in such a way that the momenta change consistently through the intermediate AIS distributions, rather than resetting them at the beginning of each Markov transition.
To make the practical applications of this work clear, we use our method, Hamiltonian Annealed Importance Sampling (HAIS), to measure the log likelihood of holdout data under a variety of directed (generative) and undirected (analysis/feed-forward) probabilistic models of natural image patches.

\section{Estimating Log Likelihood}

\subsection{Hamiltonian Annealed Importance Sampling}

Hamiltonian Monte Carlo \cite{Neal:HMC} uses an analogy to the physical dynamics of particles moving with momentum under the influence of an energy function to propose Markov chain transitions which rapidly explore the state space.  It does this by expanding the state space to include auxiliary momentum variables, and then simulating Hamiltonian dynamics to move long distances along iso-probability contours in the expanded state space. A similar technique is powerful in the context of annealed importance sampling. Additionally, by retaining the momenta variables across the intermediate distributions, significant momentum can build up as the proposal distribution is transformed into the target. This provides a mixing benefit that is unique to our formulation.

The state space $\mb X$ is first extended to $\mb Y = \left\{\mb y_{1}, \mb y_{2} \ldots \mb y_{N} \right\}$, $\mb y_{n} = \left\{ \mb x_{n}, \mb v_{n} \right\}$, where $\mb v_{n} \in \mathbb R^M$ consists of a momentum associated with each position $\mb x_{n}$.  The momenta associated with both the proposal and target distributions is taken to be unit norm isotropic gaussian.
The proposal and target distributions $q\lp \mb x \rp$ and $p\lp \mb x \rp$ are extended to corresponding distributions $q_\cup\lp \mb y \rp$ and $p_\cup\lp \mb y \rp$ over position and momentum $\mb y = \left\{ \mb x, \mb v \right\}$,
\begin{align}
p_\cup\lp \mb y \rp &= p\lp \mb x \rp \ \Phi\lp \mb v \rp = \frac
{e^{-E_{p_\cup}\lp \mb y \rp}}
{Z_{p_\cup}}
\\
q_\cup\lp \mb y \rp &= q\lp \mb x \rp \  \Phi\lp \mb v \rp = \frac
{e^{-E_{ q_\cup}\lp \mb y \rp}}
{Z_{q_\cup}}
\\
\displaybreak[0]
\Phi\lp \mb v \rp & = \frac
{e^{
   -\frac{1}{2} \mb v^T \mb v
}}
{\lp2\pi\rp^{\frac{M}{2}}}  
\\
\displaybreak[0]
E_{p_\cup}\lp \mb y \rp &=
E_p\lp \mb x \rp + \frac{1}{2} \mb v^T \mb v
\\
E_{q_\cup}\lp \mb y \rp &=
E_q\lp \mb x \rp + \frac{1}{2} \mb v^T \mb v
.
\end{align}
The remaining distributions are extended to cover both position and momentum in a nearly identical fashion: the forward and reverse chains $Q\lp \mb X \rp \rightarrow  Q_\cup\lp \mb Y \rp$, $P\lp \mb X \rp\rightarrow  P_\cup\lp \mb Y \rp$, the intermediate distributions and energy functions $\pi_{n}\lp \mb x \rp \rightarrow  \pi_{\cup \ n}\lp \mb y \rp$,  $E_{\pi_{n}}\lp \mb x \rp \rightarrow E_{\pi_{\cup \ n}}\lp \mb y \rp$,
\begin{align}
E_{\pi_{\cup \ n}}\lp \mb y \rp & = \lp1-\beta_{n}\rp E_{q_\cup}\lp \mb y \rp + \beta_{n} E_{p_\cup}\lp \mb y \rp
\\ &=
  \lp1-\beta_{n}\rp E_{q}\lp \mb x \rp + \beta_{n} E_{p}\lp \mb x \rp + \frac{1}{2} \mb v^T \mb v
,
\end{align}
and the forward and reverse Markov transition distributions $\gij{n}{n+1} \rightarrow \ygij{n}{n+1}$ and $\gijr{n+1}{n}\rightarrow \ygijr{n+1}{n}$.  Similarly, the samples $\mc S_{Q_\cup}$ now each have both position $\mb X$ and momentum $\mb V$, and are drawn from the forward chain described by $Q_\cup\lp \mb Y \rp$.

The annealed importance sampling estimate $\hat Z_p$ given in Equation \ref{eq AIS} remains {\em unchanged}, except for a replacement of $\mc S_{Q}$ with $\mc S_{Q_\cup}$ -- all the terms involving the momentum $\mb V$ conveniently cancel out, since the same momentum distribution $\Phi\lp \mb v \rp$ is used for the proposal $q_\cup\lp \mb y_{1} \rp$ and target $p_\cup\lp \mb y_{N} \rp$,
\begin{align}
\hat Z_p & =     \frac{1}{\left| \mc S_{Q_\cup} \right|}\sum_{Y \in \mc S_{Q_\cup}}
   \frac
      {e^{-E_p\lp \mb x_{N} \rp}    \Phi\lp \mb v_N \rp }
      {q\lp \mb x_{1} \rp                  \Phi\lp \mb v_1 \rp}
   \frac{e^{
      -\Ept{1}{1} + \frac{1}{2} \mb v_1^T \mb v_1
   }}
   {e^{
      - \Ept{1}{2} + \frac{1}{2} \mb v_2^T \mb v_2
   }}
   \cdots
   \frac{e^{
      -\Ept{N-1}{N-1} + \frac{1}{2} \mb v_{N-1}^T \mb v_{N-1}
   }}
   {e^{
      - \Ept{N-1}{N} + \frac{1}{2} \mb v_N^T \mb v_N
   }}
\\ & =
\frac{1}{\left| \mc S_{Q_\cup} \right|}\sum_{Y \in \mc S_{Q_\cup}}
   \frac
      {e^{-E_p\lp \mb x_{N} \rp}}
      {q\lp \mb x_{1} \rp}
   \frac{e^{
      -\Ept{1}{1}
   }}
   {e^{
      - \Ept{1}{2}
   }}
   \cdots
   \frac{e^{
      -\Ept{N-1}{N-1}
   }}
   {e^{
      - \Ept{N-1}{N}
   }}
                      \label{eq HAIS}
.
\end{align}
Thus, the momentum only matters when generating the samples $\mc S_{Q_\cup}$, by drawing from the initial proposal distribution $p_\cup\lp \mb y_{1} \rp$, and then applying the series of Markov transitions $\ygij{n}{n+1}$.

For the transition distributions, $\ygij{n}{n+1}$, we propose a new location by integrating Hamiltonian dynamics for a short time using a single leapfrog step, accept or reject the new location via Metropolis rules, and then partially corrupt the momentum.  That is, we generate a sample from $\ygij{n}{n+1}$ by following the procedure:
\begin{enumerate}
\item $\left\{ \mb x_{H}^0, \mb v_{H}^0 \right\} = \left\{ \mb x_{n}, \mb v_{n} \right\}$
\item leapfrog: \parbox[t]{3.2in}{$\mb x_{H}^\frac{1}{2} = \mb x_{H}^0 + \frac{\epsilon}{2} \mb v_{H}^0$ \vspace{3mm} \\
$\mb v_{H}^1 = \mb v_{H}^0 - \left. \epsilon 
	\pd{E_{\pi_{n}}\lp \mb x \rp}{\mb x}
\right|_{\mb x = \mb x_{H}^\frac{1}{2}}$ \\
$\mb x_{H}^1 = \mb x_{H}^\frac{1}{2} + \frac{\epsilon}{2} \mb v_{H}^1$} \vspace{2mm} \\
where the step size $\epsilon = 0.2$ for all experiments in this paper.
\item accept/reject: $\left\{ \mb x', \mb v' \right\} = \left\{ \mb x_{H}^1, -\mb v_{H}^1 \right\}$ with probability
$P_{accept} = \min\left[
   1,
   \frac{
    e^{
       -E_{\pi_{n}}\lp \mb x_{H}^1 \rp
       -\frac{1}{2}{\mb v_{H}^1}^T{\mb v_{H}^1}
   } }
   {
   e^{
       -E_{\pi_{n}}\lp \mb x_{H}^0 \rp
       -\frac{1}{2}{\mb v_{H}^0}^T{\mb v_{H}^0}
   } } \right]$, otherwise $\left\{ \mb x', \mb v' \right\} = \left\{ \mb x_{H}^0, \mb v_{H}^0 \right\}$
\item partial momentum refresh: $\tilde{\mb v}' = -\sqrt{1 - \gamma}\mb v' + \gamma \mb r$, where $r \sim \mc N\lp 0, \mb I \rp$, and $\gamma \in \lp 0, 1 \right]$ is chosen so as to randomize half the momentum power per unit simulation time \cite{Culpepper2011}.
\item $\mb y_{n+1} = \left\{ \mb x_{n+1}, \mb v_{n+1} \right\} = \left\{ \mb x', \tilde{\mb v}' \right\}$
\end{enumerate}
This combines the advantages of many intermediate distributions, which can lower the variance in the estimated $\hat Z_p$, with the improved mixing which occurs when momentum is maintained over many update steps.  For details on Hamiltonian Monte Carlo sampling techniques, and a discussion of why the specific steps above leave $\pi_{n}\lp \mb x \rp$ invariant, we recommend \cite{Culpepper2011,Neal:HMC}.

Some of the models discussed below have linear constraints on their state spaces.  These are dealt with by negating the momentum $\mb v$ and reflecting the position $\mb x$ across the constraint boundary every time a leapfrog halfstep violates the constraint.

\subsection{Log Likelihood of Analysis Models}

Analysis models are defined for the purposes of this paper as those which have an easy to evaluate expression for $E_p\lp \mb x \rp$ when they are written in the form of Equation \ref{eq qx}.  The average log likelihood $\mc L$ of an analysis model $p\lp \mb x \rp$ over a set of testing data $\mc D$ is
\begin{align}
\mc L = \frac{1}{\left| \mc D \right|} \sum_{\mb x \in \mc D} \log p\lp \mb x \rp = -\frac{1}{\left| \mc D \right|} \sum_{\mb x \in \mc D} E_p\lp \mb x \rp - \log Z_p
\end{align}
where $\left| \mc D \right|$ is the number of samples in $\mc D$,
and the $Z_p$ in the second term can be directly estimated by Hamiltonian annealed importance sampling.

\subsection{Log Likelihood of Generative Models}

Generative models are defined here to be those which have a joint distribution,
\begin{align}
p\lp \mb x, \mb a \rp
& =
p\lp \mb x | \mb a \rp p\lp \mb a \rp
=
\frac
{e^{-E_{x|a}\lp \mb x, \mb a \rp }}
{Z_{x|a}}
\frac
{e^{- E_{a}\lp \mb a \rp }}
{Z_a}
,
\end{align}
over visible variables $\mb x$ and auxiliary variables $\mb a \in \mathbb R^L$ which is easy to exactly evaluate and sample from, but for which the marginal distribution over the visible variables $p\lp \mb x \rp = \int d\mb a\ p\lp \mb x, \mb a \rp$ is intractable to compute.  The average log likelihood $\mc L$ of a model of this form over a testing set $\mc D$ is
\begin{align}
\mc L &= \frac{1}{\left| \mc D \right|} \sum_{\mb x \in \mc D} \log Z_{a|x} \\
Z_{a|x} &= \int d\mb a\ e^{    
-E_{x|a}\lp \mb x, \mb a \rp
-\log Z_{x|a}
- E_{a}\lp \mb a \rp
- \log Z_a
}
,
\end{align}
where each of the $Z_{a|x}$ can be estimated using HAIS.  Generative models take significantly longer to evaluate than analysis models, as a separate HAIS chain must be run for each test sample.

\section{Models} \label{sec models}

\begin{figure}[t]
\vskip 0.2in
\begin{center}
\includegraphics[width= 0.92\columnwidth]{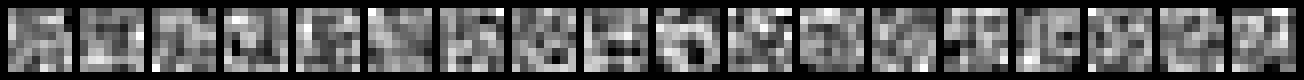}  (a) \\
\includegraphics[width= 0.92\columnwidth]{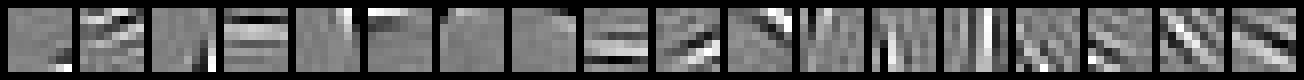}  (b) \\
\includegraphics[width= 0.92\columnwidth]{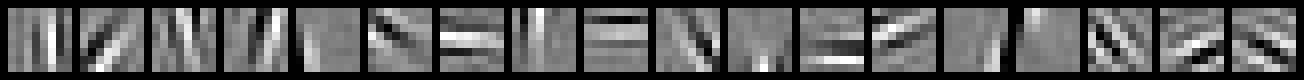}  (c) \\
\includegraphics[width= 0.92\columnwidth]{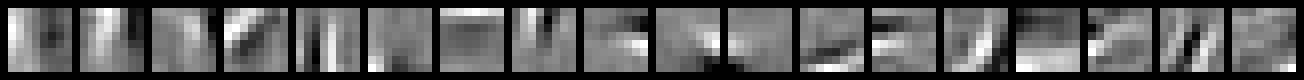} (d) \\
\includegraphics[width= 0.92\columnwidth]{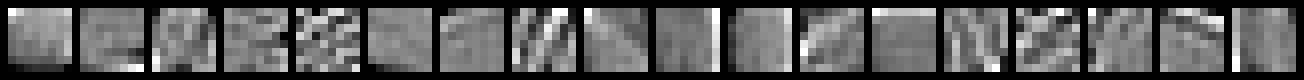} (e) \\
\includegraphics[width= 0.92\columnwidth]{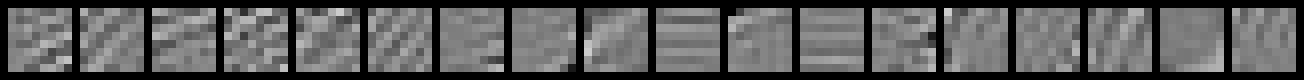} (f) \\
\includegraphics[width= 0.92\columnwidth]{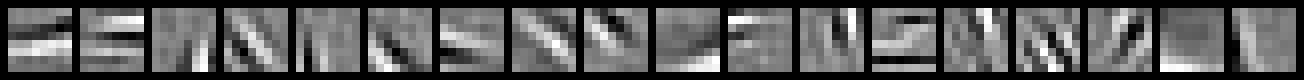} (g) \\
\includegraphics[width= 0.92\columnwidth]{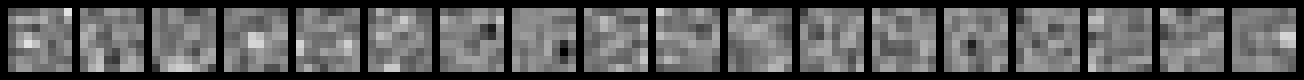} (h) \\
\includegraphics[width= 0.92\columnwidth]{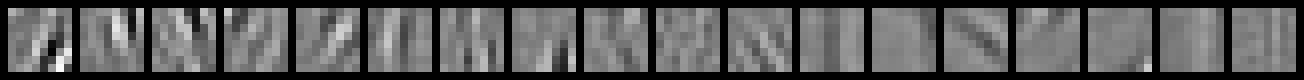} (i) \\
\caption{A subset of the basis functions and filters learned by each model. {\em (a)} Bases $\Phi$ for the linear generative model with Gaussian prior and {\em (b)} Laplace prior; {\em (c)} filters $\Phi$ for the product of experts model with Laplace experts, and {\em (d)} Student's t experts; {\em (e)} Bases $\Phi$ for the bilinear generative model and {\em (f)} the basis elements making up a single grouping from $\Psi$, ordered by and contrast modulated according to the strength of the corresponding $\Psi$ weight (decreasing from left to right); mcRBM {\em (g)} $C$ filters, {\em (h)} $W$ means, and {\em (i)} a single $P$ grouping, showing the pooled filters from $C$, ordered by and contrast modulated according to the strength of the corresponding $P$ weight (decreasing from left to right). }
\label{fig:bf}
\end{center}
\vskip -0.2in
\end{figure}

The probabilistic forms for all models whose log likelihood we evaluate are given below.  In all cases, $\mb x \in \mathbb R^M$ refers to the data vector.
\begin{enumerate}
\item linear generative:
\begin{align}
p\lp \mb x | \mb a \rp & = \frac{
   \exp\left[
       -\frac{1}{2\sigma_n^2} {\lp \mb x - \Phi \mb a \rp}^T {\lp \mb x - \Phi \mb a \rp}
   \right]
}{
   \lp 2 \pi \rp^{\frac{M}{2}} \sigma_n^M
}
\label{eq lin gen}
\end{align}
parameters: $\Phi \in \mathbb R^{M \times L}$\\
auxiliary variables: $\mb a \in \mathbb R^L$ \\
constant: $\sigma_n = 0.1$ \\
Linear generative models were tested with a two priors, as listed:
\begin{enumerate}
\item Gaussian prior: \label{sec lin gauss}
\begin{align}
p\lp \mb a \rp & = \frac{
   \exp\left[
       -\frac{1}{2} {\mb a}^T {\mb a}
   \right]
}
{
\lp 2 \pi \rp^{\frac{L}{2}}
}
\end{align}
\item Laplace prior \cite{Olshausen1997a}:
\begin{align}
p\lp \mb a \rp & = \frac{
   \exp\left[
       -\left| \left| \mb a \right| \right|_1^1
   \right]
}
{
2
}
\end{align}
\end{enumerate}
\item bilinear generative \cite{Culpepper2011}: \label{mod bilin} The form is the same as for the linear generative model, but with the coefficients $\mb a$ decomposed into two multiplicative factors, one of which is positive only, 
\begin{align}
\mb a &= \lp\Theta \mb c\rp \odot \lp\Psi \mb d\rp
\\
p\lp \mb c \rp & = \frac{
   \exp\left[
       -\left| \left| \mb c \right| \right|_1^1
   \right]
}
{
2
} \\
p\lp \mb d \rp & =
   \exp\left[
       -\left| \left| \mb d \right| \right|_1^1
   \right]
,
\end{align}
where $\odot$ indicates element-wise multiplication. \\
parameters: $\Phi \in \mathbb R^{M \times L}$, $\Theta \in \mathbb R^{L \times K_c}$, $\Psi \in \mathbb R^{L \times K_d}$ \\
auxiliary variables: $\mb c \in \mathbb R^{K_c}$, $\mb d \in \mathbb R_+^{K_d}$
\item product of experts \cite{Hinton02}: This is the analysis model analogue of the linear generative model,  \label{sec POE}
\begin{align}
p\lp \mb x \rp & = \frac{1}{Z_{POE}}
\prod_{l=1}^L \exp\lp -E_{POE}\lp \Phi_l \mb x; \lambda_l  \rp \rp
.
\end{align}
parameters: $\Phi \in \mathbb R^{L \times M}$, $\lambda \in \mathbb R_+^L$,\\
Product of experts models were tested with two experts, as listed:
\begin{enumerate}
\item Laplace expert:
\begin{align}
E_{POE}\lp u; \lambda_l  \rp = \lambda_l \left| u \right|
\end{align}
(changing $\lambda_l$ is equivalent to changing the length of the row $\Phi_l$, so it is fixed to $\lambda_l = 1$)
\item Student's t expert:
\begin{align}
E_{POE}\lp u; \lambda_l \rp = \lambda_l \log\lp 1 + u^2 \rp
\end{align}
\end{enumerate}
\item Mean and covariance restricted Boltzmann machine (mcRBM) \cite{Ranzato:2010p12217}:  This is an analysis model analogue of the bilinear generative model.  The exact marginal energy function $E_{mcR}$ is taken from the released code rather than the paper.
\begin{align}
p\lp \mb x \rp & = \frac{
\exp\left[ - E_{mcR}\lp \mb x \rp \right]
}{Z_{mcR}} \\
E_{mcR}\lp \mb x \rp & = -\sum_{k=1}^K  \log \left[ 1 + e^{
                        \frac{1}{2}
                            \sum_{l=1}^L P_{lk} \frac{
                                \left( \mb C_l \mb x \right)^2
                            }{
                                \left| \left| \mb x \right|\right|_2^2 + \frac{1}{2}
                            }
                            + b^c_k`
                        } \right]
\nonumber \\ & \qquad
                     -\sum_{j=1}^J \log \left[ 1 + e^{
                        \mb W_j \mb x + b^m_j
                        } \right]
\nonumber \\ & \qquad
                    +\frac{1}{2\sigma^2} \mb x^T \mb x
                    -\mb x^T \mb b^v
\end{align}
parameters: $P \in \mathbb R^{L \times K}$, $C \in \mathbb R^{L \times M}$, $W \in \mathbb R^{J \times M}$, $b^m \in \mathbb R^J$, $b^c \in \mathbb R^K$, $b^v \in \mathbb R^K$, $\sigma \in \mathbb R$
\end{enumerate}

\section{Training}

All models were trained on 10,000 $16x16$ pixel image patches taken at random from 4,112 linearized images of natural scenes from the van Hateren dataset \cite{hateren_schaaf_1998}.  The extracted image patches were first logged, and then mean subtracted.  They were then projected onto the top $M$ PCA components, and whitened by rescaling each dimension to unit norm.

All generative models were trained using Expectation Maximization over the full training set, with a Hamiltonian Monte Carlo algorithm used during the expectation step to maintain samples from the posterior distribution.  See \cite{Culpepper2011} for details.  All analysis models were trained using LBFGS on the minimum probability flow learning objective function for the full training set, with a transition function $\Gamma$ based on Hamiltonian dynamics.  See \cite{MPF_ICML} for details.  
No regularization or decay terms were required on any of the model parameters. 

\section{Results}
\begin{figure}[t]
\vskip 0.2in
\begin{center}
\centerline{\includegraphics[width= 0.9\columnwidth]{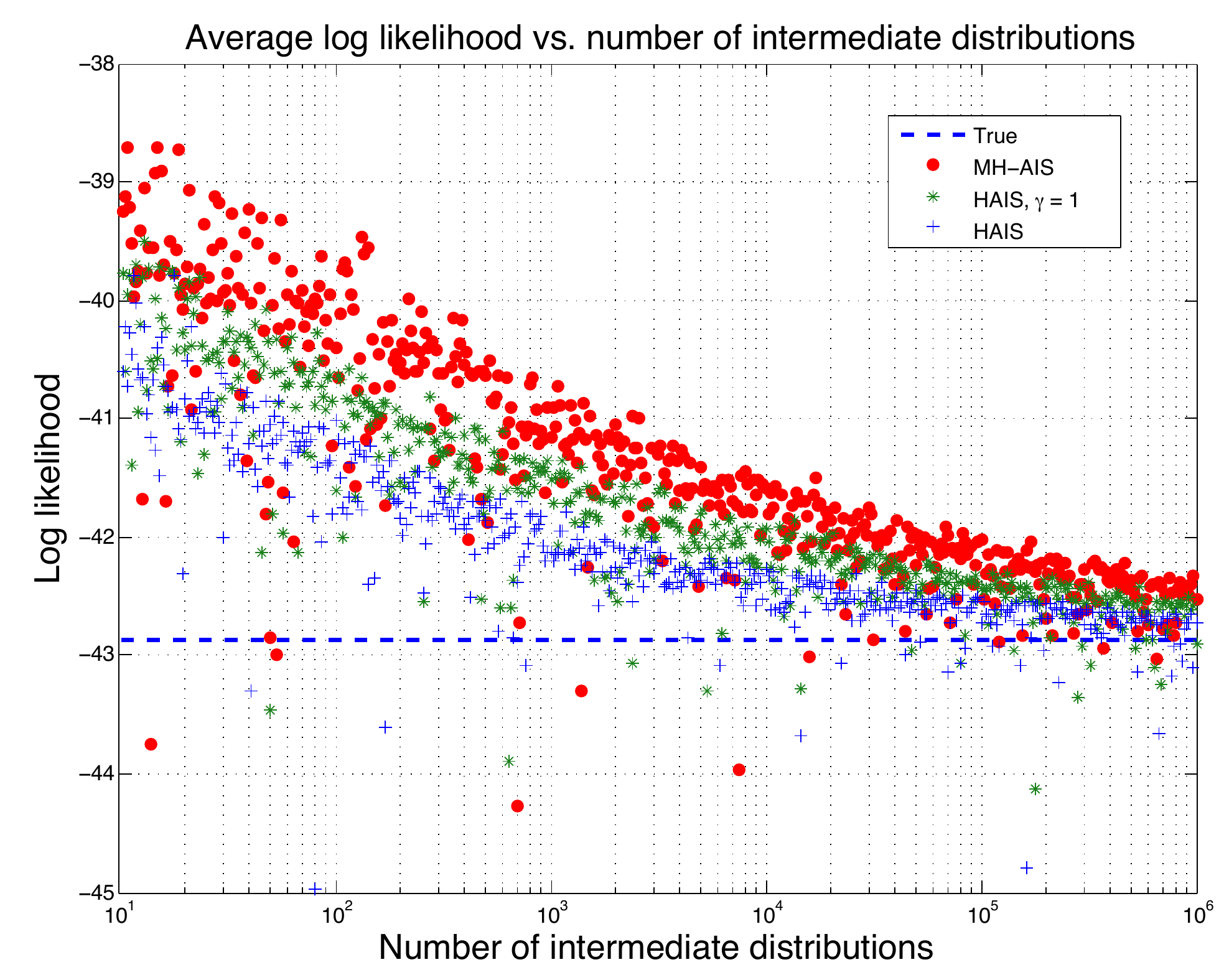}}
\caption{Comparison of HAIS with alternate AIS algorithms in a complete ($M=L=36$) POE Student's t model.  The scatter plot shows estimated log likelihoods under the test data for the POE model for different numbers of intermediate distributions $N$. The \textcolor{blue}{blue crosses} indicate HAIS.  The \textcolor{green}{green stars} indicate AIS with a single Hamiltonian dynamics leapfrog step per distribution, but no continuity of momentum. The \textcolor{red}{red dots} indicate AIS with a Gaussian proposal distribution.  The dashed \textcolor{blue}{blue line} indicates the true log likelihood of the minimum probability flow trained model.  
This product of Student's t distribution is extremely difficult to normalize numerically, as many of its moments are infinite.
}
\label{fig:converge-poe-studentt}
\end{center}
\vskip -0.2in
\end{figure}

100 images from the van Hateren dataset were chosen at random and reserved as a test set for evaluation of log likelihood.  The test data was preprocessed in an identical fashion to the training data.  Unless otherwise noted, log likelihood is estimated on the same set of 100 patches drawn from the test images, using Hamiltonian annealed importance sampling with $N=100,000$ intermediate distributions, and 200 particles.  This procedure takes about 170 seconds for the 36 PCA component analysis models tested below.  The generative models take approximately 4 hours, because models with unmarginalized auxiliary variables require one full HAIS run for each test datapoint.

\subsection{Validating Hamiltonian Annealed Importance Sampling}

\begin{figure}[t]
\vskip 0.2in
\begin{center}
\centerline{\includegraphics[width= 0.9\columnwidth]{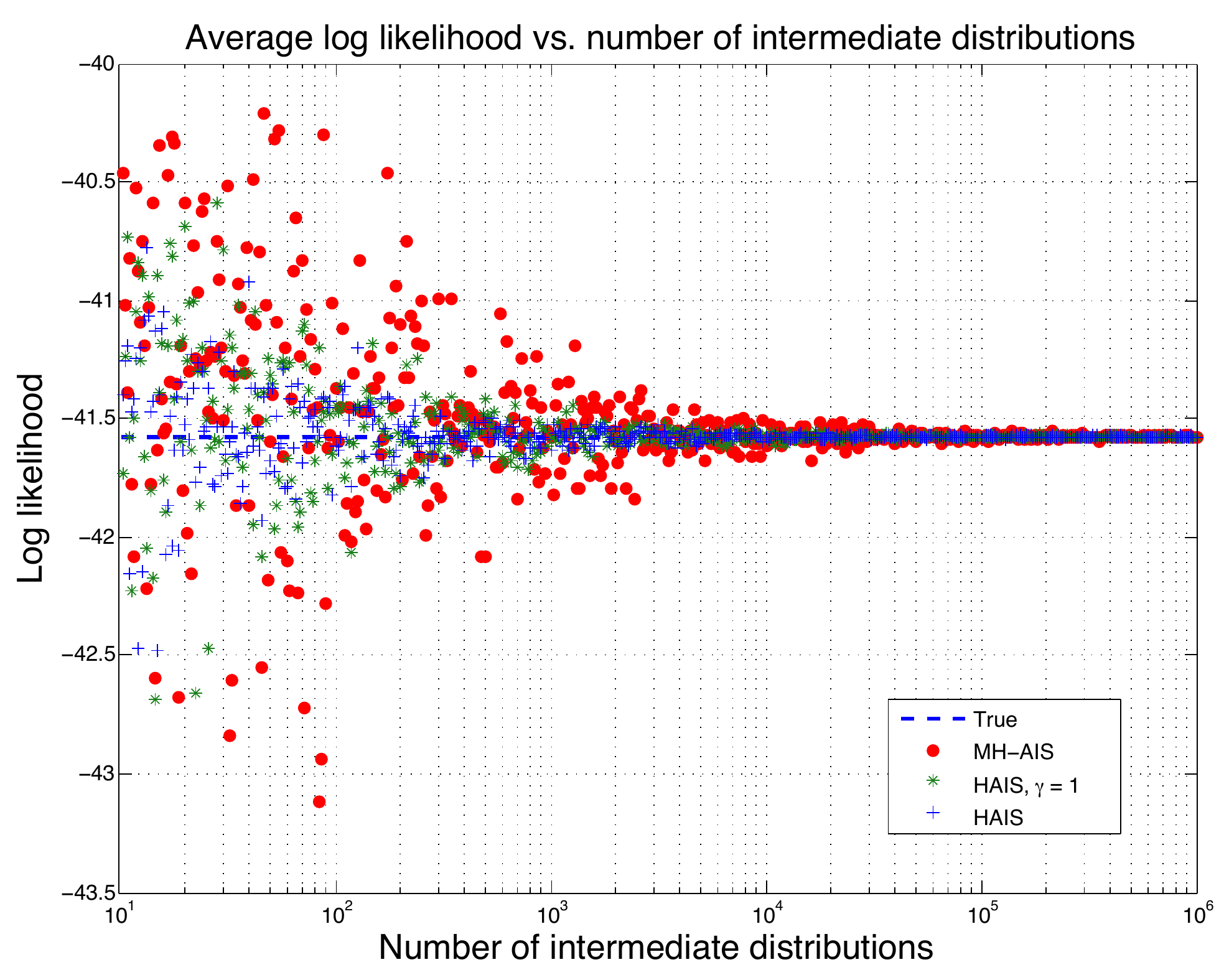}}
\caption{Comparison of HAIS with alternate AIS algorithms in a complete ($M=L=36$) POE Laplace model.  Format as in Figure \ref{fig:converge-poe-studentt}, but for a Laplace expert.
}
\label{fig:converge-poe-laplace}
\end{center}
\vskip -0.2in
\end{figure}

\begin{figure}[t]
\vskip 0.2in
\begin{center}
\centerline{\includegraphics[width= 0.9\columnwidth]{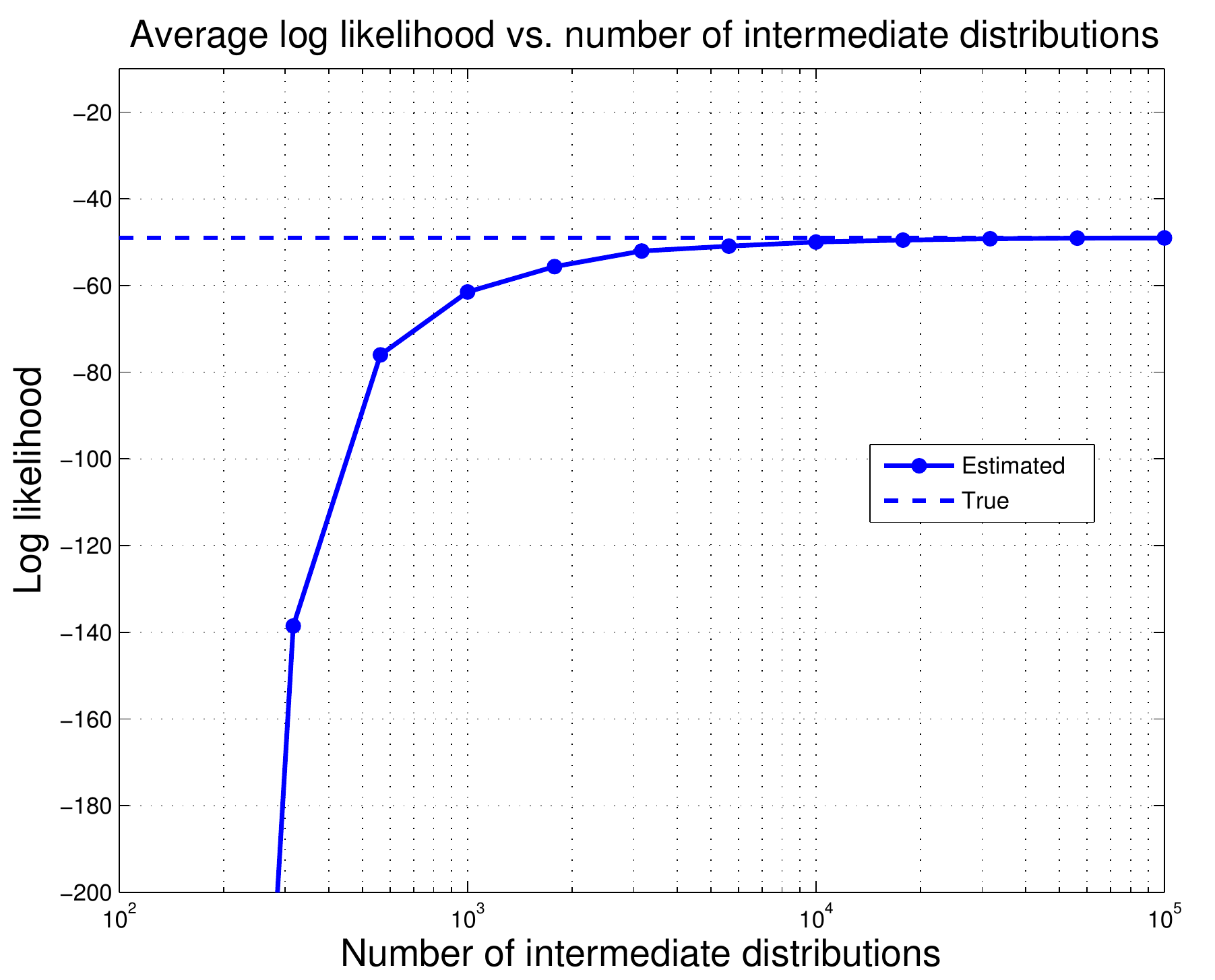}}
\caption{Convergence of HAIS for the linear generative model with a Gaussian prior. The dashed blue line indicates the true log likelihood of the test data under the model. The solid blue line indicates the HAIS estimated log likelihood of the test data for different numbers of intermediate distributions $N$.}
\label{fig:converge-linear-gauss}
\end{center}
\vskip -0.2in
\end{figure}

\begin{figure}[t]
\vskip 0.2in
\begin{center}
\centerline{\includegraphics[width= 0.9\columnwidth]{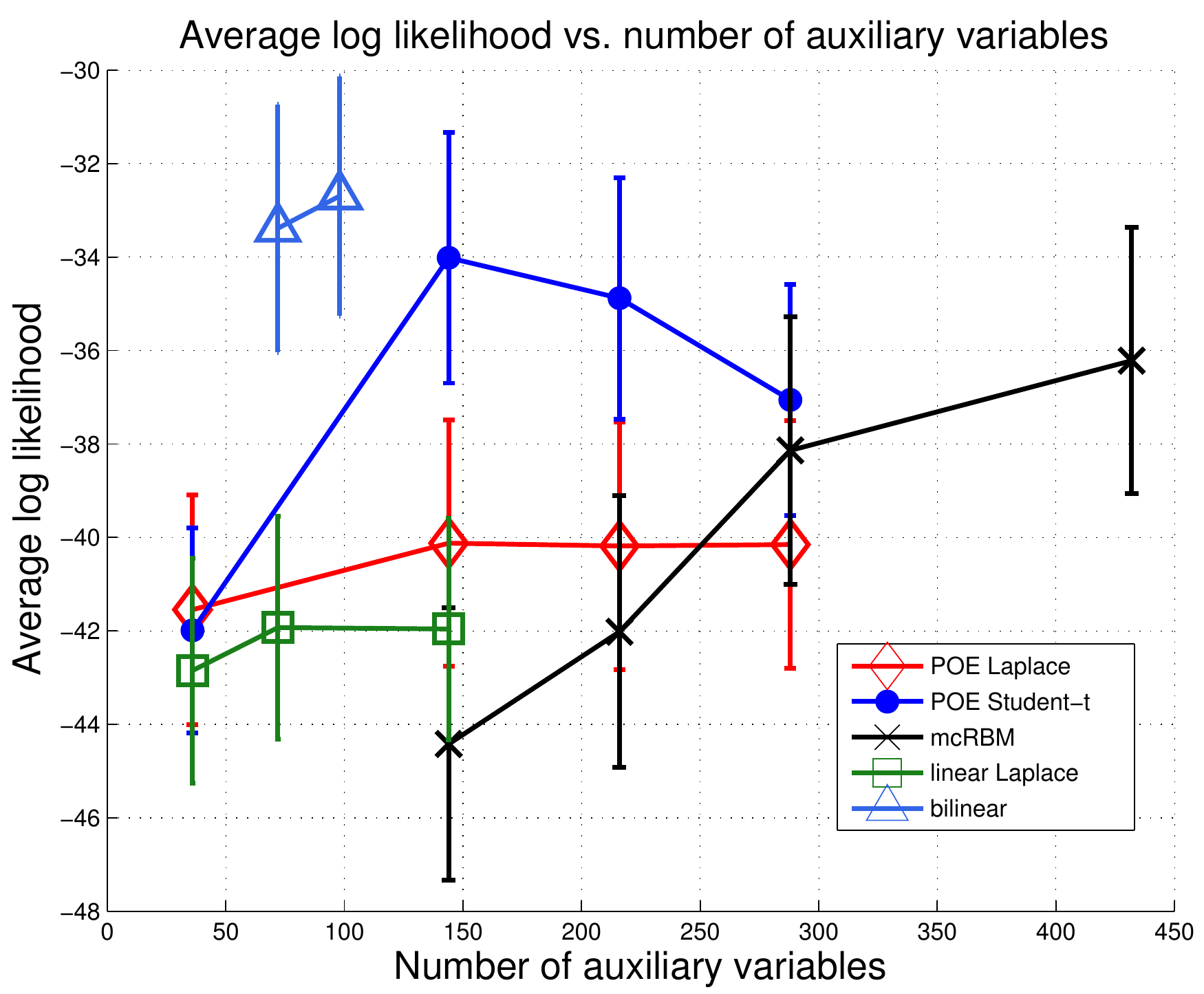}}
\caption{Increasing the number of auxiliary variables in a model increases the likelihood it assigns to the test data until it saturates, or overfits. }
\label{fig:fit}
\end{center}
\vskip -0.2in
\end{figure}

\begin{table}[t]
\caption{Average log likelihood for the test data under each of the models. The model `size' column denotes the number of experts in the POE models, the sum of the mean and covariance units for the mcRBM, and the total number of latent variables in the generative models.}
\label{table:model comparison}
\vskip 0.15in
\begin{center}
\begin{small}
\begin{sc}
\begin{tabular}{lrl}
\hline
Model & Size & Log Likelihood \\
\hline
Lin. generative, Gaussian & 36 & -49.15$\pm$ 2.31 \\
Lin. generative, Laplace & 36 & -42.85$\pm$ 2.41 \\
POE, Laplace experts & 144 & -41.54$\pm$ 2.46 \\
mcRBM & 432 & -36.01$\pm$ 2.57 \\
POE, Student's t experts & 144 &  -34.01$\pm$ 2.68 \\
Bilinear generative & 98 & -32.69$\pm$ 2.56 \\
\hline
\end{tabular}
\end{sc}
\end{small}
\end{center}
\vskip -0.1in
\end{table}

The log likelihood of the test data can be analytically computed for three of the models outlined above: linear generative with Gaussian prior (Section \ref{sec models}, model \ref{sec lin gauss}), and product of experts with a complete representation ($M = L$) for both Laplace and Student's t experts (Section \ref{sec models}, model \ref{sec POE}).  Figures \ref{fig:converge-poe-studentt}, \ref{fig:converge-poe-laplace} and \ref{fig:converge-linear-gauss} show the convergence of Hamiltonian annealed importance sampling, with 200 particles, for each of these three models as a function of the number $N$ of intermediate distributions.  Note that the Student's t expert is a pathological case for sampling based techniques, as for several of the learned $\lambda_l$ even the first moment of the Student's t-distribution was infinite.

Additionally, for all of the generative models, if $\mb \Phi = \mb 0$ then the statistical model reduces to,
\begin{align}
p\lp \mb x | \mb a \rp & = \frac{
   \exp\left[
       -\frac{1}{2\sigma_n^2} {\mb x}^T {\mb x}
   \right]
}{
   \lp 2 \pi \rp^{\frac{M}{2}} \sigma_n^M
} \, ,
\end{align}
and the log likelihood $\mathcal L$ has a simple form that can be used to directly verify the estimate computed via HAIS.  We performed this sanity check on all generative models, and found the HAIS estimated log likelihood converged to the true log likelihood in all cases.

\subsection{Speed of Convergence}

In order to demonstrate the improved performance of HAIS, we compare against two alternate AIS learning methods.  First, we compare to AIS with transition distributions $\gij{n}{n+1}$ consisting of a Gaussian ($\sigma_{diffusion} = 0.1$) proposal distribution and Metropolis-Hastings rejection rules.  Second, we compare to AIS with a single Hamiltonian leapfrog step per intermediate distribution $\pit{n}{n}$, and unit norm isotropic Gaussian momentum.  Unlike in HAIS however, in this case we randomize the momenta before each update step, rather than allowing them to remain consistent across intermediate transitions.  As can be seen in Figures \ref{fig:converge-poe-studentt} and \ref{fig:converge-poe-laplace}, HAIS requires fewer intermediate distributions by an order of magnitude or more.

\subsection{Model Size}

By training models of different sizes and then using HAIS to compute their likelihood, we are able to explore how each model behaves in this regard, and find that three have somewhat different characteristics, shown in Figure~\ref{fig:fit}. The POE model with a Laplace expert has relatively poor performance and we have no evidence that it is able to overfit the training data; in fact, due to the relatively weak sparsity of the Laplace prior, we tend to think the only thing it can learn is oriented, band-pass functions that more finely tile the space of orientation and frequency. In contrast, the Student-t expert model rises quickly to a high level of performance, then overfits dramatically. Surprisingly, the mcRBM performs poorly with a number of auxiliary variables that is comparable to the best performing POE model. One explanation for this is that we are testing it in a regime where the major structures designed into the model are not of great benefit. That is, the mcRBM is primarily good at capturing long range image structures, which are not sufficiently present in our data because we use only 36 PCA components. Although for computational reasons we do not yet have evidence that the mcRBM can overfit our dataset, it likely does have that power. We expect that it will fare better against other models as we scale up to more sizeable images. Finally, we are excited by the superior performance of the bilinear generative model, which outperforms all other models with only a small number of auxiliary variables. We suspect this is mainly due to the high degree of flexibility of the sparse prior, whose parameters (through $\Theta$ and $\Psi$) are learned from the data. The fact that for a comparable number of ``hidden units'' it outperforms the mcRBM, which can be thought of as the bilinear generative model's `analysis counterpart', highlights the power of this model.

\subsection{Comparing Model Classes}

As illustrated in Table \ref{table:model comparison}, we used HAIS to compute the log likelihood of the test data under each of the image models in Section \ref{sec models}.  The model sizes are indicated in the table -- for both POE models and the mcRBM they were chosen from the best performing datapoints in Figure \ref{fig:fit}.  In linear models, the use of sparse priors or experts leads to a large ($> 6\ nat$) increase in the log likelihood over a Gaussian model.  The choice of sparse prior was similarly important, with the POE model with Student's t experts performing more than $7\ nats$ better than the POE or generative model with Laplace prior or expert.  Although previous work \cite{Ranzato:2010p12217,Culpepper2011} has suggested bilinear models outperform their linear counterparts, our experiments show the Student's t POE performing within the noise of the more complex models.  One explanation is the relatively small dimensionality (36 PCA components) of the data -- the advantage of bilinear models over linear is expected to increase with dimensionality. 
Another is that Student's t POE models are in fact better than previously believed. Further investigation is underway.  The surprising performance of the Student's t POE, however, highlights the power and usefulness of being able to directly compare the log likelihoods of probabilistic models.
%

\section{Conclusion}

By improving upon the available methods for partition function estimation, we have made it possible to directly compare large probabilistic models in terms of the likelihoods they assign to data. This is a fundamental measure of the quality of a model -- especially a model trained in terms of log likelihood -- and one which is frequently neglected due to practical and computational limitations.  It is our hope that the Hamiltonian annealed importance sampling technique presented here will lead to better and more relevant empirical comparisons between models.

\chapter{Hamiltonian Monte Carlo}
\label{chap HMC}



Sampling is critical for many tasks involved in learning and working with probabilistic models.  As discussed in Section \ref{sec HMC intro}, Hamiltonian Monte Carlo (HMC) is the current state of the art technique for sampling from high dimensional probabilistic models over continuous state spaces.  In this chapter, two extensions to Hamiltonian Monte Carlo which allow more rapid exploration of the state space are presented.    Material in this chapter is taken from \cite{Sohl-Dickstein2012}.

\section{Reduced Momentum Flips}

Hamiltonian dynamics with partial momentum refreshment, in the style of \cite{Horowitz1991}, explore the state space more slowly than they otherwise would due to the momentum reversals which occur on proposal rejection.  These cause trajectories to double back on themselves, leading to random walk behavior on timescales longer than the typical rejection time, and leading to slower mixing.  I present a technique by which the number of momentum reversals can be reduced.  This is accomplished by maintaining the net exchange of probability between states with opposite momenta, but reducing the rate of exchange in both directions such that it is 0 in one direction.  An experiment illustrates these reduced momentum flips accelerating mixing for a particular distribution.

\subsection{Formalism}\label{formalism} 

A state $\mb \zeta\in R^{N\times 2}$ consists of a position $\mb x\in\mathcal R^N$ and an auxiliary momentum $\mb v\in\mathcal R^N$, $\mb \zeta = \left\{ \mb x, \mb v \right\}$.  The state space has an associated Hamiltonian
\begin{align}
H\left( \zeta \right) &= E\left( \mb x \right) + \frac{1}{2} \mb v^T \mb v
,
\end{align}
and a joint probability distribution
\begin{align}
p\left( \mb x, \mb v \right) &= p\left( \zeta \right) = \frac{1}{Z} \exp\left( -H\left( \zeta \right) \right)
,
\end{align}
where the normalization constant $Z$ is the partition function.

The momentum flip operator $F: \mathcal R^{N\times 2} \rightarrow \mathcal R^{N\times 2}$ negates the momentum.  It has the properties:
\begin{itemize}
  \item $F$ negates the momentum, $F \zeta = F\left\{ \mb x, \mb v\right\} = \left\{ \mb x, -\mb v\right\}$
  \item $F$ is its own inverse, $F^{-1} = F$, $FF \zeta = \zeta$.
  \item $F$ is volume preserving, $\det \left(\pd{ \left( F\zeta\right) }{\zeta^T}  \right) = 1$
  \item $F$ doesn't change the probability of a state, $p\left(\zeta\right) = p\left(F\zeta\right)$
\end{itemize}

The leapfrog integrator $L\left( n, \epsilon \right): \mathcal R^{N\times 2} \rightarrow \mathcal R^{N\times 2}$ integrates Hamiltonian dynamics for the Hamiltonian $H\left( \zeta \right)$, using leapfrog integration, for $n\in\mathcal Z^+$ integration steps with stepsize $\epsilon\in\mathcal R^+$.  We assume that $n$ and $\epsilon$ are constants, and write this operator simply as $L$.  The leapfrog integrator $L$ has the following relevant properties:
\begin{itemize}
  \item $L$ is volume preserving, $\det \left(\pd{ \left( L\zeta\right) }{\zeta^T}  \right) = 1$
  \item $L$ is exactly reversible using momentum flips, $L^{-1} = FLF$, $\zeta = FLFL\zeta$
\end{itemize}

During sampling, state updates are performed using a transition operator $T\left( r\right): \mathcal R^{N\times 2} \rightarrow \mathcal R^{N\times 2}$, where $r\sim U\left( [0, 1) \right)$ is drawn from the uniform distribution between 0 and 1,
\begin{align}
T\left( r\right) \zeta
 =
	\left\{\begin{array}{ccc}
L \zeta & & r < P_{leap}\left( \zeta \right) \\
F \zeta & & P_{leap} \leq r < P_{leap}\left( \zeta \right) + P_{flip}\left( \zeta \right) \\
\zeta    & & P_{leap} + P_{flip}\left( \zeta \right) \leq r
	\end{array}\right.
\label{trans}
.
\end{align}
$T\left( r\right)$ additionally depends on an acceptance probability for the leapfrog dynamics, $P_{leap}\left( \zeta \right)\in[0,1]$, and a probability of negating the momentum, $P_{flip}\left( \zeta \right)\in[0,1-P_{leap}\left( \zeta \right)]$.  These must be chosen to guarantee that $p\left( \zeta \right)$ is a fixed point of $T$.\footnote{This fixed point requirement can be written as $p\left( \zeta \right) = \int d{\zeta'} p\left( \zeta' \right) \int_0^1 dr \delta\left( \zeta - T\left(r\right) \zeta' \right)$.}

\begin{figure}
\begin{center}
\parbox[c]{0.95\linewidth}{
\begin{tabular}{cc}
\begin{tabular}{c}\includegraphics[width=0.46\linewidth]{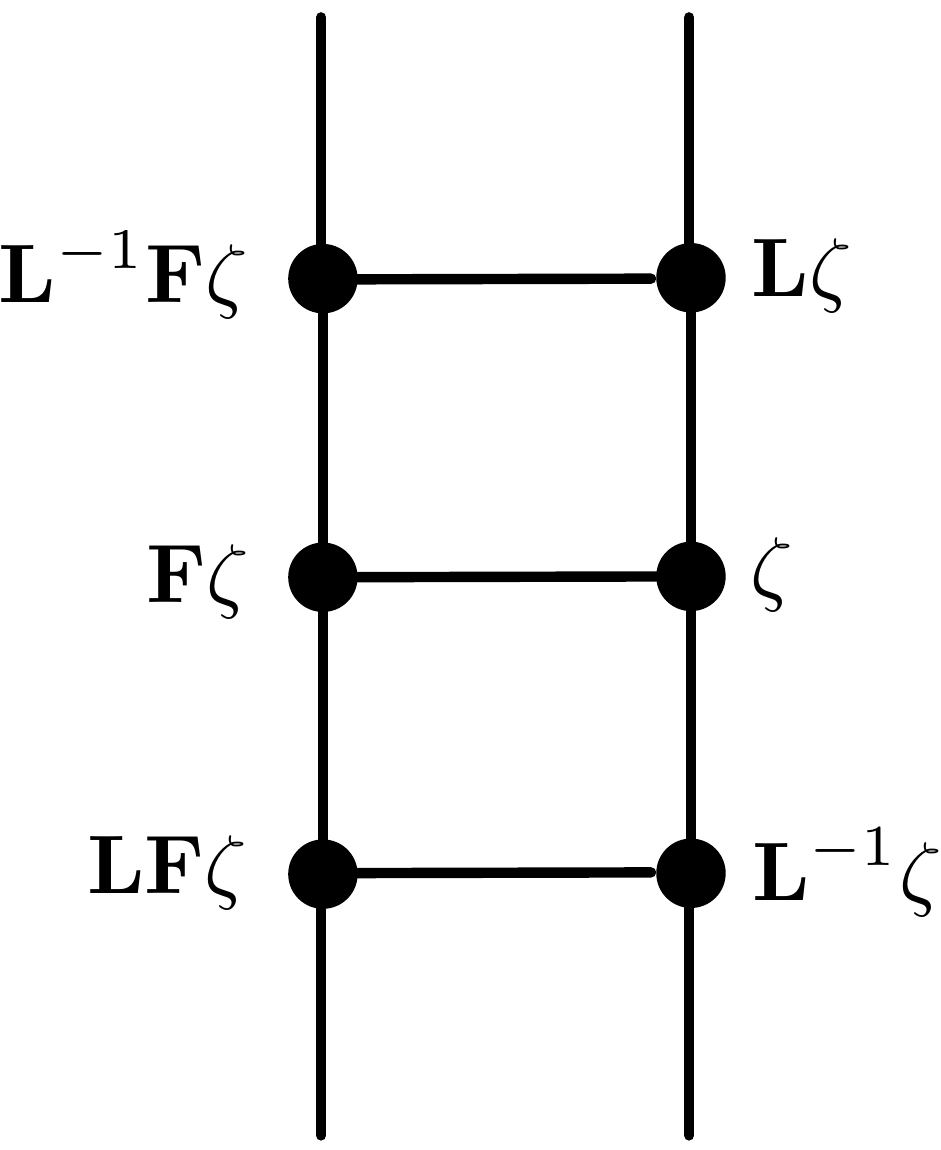}\\(a)\end{tabular}
&
\begin{tabular}{c}\includegraphics[width=0.27\linewidth]{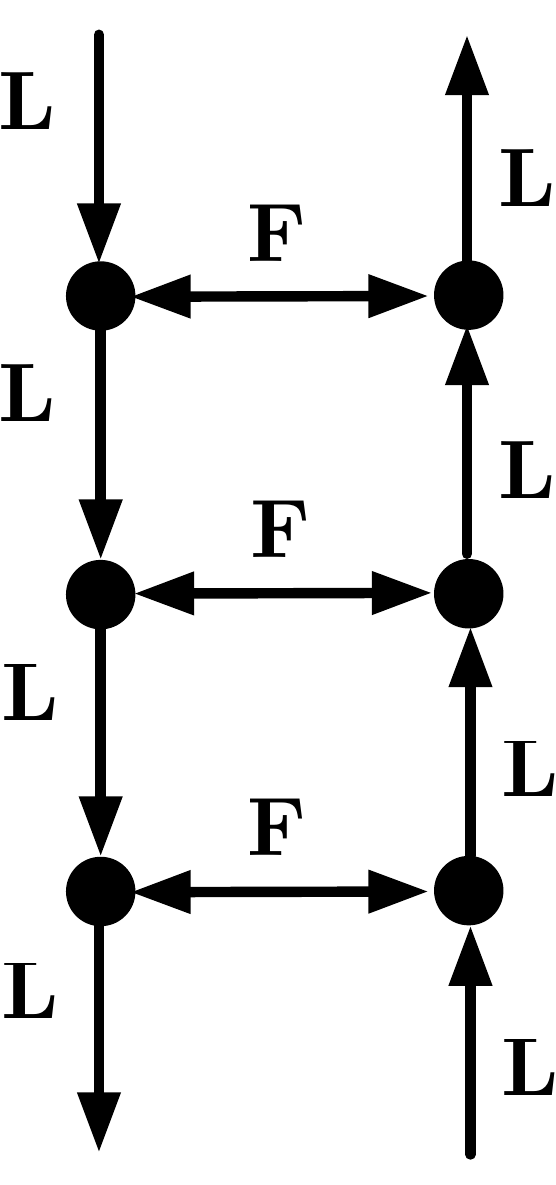}\\(d)\end{tabular}
\end{tabular}
} 
\end{center}
\caption{
This diagram illustrates the possible transitions between states using the Markov transition operator from Equation \ref{trans}.  In \emph{(a)} the relevant states, represented by the nodes, are labeled.  In \emph{(b)} the possible transitions, represented by the arrows, are labeled.    In Section \ref{fixed point}, the net probability flow into and out of the state $\zeta$ is set to 0.
}
\label{transitions}
\end{figure}

\subsection{Making the distribution of interest a fixed point}\label{fixed point}

In order to make $p\left(\zeta\right)$ a fixed point, we will choose the Markov dynamics $T$ so that on average as many transitions enter as leave state $\zeta$ at equilibrium.  This is {\em not} pairwise detailed balance --- instead we are directly enforcing zero net change in the probability of each state by summing over all allowed transitions into or out of the state.  This constraint is analogous to Kirchhoff's current law, where the total current entering a node is set to 0.  As can be seen from Equation \ref{trans} and the definitions in Section \ref{formalism}, and as is illustrated in Figure \ref{transitions}, 
a state $\zeta$ can only lose probability to the two states $L \zeta$ and $F \zeta$, and gain probability from the two states $L^{-1} \zeta$ and $F^{-1}\zeta$.  Equating the rates of probability inflow and outflow, we find
\begin{align}
p\left( \zeta \right) P_{leap}\left( \zeta \right) +
p\left( \zeta \right) P_{flip}\left( \zeta \right)
&=
p\left( L^{-1}\zeta \right) P_{leap}\left( L^{-1}\zeta \right) +
p\left( F^{-1}\zeta \right) P_{flip}\left( F^{-1}\zeta \right)
\\
&=
p\left( L^{-1}\zeta \right) P_{leap}\left( L^{-1}\zeta \right) +
p\left( \zeta \right) P_{flip}\left( F\zeta \right)
\\
P_{flip}\left( \zeta \right) - P_{flip}\left( F\zeta \right)
&=
\frac{p\left( L^{-1}\zeta \right)}{p\left( \zeta \right) } P_{leap}\left( L^{-1}\zeta \right) - P_{leap}\left( \zeta \right)
\label{balance intermediate}
.
\end{align}

We choose the standard Metropolis-Hastings acceptance rules for $P_{leap}\left( \zeta \right)$,
\begin{align}
P_{leap}\left( \zeta \right) = \min\left( 1, \frac{p\left(L\zeta\right)}{p\left(\zeta\right)} \right)
.
\end{align}
Substituting this in to Equation \ref{balance intermediate}, we find
\begin{align}
P_{flip}\left( \zeta \right) - P_{flip}\left( F\zeta \right)
&=
\frac{p\left( L^{-1}\zeta \right)}{p\left( \zeta \right) } \min\left( 1, \frac{p\left(LL^{-1}\zeta\right)}{p\left(L^{-1}\zeta\right)} \right) - \min\left( 1, \frac{p\left(L\zeta\right)}{p\left(\zeta\right)} \right)
\\
&=
\min\left( 1, \frac{p\left( L^{-1}\zeta \right)}{p\left( \zeta \right) } \right) - \min\left( 1, \frac{p\left(L\zeta\right)}{p\left(\zeta\right)} \right)
\\
&=
\min\left( 1, \frac{p\left( LF\zeta \right)}{p\left( \zeta \right) } \right) - \min\left( 1, \frac{p\left(L\zeta\right)}{p\left(\zeta\right)} \right)
\label{balance}
.
\end{align}
Satisfying Equation \ref{balance} we choose\footnote{To recover standard HMC, instead set $P_{flip}\left( \zeta \right) = 1 - P_{leap}\left( \zeta \right)$.  One can verify by substitution that this satisfies Equation \ref{balance}.}
 the following form for $P_{flip}\left( \zeta \right)$,
\begin{align}
P_{flip}\left( \zeta \right)
&=
\max\left( 0,
\min\left( 1, \frac{p\left( LF\zeta \right)}{p\left( \zeta \right) } \right) - \min\left( 1, \frac{p\left(L\zeta\right)}{p\left(\zeta\right)} \right)
\right)
\label{pflip}
.
\end{align}
Note that $P_{flip}\left( \zeta \right) \leq 1 - P_{leap}\left( \zeta \right)$, where $1 - P_{leap}\left( \zeta \right)$ is the rejection rate, and thus the momentum flip rate, in standard HMC.  Using this form for $P_{flip}\left( \zeta \right)$ will generally reduce the number of momentum flips required.

\begin{figure}
\begin{center}
\parbox[c]{0.6\linewidth}{
\includegraphics[width=1.0\linewidth]{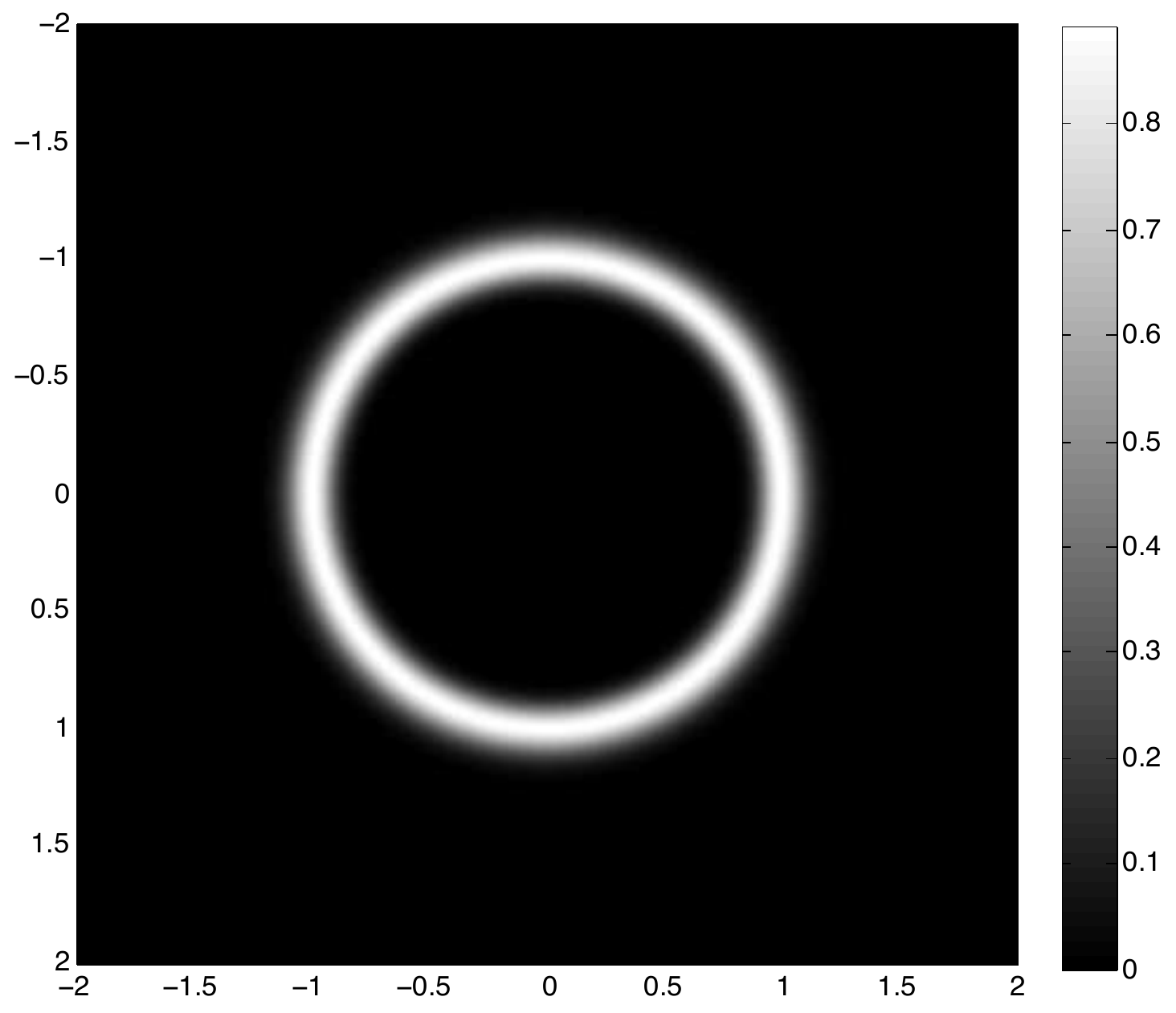}
} 
\end{center}
\caption{
A two dimensional image of the distribution used in Section \ref{ex section}.  Pixel intensity corresponds to the probability density function at that location.
}
\label{distribution}
\end{figure}

\begin{figure}
\begin{center}
\parbox[c]{0.6\linewidth}{
\includegraphics[width=1.0\linewidth]{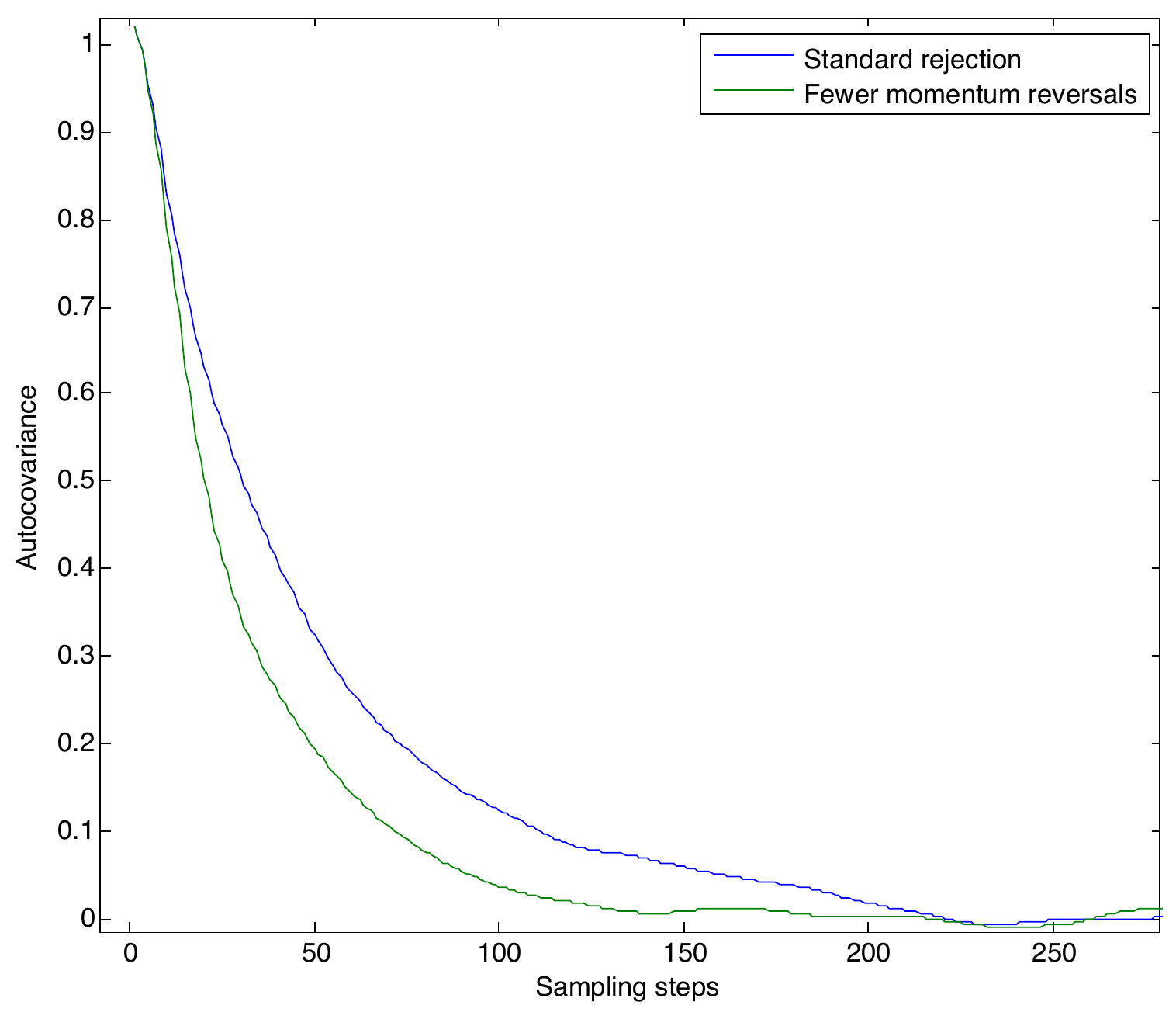}
} 
\end{center}
\caption{
The covariance between samples as a function of the number of intervening sampling steps for HMC with standard rejection and rejection with fewer momentum reversals.  Reducing the number of momentum reversals causes faster mixing, as evidenced by the faster falloff of the autocovariance.
}
\label{autocovariance}
\end{figure}
\subsection{Example} \label{ex section}

In order to demonstrate the accelerated mixing provided by this technique, samples were drawn from a simple distribution with standard rejection, and with separate rejection and momentum flipping rates as described above.  In both cases, the leapfrog step length $\epsilon$ was set to 0.1, the number of integration steps $n$ was set to 1, and the momentum corruption rate $\beta$ was set so as to corrupt half the momentum per unit stimulation time.  Both samplers were run for $100,000$ sampling steps.  The distribution used was described by the energy function
\begin{align}
E &= 100 \log^2\left( \sqrt{x_1^2 + x_2^2} \right)
.
\end{align}
A 2 dimensional image of this distribution can be seen in Figure \ref{distribution}.  The autocovariance of the returned samples can be seen, as a function of the number of intervening sampling steps, in Figure \ref{autocovariance}.  Sampling using the technique presented here led to more rapid decay of the autocovariance, consistent with faster mixing.

\chapter{Conclusion}
\label{chap conclusion}

Scientific progress is driven by our ability to build models of the world.  When investigating complex or large systems, the tools available to build probabilistic models are frequently inadequate.  In this thesis I have introduced several new tools that address some of the most pressing issues in probabilistic modeling. 

Minimum Probability Flow learning (MPF) is a novel, general purpose framework for parameter estimation in probabilistic models that outperforms current techniques in both learning speed and accuracy. MPF works for any parametric model without hidden state variables, including those over both continuous and discrete state space systems.  It avoids explicit calculation of the partition function by employing deterministic dynamics in place of the slow sampling required by many existing approaches.
Because MPF provides a simple and well-defined objective function, it can be minimized quickly using existing higher order gradient descent techniques.
 Furthermore, the objective function is convex for models in the exponential family, ensuring that the global minimum can be found with gradient descent.
Extensions to MPF allow it to be used in conjunction with sampling algorithms and persistent particles for even faster performance.  
Minimum velocity learning, score matching, and some forms of contrastive divergence are special cases of MPF for specific choices for its dynamics.

The natural gradient is a powerful concept, but can be difficult to understand in its traditional presentation.  I have made a connection between the natural gradient and the common concept of signal whitening, and additionally provided cookbook techniques for the application of the natural gradient to learning problems.  This should lower the barrier to understanding and using this technique in learning problems.  Both the natural gradient and MPF allow model fitting to be performed more quickly and accurately, and in situations in which it was previously impractical or impossible.

Hamiltonian Annealed Importance Sampling (HAIS) allows the partition function of non-analytically-normalizable probabilistic models to be computed many times faster than with competing techniques.  By improving upon the available methods for partition function estimation, it makes it possible to directly compare large probabilistic models in terms of the likelihoods they assign to data. This is a fundamental measure of the quality of a model, but one which is very frequently neglected in the literature due to practical and computational limitations.  It is my hope that HAIS will lead to more meaningful comparisons between models.

Improvements to Hamiltonian Monte Carlo sampling make many tasks, such as averaging over a distribution, more practical for complex and computationally expensive probabilistic models.  I have reduced the time required to generate independent samples from a distribution via Hamiltonian Monte Carlo, by reducing the frequency of momentum flips which cause the sampler to retrace its steps.  This will improve the practicality of sampling from a distribution, and lead to more frequent use of samples rather than less accurate approximations or maximum a posteriori estimates.

It is my hope that, taken together, these contributions will improve the ability of scientists and engineers to build and manipulate probabilistic models of the world.

\begin{appendices}
\chapter{Derivation of MPF objective by Taylor expanding KL divergence}
\label{app:MPF Taylor}
\label{app:KL}

The minimum probability flow learning objective function $K\left(\theta\right)$ is found by taking up to the first order terms in the Taylor expansion of the KL divergence between the data distribution and the distribution resulting from running the dynamics for a time $\epsilon$: 
\begin{align}
K\left( \theta \right) & \approx  D_{KL}\left(
   \mb{p^{(0)}} ||\mb{p^{(t)}}
   \left(\theta\right)\right)\Big |_{t=0}
\nonumber \\ & \qquad
   + \epsilon \frac
   {\partial D_{KL}\left(
   \mb{p^{(0)}} ||\mb{p^{(t)}}
   \left(\theta\right)\right)}
   {\partial t}\Big |_{t=0} \\
&= 0
   + \epsilon \frac
   {\partial D_{KL}\left(
   \mb{p^{(0)}} ||\mb{p^{(t)}}
   \left(\theta\right)\right)}
   {\partial t}\Big |_{t=0} \\
&= \epsilon \pd{}{t}\left.\left(\sum_{i\in \mathcal{D}} \p{i}{0}\log\frac{\p{i}{0}}{\p{i}{t}}\right)\right|_0 \\
&= -\epsilon \sum_{i\in \mathcal{D}}\frac{\p{i}{0}}{\p{i}{0}}\left.\pd{\p{i}{t}}{t}\right|_0 \\
&= -\epsilon \left.\sum_{i\in \mathcal{D}}\pd{\p{i}{t}}{t}\right|_0 \\\displaybreak[0]
&= -\epsilon \left.\left(\pd{}{t}\sum_{i\in \mathcal{D}}\p{i}{t}\right)\right|_0  \label{eqn:sumrate} \\ \displaybreak[0]
&= -\epsilon \left.\pd{}{t}\left(1-\sum_{i\notin \mathcal{D}}\p{i}{t}\right)\right|_0 \\ \displaybreak[0]
&= \epsilon \left.\sum_{i\notin \mathcal{D}}\pd{\p{i}{t}}{t}\right|_0 \\\displaybreak[0]
&= \epsilon \sum_{i\notin \mathcal{D}}\sum_{j\in \mathcal{D}}\Gamma_{ij}\p{j}{0} \\
&= \frac{\epsilon}{|\mathcal{D}|} \sum_{i\notin \mathcal{D}}\sum_{j\in \mathcal{D}}\Gamma_{ij},
\end{align}
where we used the fact that $\sum_{i\in \mathcal{D}} p_i^{(t)} + \sum_{i\notin \mathcal{D}} p_i^{(t)} = 1$.
This implies that the rate of growth of the KL divergence at time $t=0$ equals the total initial flow of probability from states with data into states without.

\chapter{Convexity of MPF objective function}
\label{app:convex}

As observed by Macke and Gerwinn \cite{macke}, the MPF objective function is convex for models in the exponential family.

We wish to minimize
\begin{align}
K &=
   \sum_{i \in D} \sum_{j \in D^C} \Gamma_{ji} p_i^{(0)} .
\end{align}

$K$ has derivative
\begin{align}
\pd{K}{\theta_m} &= \sum_{i \in D}\sum_{j \in D^c} \left( \pd{\Gamma_{ij}}{\theta_m} \right) p_i^{(0)} \\
&= \frac{1}{2} \sum_{i \in D}\sum_{j \in D^c} \Gamma_{ij} \left(  \pd{E_j}{\theta_m} - \pd{E_i}{\theta_m}  \right) p_i^{(0)},
\end{align}
and Hessian
\begin{align}
\pd{^2 K}{\theta_m \partial \theta_n}
   &=
   	\frac{1}{4} \sum_{i \in D}\sum_{j \in D^c} \Gamma_{ij} \left(  \pd{E_j}{\theta_m} - \pd{E_i}{\theta_m}  \right)\left(  \pd{E_j}{\theta_n} - \pd{E_i}{\theta_n}  \right) p_i^{(0)}  \nonumber \\
   	&+
   	\frac{1}{2} \sum_{i \in D}\sum_{j \in D^c} \Gamma_{ij} \left(  \pd{^2 E_j}{\theta_m \partial \theta_n} - \pd{^2 E_i}{\theta_m \partial \theta_n}  \right) p_i^{(0)} .
\end{align}
The first term in the Hessian is a weighted sum of outer products, with non-negative weights $\frac{1}{4} \Gamma_{ij} p_i^{(0)}$, and is thus positive semidefinite.  The second term is $0$ for models in the exponential family (those with energy functions linear in their parameters).

Parameter estimation for models in the exponential family is therefore convex using minimum probability flow learning.

\chapter{Lower Bound on Log Likelihood Using MPF}
\label{app:bound}

In this appendix, a lower bound on the log probability of data states is derived in terms of the MPF objective function.  Although this bound is of theoretical interest, evaluating it requires calculating the first non-zero eigenvalue of the probability flow matrix $\mb \Gamma$.  This is typically an intractable task, and is equivalent to computing the mixing time for a Markov Chain Monte Carlo algorithm.

The probability flow matrix $\mb \Gamma$ can be written 
\begin{align}
\mb \Gamma &= \mb D + \mb V \mb g \mb V^{-1}
,
\end{align}
where $\mb V$ is a diagonal matrix with entries $V_{ii} = \exp \left[ -\frac{1}{2} \mb E_i  \right]$, $\mb D$ is a diagonal matrix with entries $D_{ii} = -\sum_j \left[ \mb V \mb g \mb V^{-1} \right]_{ji}$, and $\mb g$ is the symmetric connectivity matrix.

As observed by Surya Ganguli (personal communication), we can relate $\mb \Gamma$ to a symmetric matrix $\mb B$ by an eigenvalue-maintaining transformation,
\begin{align}
\mb B &= \mb V^{-1} \mb \Gamma \mb V \\
\mb \Gamma &= \mb V \mb B \mb V^{-1}
.
\end{align}
The eigendecomposition of $\mb B$ is
\begin{align}
\mb B &= \mb U^{\mb B} \mb \Lambda \left( \mb U^{\mb B} \right)^T
,
\end{align}
where the eigenvalues $\mb \Lambda$ are identical to the eigenvalues for $\mb \Gamma$.  Single eigenvalues will be written $\lambda_i \equiv \Lambda_{ii}$.  The eigenvectors $\mb U^{\mb B}$ of $\mb B$ are orthogonal and taken to be unit length.  The lengths of the eigenvectors for $\mb \Gamma$ are chosen relative to the eigenvectors for $\mb B$,
\begin{align}
\mb U^{\mb \Gamma} &= \mb V \mb U^{\mb B}
.
\end{align}
Additionally, the eigenvectors and eigenvalues are assumed to be sorted in decreasing order, with $\lambda_1 = 0$, and the corresponding eigenvector of $\mb\Gamma$, $U^\Gamma_{\cdot 1}$, being a scaled version of the model distribution 
\begin{align}
p_i^{(\infty)} &=
\frac{
	\exp\left[  - E_i \right]
	}
	{\sum_j 
				\exp\left[  - E_j \right] 
	} 
,
\end{align}
with the scaling factor determined by the unit length constraint on $U^{\mb B}_{\cdot 1}$.  The analytic form for the 1st eigenvector of both $\mb \Gamma$ and $\mb B$ is,
\begin{align}
U^{\mb B}_{i 1} &= \frac
			{V^{-1}_{ii} p_i^{(\infty)} }
			{\left( \sum_j 
				\left( V_{jj}^{-1} p_j^{(\infty)} \right)^2  \right)^\frac{1}{2}}
= \frac{
	\exp\left[  - \frac{1}{2} E_i \right]
	}
	{\left( \sum_j 
				\exp\left[  - E_j \right] \right)^\frac{1}{2}
	} \\
U^{\mb \Gamma}_{i 1} &= 
\exp \left[ -\frac{1}{2} E_i  \right]  U^{\mb B}_{i 1} 
= \frac{
	\exp\left[  - E_i \right]
	}
	{\left( \sum_j 
				\exp\left[  - E_j \right] \right)^\frac{1}{2}
	} 
	.
\end{align}
The log probability can then be related to the eigenvector $U^{\mb B}_{\cdot 1}$,
\begin{align}
\log p_i^{(\infty)} & = \log U^{\mb B}_{i 1} 
- \frac{1}{2}E_i  
+ \log \frac{
	\left( \sum_j 
				\exp\left[  - E_j \right] \right)^\frac{1}{2}
	}{
	 \sum_j 
				\exp\left[  - E_j \right] 
	} \\
	& = \log U^{\mb B}_{i 1} 
+ \frac{1}{2}\left(  -E_i  -
	 \log \sum_j 
				\exp\left[ - E_j \right] 
				\right)
		\\
	& = 2\log U^{\mb B}_{i 1}
		\label{bound 1 equality}
.
\end{align}

We now relate the entries in $U^{\mb B}_{\cdot 1}$ to the initial flow rates $\pd{p_i^{(0)}}{t}$, restricted to data states $i \in \mathcal D$.  We make the simplifying assumption that the data is sufficiently sparse, and the connectivity matrix $\mb g$ has been chosen in such a way, that there is no direct flow of probability between data states.  Under this assumption, and for data states $i \in \mathcal D$,
\begin{align}
\pd{p_i^{(0)}}{t} &= \Gamma_{ii} p_i^{(0)} 
\label{dpdt intro}
.
\end{align}
Noting that $\Gamma$ can be written $\mb V \mb U^{\mb B} \mb \Lambda \left( \mb U^{\mb B} \right)^T \mb V^{-1}$, we expand Equation \ref{dpdt intro},
\begin{align}
\pd{p_i^{(0)}}{t} &= V_{ii} \sum_{j=1}^N U^{\mb B}_{i j} \lambda_j U^{\mb B}_{i j} V_{ii}^{-1}  p_i^{(0)} \\
& =  p_i^{(0)} \sum_{j=1}^N \left(U^{\mb B}_{i j}\right)^2 \lambda_j
.
\end{align}
Remembering that the eigenvalues are in decreasing order, with $\lambda_1 = 0$ and the remaining eigenvalues negative, and also remembering that $\mb U^{\mb B}$ is orthonormal, we write the inequality
\begin{align}
\pd{p_i^{(0)}}{t} &\leq p_i^{(0)}  \lambda_{2}  \sum_{j=2}^N \left(U^{\mb B}_{i j}\right)^2 =  p_i^{(0)}  \lambda_{2}  \left[ 1 - \left(U^{\mb B}_{i 1}\right)^2 \right] \\
U^{\mb B}_{i 1} & \geq 
\left( \max \left[ 0, 
\left(
1 - {
	\pd{p_i^{(0)}}{t}
	}{
	\left(p_i^{(0)} \lambda_{2}  \right)^{-1}
	}
\right) \right]
\right)^\frac{1}{2}
. \label{bound 2}
\end{align}

Combining Equations \ref{bound 1 equality} and \ref{bound 2}, for data states $i \in \mathcal D$ we find
\begin{align}
\log p_i^{(\infty)} \geq \log \max \left[ 0, 
\left(
1 - {
	\pd{p_i^{(0)}}{t}
	}{
	\left(p_i^{(0)}  \lambda_{2}  \right)^{-1}
	}
\right) \right]
 .
 \label{log p bound}
\end{align}

For sufficiently small magnitude values of ${
	\pd{p_i^{(0)}}{t}
	}{
	\left(p_i^{(0)}  \lambda_{2}  \right)^{-1}
	}$ (for instance, because of very small probability flow $\pd{p_i^{(0)}}{t}$, or very negative first non-zero eigenvalue $\lambda_2$), we can make the approximation
\begin{align}
\log \max \left[ 0, 
\left(
1 - {
	\pd{p_i^{(0)}}{t}
	}{
	\left(p_i^{(0)}  \lambda_{2}  \right)^{-1}
	}
\right) \right]
& \approx
- {
	\pd{p_i^{(0)}}{t}
	}{
	\left(p_i^{(0)}  \lambda_{2}  \right)^{-1}
	}
.
\end{align}	

The mixing time of a Monte Carlo algorithm with a transition matrix corresponding to $\mb\Gamma$ can be upper and lower bounded using $\lambda_2$.  However, $\lambda_2$ is generally difficult to find, and frequently little can be said about Monte Carlo mixing times.  Equation \ref{log p bound} can be rewritten as a bound on $\lambda_2$ as follows,
\begin{align}
p_i^{(\infty)} & \geq
1 - {
	\pd{p_i^{(0)}}{t}
	}{
	\left(p_i^{(0)}  \lambda_{2}  \right)^{-1}
	}
	\\
\lambda_{2} \left( p_i^{(\infty)} - 1 \right) & \leq
- {
	\pd{p_i^{(0)}}{t}
	}
	\frac{1}{p_i^{(0)}}
	\\
\lambda_{2} & \geq
 {
	\pd{p_i^{(0)}}{t}
	}
	\frac{1}{p_i^{(0)}}
\frac{1}
{\left( 1 - p_i^{(\infty)}  \right)}
\\
\pd{p_i^{(0)}}{t} & =
-\sum_{j\neq i} g_{ji} \exp\left( \frac{1}{2} \left[ E_i - E_j \right] \right) p_i^{(0)}
\\
\lambda_{2} & \geq
\frac{
-\sum_{j\neq i} g_{ji} \exp\left( \frac{1}{2} \left[ E_i - E_j \right] \right)
}
{\left( 1 - p_i^{(\infty)}  \right)}.
\label{final lambda}
\end{align}
$p_i^{(\infty)}$ will typically be much smaller than one, and replacing it with an upper bound will thus have only a small effect on the tightness of the bound in Equation \ref{final lambda},
\begin{align}
\lambda_{2} & \geq
\frac{
-\sum_{j\neq i} g_{ji} \exp\left( \frac{1}{2} \left[ E_i - E_j \right] \right)
}
{\left( 1 - p_i^{(\infty)}  \right)}.
\geq
\frac{
-\sum_{j\neq i} g_{ji} \exp\left( \frac{1}{2} \left[ E_i - E_j \right] \right)
}
{\left( 1 - p_{i\textrm{\ upper\ bound}}^{(\infty)}  \right)}
.
\label{final lambda bound}
\end{align}
Equation \ref{final lambda bound} holds for any system state $i$, and the bound on $\lambda_2$ can be written in terms of the tightest bound for any system state,
\begin{align}
\lambda_{2} & \geq
\max_i \left[
\frac{
-\sum_{j\neq i} g_{ji} \exp\left( \frac{1}{2} \left[ E_i - E_j \right] \right)
}
{\left( 1 - p_{i\textrm{\ upper\ bound}}^{(\infty)}  \right)}
\right]
.
\label{final lambda bound max}
\end{align}

\chapter{Score Matching (SM) is a special case of MPF}
\label{app SM deriv}
\label{app:score_matching}

Score matching, developed by Aapo Hyv\"arinen \cite{Hyvarinen05}, is a method that learns parameters in a probabilistic model using only derivatives of the energy function evaluated over the data distribution (see Equation \eqref{eq:score matching}).  This sidesteps the need to explicitly sample or integrate over the model distribution. In score matching one minimizes the expected square distance of the score function with respect to spatial coordinates given by the data distribution from the similar score function given by the model distribution.  A number of connections have been made between score matching and other learning techniques \cite{Hyvarinen:2007p5984,sohldickstein,Movellan:2008p7643,siwei2009}.  Here we show that in the correct limit, MPF also reduces to score matching.

For a $d$-dimensional, continuous state space, we can write the MPF objective function as
\begin{align}
K_{\mathrm{MPF}} & = \frac{1}{N}\sum_{x\in\mathcal{D}} \int \dd^d y\; \Gamma(y,x)\nonumber \\
&= \frac{1}{N}\sum_{x\in\mathcal{D}} \int \dd^d y\; g(y,x) e^{(E(y|\theta)-E(x|\theta))},
\end{align}
where the sum $\sum_{x\in\mathcal{D}}$ is over all data samples, and $N$ is the number of samples in the data set $\mathcal{D}$.
Now we assume that transitions are only allowed from states $x$ to states $y$ that are within
a hypercube of side length $\epsilon$ centered around $x$ in state space. (The master equation
will reduce to Gaussian diffusion as $\epsilon\rightarrow 0$.) Thus, the function $g(y,x)$ will
equal 1 when $y$ is within the $x$-centered cube (or $x$ within the $y$-centered cube) and 0 otherwise. Calling this cube $C_\epsilon$,
and writing $y=x+\alpha$ with $\alpha\in C_\epsilon$, we have
\begin{align}
K_{\mathrm{MPF}} = \frac{1}{N}\sum_{x\in\mathcal{D}} \int_{C_\epsilon} \dd^d \alpha\;
e^{(E(x+\alpha|\theta)-E(x|\theta))}.
\end{align}
If we Taylor expand in $\alpha$ to second order and ignore cubic and higher terms, we get
\begin{align}
K_{\mathrm{MPF}} &\approx \frac{1}{N}\sum_{x\in\mathcal{D}} \int_{C_\epsilon} \dd^d\alpha\;
(1) \nonumber\\
&- \frac{1}{N}\sum_{x\in\mathcal{D}} \int_{C_\epsilon} \dd^d\alpha\;
\frac{1}{2}\sum_{i=1}^d\alpha_i\nabla_{x_i}E(x|\theta)\nonumber \\
&+\frac{1}{N}\sum_{x\in\mathcal{D}} \int_{C_\epsilon} \dd^d\alpha\;
\frac{1}{4}\Biggl(\frac{1}{2}\biggl[\sum_{i=1}^d
\alpha_i\nabla_{x_i}E(x|\theta)\biggr]^2 \nonumber \\
&\quad-\sum_{i,j=1}^d\alpha_i\alpha_j\nabla_{x_i}\nabla_{x_j}E(x|\theta)\Biggr).
\end{align}
This reduces to
\begin{align}
K_{\mathrm{MPF}}&\approx\frac{1}{N}\sum_{x\in\mathcal{D}}\Biggl[\epsilon^d + \frac{1}{4}\Biggl(\frac{1}{2}\frac{2}{3}\epsilon^{d+2}
\biggl[\sum_{i=1}^d\nabla_{x_i}E(x|\theta)\biggr]^2 \nonumber \\ &\quad-\frac{2}{3}\epsilon^{d+2}\nabla_{x_i}^2E(x|\theta)\Biggr)\Biggr],
\end{align}
which, removing a constant offset and scaling factor, is exactly equal to the score matching objective function,
\begin{align}
K_{\mathrm{MPF}} &\sim \frac{1}{N}\sum_{x\in\mathcal{D}}
\biggl[ \frac{1}{2}\nabla E(x|\theta)\cdot\nabla E(x|\theta) -\nabla^2E(x|\theta)\biggr]\label{eq:score matching}\\
&= K_{\mathrm{SM}}
.
\end{align}
Score matching is thus equivalent to MPF when the connectivity function $g(y,x)$ is non-zero only for states infinitesimally close to each other. It should be noted that the score matching estimator has a closed-form solution when the model distribution belongs to the exponential family \cite{hyvarinen2007}, so the same can be said for MPF in this limit.

\chapter{MPF objective function for an Ising model}
\label{app MPF Ising}

This appendix derives the MPF objective function for the case of an Ising model.  In Section \ref{sec single bit}, connectivity is set between all states which differ by a single bit flip.  In Section \ref{sec all bit}, an additional connection is included between states which differ in all bits.  This additional connection is particularly beneficial in cases (such as spike train data) where unit activity is extremely sparse.  Code implementing MPF for the Ising model is available at \cite{MPFcode}.

The MPF objective function is
\begin{align}
K\left( \mb J \right) & = \sum_{\mb x \in \mathcal D} \sum_{\mb x' \notin \mathcal D} g\left( \mb x, \mb x' \right) \exp\left( \frac{1}{2}\left[
E(\mb x; \mb J ) - E(\mb x'; \mb J) \right] \right)
,
\end{align}
where $g\left( \mb x, \mb x' \right) = g\left( \mb x', \mb x \right) \in \left\{ 0, 1 \right\}$ is the connectivity function, $E(\mb x; \mb J )$ is an energy function parameterized by $\mb J$, and $\mathcal D$ is the list of data states.  For the Ising model, the energy function is
\begin{align}
E\left( \mb x; \mb J\right) = \mb x^T \mb J \mb x
\end{align}
where $\mb x \in \left\{ 0, 1 \right\}^N$, $\mb J \in \mathcal R^{N\times N}$, and $\mb J$ is symmetric ($\mb J = \mb J^T$).

\section{Single Bit Flips}\label{sec single bit}

We consider the case where the connectivity function $g\left( \mb x, \mb x' \right)$ is set to connect all states which differ by a single bit flip,
\begin{align}
g\left( \mb x, \mb x' \right)
 =
	\left\{\begin{array}{ccrl}
1 & & \mb x \mathrm{\ and\ } \mb x' \mathrm{\ differ\ by\ a\ single\ bit\ flip,\ }  & \sum_n \left| x_n - x_n' \right| = 1  \\
0 & & \mathrm{otherwise} & 
	\end{array}\right.
.
\end{align}
The MPF objective function in this case is
\begin{align}
K\left( \mb J \right) = \sum_{\mb x \in \mathcal D} \sum_{n=1}^N \exp\left( \frac{1}{2}\left[
E(\mb x; \mb J) - E(\mb x + {\mb d}(\mb x, n); \mb J) \right] \right)
\end{align}
where the sum over $n$ is a sum over all data dimensions, and the bit flipping function ${\mb d}(\mb x, n) \in \left\{ -1, 0, 1 \right\}^N$ is
\begin{align}
{\mb d}(\mb x, n)_i =
	\left\{\begin{array}{ccc}
0 & & i \neq n \\
-(2 x_i - 1) & & i = n
	\end{array}\right.
\end{align}

For the Ising model, this MPF objective function becomes (using the fact that $\mb J = \mb J^T$)
\begin{align}
K\left( \mb J \right) & = \sum_{\mb x \in \mathcal D} \sum_n \exp\left( \frac{1}{2}\left[
\mb x^T \mb J \mb x
- (\mb x + {\mb d}(\mb x, n))^T \mb J (\mb x + {\mb d}(\mb x, n))
\right] \right) \\
& = \sum_{\mb x \in \mathcal D} \sum_n \exp\left( \frac{1}{2}\left[
\mb x^T \mb J \mb x
- \left(
\mb x^T \mb J \mb x
+
2 \mb x^T \mb J {\mb d}(\mb x, n)
+
{\mb d}(\mb x, n)^T \mb J {\mb d}(\mb x, n)
\right) 
\right] \right)
 \\
& = \sum_{\mb x \in \mathcal D} \sum_n \exp\left( -\frac{1}{2}\left[
2 \mb x^T \mb J {\mb d}(\mb x, n)
+
{\mb d}(\mb x, n)^T \mb J {\mb d}(\mb x, n)
\right]
\right)  \\
& = \sum_{\mb x \in \mathcal D} \sum_n \exp\left( -\frac{1}{2}\left[
2 \sum_i x_i J_{in} \left( 1 - 2 x_n  \right)
+
J_{nn}
\right]
\right)\\
& = \sum_{\mb x \in \mathcal D} \sum_n \exp\left( \left[
\left( 2 x_n - 1 \right) \sum_i x_i J_{in}
-
\frac{1}{2}J_{nn}
\right]
\right)
\label{K final single bit}
.
\end{align}

Assume the symmetry constraint on $\mb J$ is enforced by writing it in terms of another possibly asymmetric matrix $\mb J' \in \mathcal R^{N\times N}$,
\begin{align}
\mb J = \frac{1}{2} \mb J' + \frac{1}{2} \mb {J'}^T
.
\end{align}
The derivative of the MPF objective function with respect to $\mb J'$ is
\begin{align}
\pd{K\left( \mb J' \right)}{{J'}_{lm}}  & =
\frac{1}{2}\sum_{\mb x \in \mathcal D} \exp\left( \left[
\left( 2 x_m - 1 \right) \sum_i x_i {J}_{im}
-
\frac{1}{2}{J}_{mm}
\right]
\right)
	\left[
		\left( 2 x_m - 1 \right) x_l
		-
		\delta_{lm} \frac{1}{2}
	\right] \nonumber \\ & \qquad 
+
\frac{1}{2}\sum_{\mb x \in \mathcal D} \exp\left( \left[
\left( 2 x_l - 1 \right) \sum_i x_i {J}_{il}
-
\frac{1}{2}{J}_{ll}
\right]
\right)
	\left[
		\left( 2 x_l - 1 \right) x_m
		-
		\delta_{ml} \frac{1}{2}
	\right]
,
\end{align}
where the second term is simply the first term with indices $l$ and $m$ reversed.

Note that both the objective function and gradient can be calculated using matrix operations (no for loops).  See the released Matlab code.

\section{All Bits Flipped}\label{sec all bit}

We consider the case where the connectivity function $g\left( \mb x, \mb x' \right)$ is set to connect all states which differ by a single bit flip, and all states which differ in all bits,
\begin{align}
g\left( \mb x, \mb x' \right)
 =
	\left\{\begin{array}{ccrl}
1 & & \mb x \mathrm{\ and\ } \mb x' \mathrm{\ differ\ by\ a\ single\ bit\ flip,\ } & \sum_n \left| x_n - x_n' \right| = 1 \\ 
1 & & \mb x \mathrm{\ and\ } \mb x' \mathrm{\ differ\ in\ all\ bits,\ } & \sum_n \left| x_n - x_n' \right| = N \\ 
0 & & \mathrm{otherwise} & 
	\end{array}\right.
.
\end{align}
This extension to the connectivity function aids MPF in assigning the correct relative probabilities between data states and states on the opposite side of the state space from the data, even in cases (such as sparsely active units) where the data lies only in a very small region of the state space.

MPF functions by comparing the relative probabilities of the data states and the states which are connected to the data states.  If there is a region of state space in which no data lives, and to which no data states are connected, then MPF is blind to that region of state space, and may assign an incorrect probability to it.  This problem has been observed fitting an Ising model to sparsely active neural data.  In this case, MPF assigns too much probability to states with many units on simultaneously.  However, if an additional connection is added between each state and the state with all the bits flipped, then there are comparison states available which have many units on simultaneously.  With this extra connection, MPF better penalizes non-sparse states, and the fit gets much better.

The modified objective function has the form, 
\begin{align}
K\left( \mb J \right) & = K_{single}\left( \mb J \right) + K_{all}\left( \mb J \right)
.
\end{align}
We can take the first term, which deals only with single bit flips, from Equation \ref{K final single bit},
\begin{align}
K_{single}\left( \mb J \right) & = \sum_{\mb x \in \mathcal D} \sum_n \exp\left( \left[
\left( 2 x_n - 1 \right) \sum_i x_i J_{in}
-
\frac{1}{2}J_{nn}
\right]
\right)
.
\end{align}
The second term is
\begin{align}
K_{all} & = \sum_{\mb x \in \mathcal D} \exp\left( \frac{1}{2}\left[
E(\mb x; \mb J) - E(\mb 1 - \mb x; \mb J) \right] \right) \\
& = \sum_{\mb x \in \mathcal D} \exp\left( \frac{1}{2}\left[
\mb x^T \mb J \mb x - 
\left( \mb 1 - \mb x\right)^T \mb J \left( \mb 1 - \mb x\right) \right] \right)
,
\end{align}
where $\mb 1$ is the vector of all ones.

The contribution to the derivative from the second term is
\begin{align}
\pd{K_{all}}{J_{lm}} & = \frac{1}{2}\sum_{\mb x \in \mathcal D} \exp\left( \frac{1}{2}\left[
\mb x^T \mb J \mb x - 
\left( \mb 1 - \mb x\right)^T \mb J \left( \mb 1 - \mb x\right) \right] \right)
\left[
x_l x_m - \left( 1 - x_l \right) \left( 1 - x_m \right)
\right]
.
\end{align}

\chapter{MPF objective function for a Restricted Boltzmann Machine (RBM)}
\label{app MPF RBM}

This appendix derives the MPF objective function for the case of a Restricted Boltzmann Machine (RBM), with the connectivity function $g_{ij}$ chosen to connect states which differ by a single bit flip.

The energy function over the visible
units for an RBM is found by marginalizing out the hidden units.  This
gives an energy function of:
\begin{align}
E(\mb x) &= -\sum_i \log ( 1 + \exp ( -W_i \mb x ) )
\end{align}
where $W_i$ is a vector of coupling parameters and $\mb x$ is the binary input
vector.  The MPF objective function for this is
\begin{align}
K &= \sum_{\mb x \in \mathcal D} \sum_n \exp\left( \frac{1}{2}\left[
E(\mb x) - E(\mb x + {\mb d}(\mb x, n)) \right] \right)
\end{align}
where the sum over $n$ indicates a sum over all data dimensions, and
the function ${\mb d}(\mb x, n)$ is
\begin{align}
{\mb d}(\mb x, n)_i =
	\left\{\begin{array}{ccc}
0 & & i \neq n \\
-(2 x_i - 1) & & i = n
	\end{array}\right.
\end{align}

Substituting into the objective function
\begin{align}
K = \sum_{\mb x \in \mathcal D} \sum_n \exp\left( \frac{1}{2}\left[
-\sum_i \log ( 1 + \exp ( -W_i \mb x ) )
+
\sum_i \log ( 1 + \exp ( -W_i \mb x + W_i {\mb d}(\mb x, n)  ) )
\right] \right)
\end{align}

Matlab code is available at \cite{MPFcode}.  It implements the sum over $n$ in a for loop, and
calculates the change in $W_i \mb x$ caused by $W_i {\mb d}( \mb x, n)$ for all samples simultaneously.  Note that the for loop could also
be performed over samples, with the change induced by each bit flip being
calculated by matrix operations.  If the code is run with a small
batch size, this implementation would be faster.  A clever programmer might find a way to replace both for loops with matrix operations.

\end{appendices}

\singlespace

\bibliography{thesis}
\bibliographystyle{named}

\printindex

\end{document}